%% file: acl_latex.tex
\pdfoutput=1
\documentclass[11pt]{article}

\usepackage[final]{acl}

\usepackage{times}
\usepackage{latexsym}

\usepackage[T1]{fontenc}
\usepackage[utf8]{inputenc}

\usepackage{microtype}

\usepackage{inconsolata}

\usepackage{graphicx}
\usepackage{amsmath}
\usepackage{amssymb}
\usepackage{booktabs}
\usepackage{multirow}
\usepackage{tcolorbox}
\usepackage{graphicx}
\usepackage{caption}
\usepackage{adjustbox}
\usepackage{tikz}
\usepackage{xcolor}
\usepackage{subcaption}
\usetikzlibrary{positioning, fit, backgrounds, shadows, calc, shapes.geometric}

\definecolor{darkgreen}{rgb}{0.0, 0.7, 0.0}
\definecolor{indigo}{RGB}{57, 73, 163}

\title{Teaching Small Language Models to Learn Logic through Meta-Learning}

\author{
 \textbf{Leonardo Bertolazzi\textsuperscript{1}},
 \textbf{Manuel Vargas Guzmán\textsuperscript{2}}, \\
 \textbf{Raffaella Bernardi\textsuperscript{3}},
 \textbf{Maciej Malicki\textsuperscript{2}},
 \textbf{Jakub Szymanik\textsuperscript{1}},
\\
\\
 \textsuperscript{1}University of Trento,
 \textsuperscript{2}University of Warsaw,
 \textsuperscript{3}Free University of Bozen-Bolzano
\\
 \small{
   \textbf{Correspondence:} \href{mailto:leonardo.bertolazzi@unitn.it}{leonardo.bertolazzi@unitn.it}
 }
}

\begin{document}
\maketitle

\begin{abstract}
Large language models (LLMs) are increasingly evaluated on reasoning tasks, yet their logical abilities remain contested. To address this, we study LLMs’ reasoning in a well-defined fragment of logic: syllogistic reasoning. We cast the problem as premise selection and construct controlled datasets to isolate logical competence. Beyond evaluation, an open challenge is enabling LLMs to acquire abstract inference patterns that generalize to novel structures. We propose to apply few-shot meta-learning to this domain, thereby encouraging models to extract rules across tasks rather than memorize patterns within tasks. Although meta-learning has been little explored in the context of logic learnability, our experiments show that it is effective: small models (1.5B–7B) fine-tuned with meta-learning demonstrate strong gains in generalization, with especially pronounced benefits in low-data regimes. These meta-learned models outperform GPT-4o and o3-mini on our syllogistic reasoning task.
\end{abstract}

\section{Introduction}
Historically, a central debate in AI has focused on whether neural networks, which are often characterized as associative statistical systems, can capture abstract rules and learn to apply them systematically~\citep{fodor1988connectionism, smolenski1990tensor, lake2018scan, hupkes2020comp, kim2020cogs, mccurdy2024survey}.

In recent years, this debate has resurfaced when investigating Large Language Models' (LLMs) reasoning capabilities~\citep{huang2023reasoning, mondorf2024reasoning, liu2025logical}. Their ability to perform rigorous logical reasoning is often contested, raising the question of whether they achieve true generalization or merely rely on memorized patterns from their training data~\citep{balloccu2024leak, singh2024datacontaminationllms}. Even with extensive scientific, mathematical, and programming data, LLMs still fail to generalize beyond seen inference patterns \citep{clark2020soft, saparov2023ood, mirzadeh2025gsmsymbolic, gulati2024putnamaxiom, huang2025math, shi2023gsm-ic, yoran2024retrievalcontext}. 

\begin{figure}[t]
\centering
\begin{minipage}{\textwidth}
    \input{tikz/schema}
\end{minipage}
\caption{\textbf{Overview of a ML episode}. Given a set of premises (the knowledge base, $\mathcal{KB}$), a set of task demonstrations (or Study Examples), and a Query Hypothesis $x^\mathrm{query}$ that is entailed from $\mathcal{KB}$, models must generate the \emph{minimal} subset of premises, the Query Premises $y^\mathrm{query}$, from which $x^\mathrm{query}$ can be derived. During each ML episode, by being trained on the Study Examples, models learn to extract the abstract logical patterns. The examples show how we frame syllogistic inferences as a premise selection task. The dataset is built with pseudwords, where here we have variables for space reasons.}
\label{fig:fig_1}
\end{figure}

In this context, few-shot meta-learning (ML) approaches~\citep{irie2024} have emerged as promising methods for inducing out-of-distribution (OOD) generalization and rapid domain adaptation in LLMs. Specifically, this class of methods has proven effective in few-shot task generalization~\citep{min-etal-2022-metaicl, chen-etal-2022-meta}, mitigating catastrophic forgetting~\citep{irie2025continual}, and inducing systematic generalization in rule induction tasks~\citep{lake-baroni2023}.

However, it remains unclear whether these findings would translate to the domain of deductive reasoning and logic, where models are tasked to learn and apply more complex rules than what has been studied in previous research. 
Can few-shot meta-learning (ML) enable LLMs to learn the abstract logical patterns underlying reasoning and generalization within a formal system?
To answer this foundational question, we propose, for the first time, to apply ML to enhance the ability of LLMs to learn logic within a rigorously defined and controlled fragment of natural language~\citep{pratt2004fragment}.
As illustrated in Figure~\ref{fig:fig_1}, we evaluate the effectiveness of ML using a logical reasoning task grounded in syllogistic logic \citep{smiley1973, guzman2024syllogistic} and frame it as a premise selection task. This is a minimal but still interesting and non-trivial fragment of logic consisting of a few clear logical rules. 
Each problem presents a knowledge base of atomic logical statements. Models are tasked with identifying the \emph{minimal} subset of premises that logically entail a given test hypothesis. This premise selection task captures a core aspect of deductive reasoning: determining which known facts are necessary and sufficient to justify a conclusion. 
We apply ML to small LMs from the Qwen-2.5 family \citep{qwen2025qwen2.5}, ranging from 1.5B to 7B parameters. Specifically, we assess the generalization capabilities induced by ML, such as systematically performing inferences over unseen sets of premises, as well as over more complex (longer) or simpler (shorter) sets of premises than those encountered during training.\footnote{Our code and data are available at: \url{https://github.com/leobertolazzi/meta-learning-logic.git}.}

Our main contributions are as follows: 
\begin{itemize}
    \item We introduce a new synthetic dataset based on syllogistic logic to directly study logical reasoning generalization in LLMs within a controlled setting.
    \item We show that ML enables small LMs to better generalize in OOD syllogistic reasoning problems with particularly strong performance in smaller models and low-data regimes.
    \item We demonstrate that small LMs fine-tuned with ML can outperform state-of-the-art LLMs such as GPT-4o and o3-mini, on our syllogistic premise selection task.
\end{itemize}  

\section{Background}

\subsection{Syllogistic Logic}
In our experiments, we focus on the syllogistic fragment of first-order logic. 
Originally, syllogisms have been studied by Aristotle as arguments composed of two premises and a conclusion, such as: “\textit{All dogs are mammals; some pets are not mammals; therefore, some pets are not dogs.}” This basic form can be extended to include inferences involving more than two premises (see \citealt{lukasiewicz1951, smiley1973}). 

\begin{figure}[t]
\centering
\includegraphics[width=0.60\linewidth]{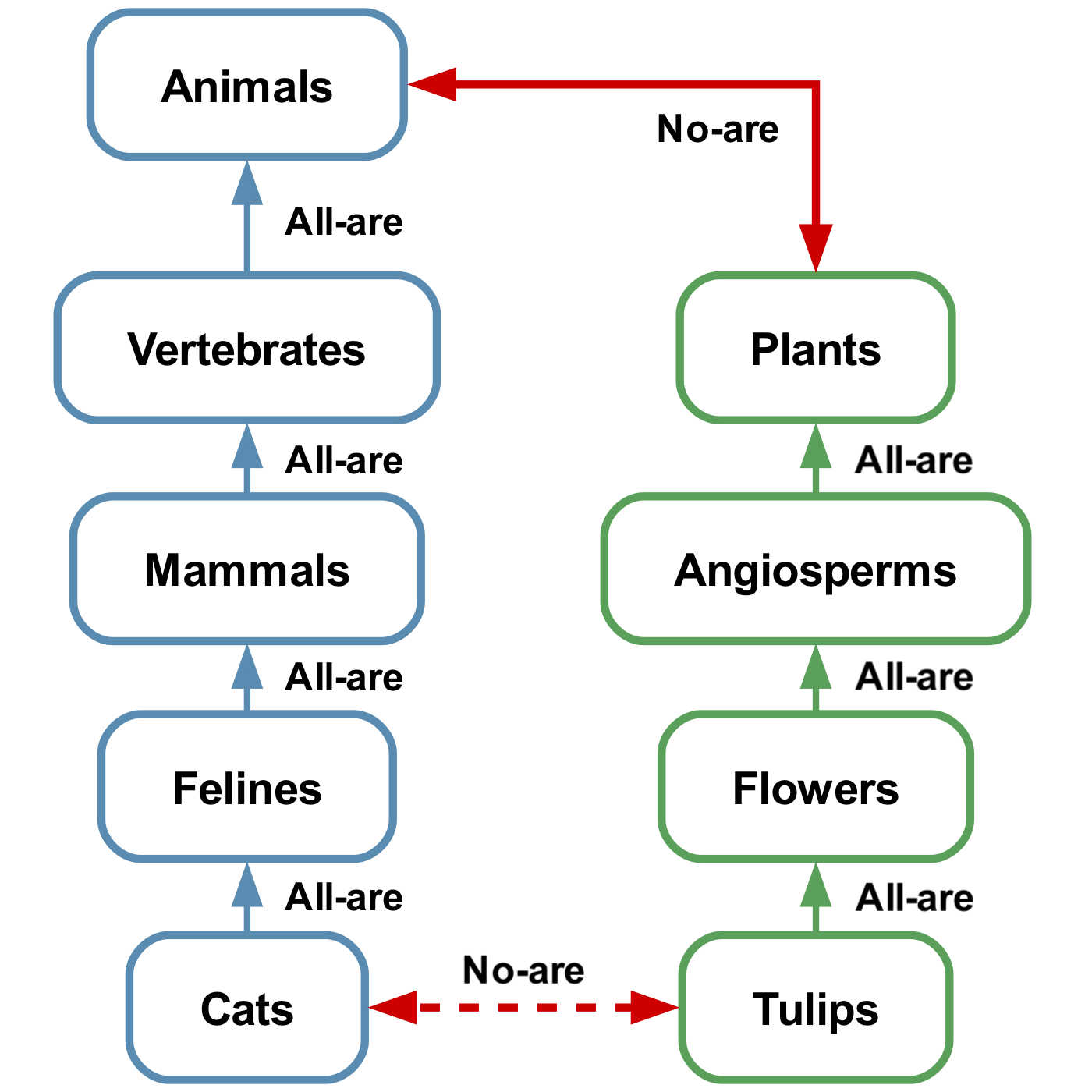}
\caption{\textbf{Example inference.} Edges labeled “All-are” denote universal affirmatives (e.g., \textit{All cats are felines}). The solid red edge is a universal negative (\textit{No animals are plants}). From these “\emph{atomic facts}” we infer \textit{No cats are tulips} (dashed red edge). Formally, this is expressed as $\{Aa - b, \; Ac - d, \; Ebd\} \vDash Eac$, the type 6 inference \citep{smiley1973}. Here we use words to better explain the inference, the syntetic dataset models see consist of pseudowords.}
\label{fig:inference}
\end{figure}

\paragraph{Syntax and semantics.}
The language of syllogistic logic comprises a finite set of atomic terms $\{ a, b, c, \ldots \}$ and four quantifier labels $A, E, I$, and $O$.
Well-formed formulas consists of $Aab$ (``All $a$ are $b$''), $Eab$ (``No $a$ are $b$''), $Iab$ (``Some $a$ are $b$''), and $Oab$ (``Some $a$ are not $b$'').
Finally, an \emph{A-chain}, denoted as $Aa-b$, represents the single formula $Aab$ or a sequence of formulas $Aac_1$, $Ac_1c_2$, $\dots$, $Ac_{n-1}c_n$, $Ac_nb$ for $n \ge 1$. 
A \emph{knowledge base} ($\mathcal{KB}$) is defined as a finite set of formulas (premises). 

An \emph{inference} $\mathcal{F} \vDash F$ (i.e., deriving a conclusion from a set of premises) holds when the conclusion $F$ is true in every interpretation
(an assignment of non-empty sets to terms) where all formulas in $\mathcal{F}$ are true. 
A set of formulas is \emph{consistent} if there exists at least one interpretation in which all formulas are simultaneously true. 

\paragraph{Minimal inferences.}
We aim for models to identify the minimal set of premises in a knowledge base to derive a given hypothesis.
Formally, we are interested in inferences $\mathcal{F} \vDash F$ such that $\mathcal{F}' \not\vDash F$ for any proper subset $\mathcal{F}' \subsetneq \mathcal{F}$. For example, $\{Abc, Abd\} \vDash Icd$ is minimal, while $\{Aab, Abc, Abd\} \vDash Icd$ is not because $Aab$ is not needed to infer the conclusion. 

There are seven types of minimal syllogistic inferences. To facilitate understanding, Figure~\ref{fig:inference} provides an intuitive representation of a type 6 inference. Further details about syllogistic logic can be found in Appendix~\ref{app:semantics}.

\subsection{Meta-learning in Autoregressive Models}
Meta-learning, or ``learning to learn'', is a paradigm that aims to enable machine learning models to acquire transferable knowledge across multiple tasks, allowing rapid adaptation to new tasks with minimal data. 
Among the numerous existing meta-learning frameworks \citep{hospedales2022meta}, our method is mainly inspired by Meta-learning Sequence Learners (MSL) \citep{irie2024}.

\paragraph{Data organization.} In standard supervised learning, data consists of a static dataset $\mathcal{D}_{\mathrm{train}} = \{(x_i,y_i)\}_{i=1}^N$ where inputs $x_i$ are mapped to targets $y_i$ under a fixed distribution $p(x,y)$. By contrast, meta-learning organizes data into  episodes $\mathcal{T} = (S^{\mathrm{supp}}, S^{\mathrm{query}})$ drawn from $p(\mathcal{T})$, where $S^{\mathrm{supp}} = \{(x_i,y_i)\}_{i=1}^K$ is the \textit{support set} containing task demonstrations, or study examples, and $S^{\mathrm{query}} = \{(x_j,y_j)\}_{j=1}^M$ is the \textit{query set} for evaluation. We consider the simplest scenario where $|S^{\mathrm{query}}| = 1$, containing a single example $(x^{\mathrm{query}}, y^{\mathrm{query}})$. We adapt this episodic formulation to our task, as shown in Figure~\ref{fig:fig_1}.

\paragraph{Optimization.} The fundamental difference between the two paradigms appears in their optimization objectives. Standard supervised learning finds parameters $\theta^*$ that maximize the likelihood: 
$\theta^* = \mathrm{argmax}_\theta\sum_{(x, y) \in \mathcal{D}_{\mathrm{train}}} \log p_\theta(y \mid x)$.
\noindent while meta-learning finds parameters $\theta^*$ that maximize the expected likelihood across tasks:
$ \theta^* = \mathrm{argmax}_\theta \quad \mathbb{E}_{\mathcal{T}} \left[ \log p_\theta(y^{\mathrm{query}} \mid x^{\mathrm{query}}, S^{\mathrm{supp}}) \right]$.

For autoregressive models, the probability $p_\theta(y^{\mathrm{query}} \mid x^{\mathrm{query}}, S^{\mathrm{supp}})$ is computed by conditioning on the support set $S^{\mathrm{supp}}$ as part of the input context, formatted as a sequence of input-output pairs preceding the query. This approach forces the model to develop the capabilities of recognizing and applying task patterns from the support examples to generate appropriate query outputs, namely, in our case, the minimal set of premises.

\section{Method}
\label{sec:method}

\subsection{Data Generation}
\label{sec:data}

In this section, we describe the methodology employed to construct textual datasets designed for the task of logical premise selection. 
The process begins with the random generation of graph-like structures representing $\mathcal{KB}s$. 
These are then translated into text using fixed syntactic templates and assigning pseudowords to nodes. 

\paragraph{Abstract representation.} 
To avoid ambiguity in premise selection, we use only \emph{non-redundant} $\mathcal{KB}s$, where for each derivable hypothesis $F$, there is a unique $\mathcal{F} \subseteq \mathcal{KB}$ such that $\mathcal{F} \vDash F$ is minimal.
We represent $\mathcal{KB}s$ as graphs, with constants as nodes and quantifiers as edges.\footnote{A visual representation of $\mathcal{KB}s$ and the seven types of inferences as graphs can be seen in Appendix \ref{app:kb_visualization}.} Synthetic $\mathcal{KB}s$ are generated by constructing such graphs. 
To ensure non-redundancy, $A$-formulas form disjoint subgraphs with at most one path between any two nodes. We created three independent sets of consistent $\mathcal{KB}s$ for training, validation, and testing to ensure diversity across splits.\footnote{See Appendix \ref{app:kb_generation} for the exact algorithms used to generate $\mathcal{KB}$s}

\paragraph{Textual translation.} To translate a given $\mathcal{KB}_i$ into a textual string, we: (1) assign a unique identifier $x_1, \ldots, x_n$ to each node within $\mathcal{KB}_i$; (2) map each edge to a fixed template connecting nodes $x_i$ and $x_j$ based on the quantifier represented by the edge (e.g., $Ax_ix_j$ becomes “All $x_i$ are $x_j$”); and (3) assign each unique node identifier $x_1, \ldots, x_n$ to a random English-like pseudoword (e.g., $x_1$ = wug, $x_2$ = blump).\footnote{Further details on the vocabulary of pseudowords we used are provided in Appendix \ref{app:vocabulary}.}

As illustrated in Figure~\ref{fig:fig_1}, we structured each datapoint in the three parts to begin with the token ``\texttt{knowledge base:}'', followed by the full sequence of premises, separated by commas. This is immediately followed by the special tag \texttt{<QUERY>}, and then the token ``\texttt{hypothesis:}'', which introduces the target hypothesis. Next comes the token ``\texttt{premises:}'', followed by the specific comma-separated premises that entail the hypothesis. To increase variability, we applied ten random pseudoword assignments and three random permutations of premise order for each $\mathcal{KB}$, resulting in multiple variants per datapoint.

Within each $\mathcal{KB}$, valid hypotheses can be inferred by minimal sets of premises of varying lengths.
We define the \textbf{length} of a inference as the total length of all $A$-chains it contains, which corresponds to the total number of $A$-formulas among its premises. For a given inference type $t$, we denote the maximum and minimum lengths as $\mu(t)$ and $\sigma(t)$, respectively.

We generated enough $\mathcal{KB}$s to obtain 1000 training, 5 validation, and 100 test examples for each inference type and length combination in the range from 0 to 19.\footnote{Note that some inference types (e.g., type 3) span the full range of lengths from 0 to 19, while others span only a subrange (e.g., type 2 spans from 1 to 10). See all type-length combinations within the generated $\mathcal{KB}$s in Figure \ref{fig:data-stats} in Appendix \ref{app:data_stats}.} This range was chosen to allow experiments with generalization to both unseen shorter and longer inferences. Full dataset statistics, including the number of generated $\mathcal{KB}$s per split, are reported in Appendix \ref{app:data_stats}.

\begin{figure*}[t!]
\centering
\input{tikz/generalization.tex}
\caption{\textbf{Length generalization}. We evaluate models on two types of length generalization: models trained on more complex (i.e., longer) inferences are tested on simpler (i.e., shorter) ones (Top) and vice versa (Bottom). The examples illustrate type 2 inferences.}
\label{fig:generalization}
\end{figure*}

\subsection{Meta-learning for Logic}
\label{sec:mind}

When applying meta-learning principles to the framework of syllogistic logic, we conceptualize the premises within a $\mathcal{KB}$ as \emph{atomic facts}. The seven types of syllogism (as detailed in Table~\ref{tab:inferences}, in the Appendix) are treated as \emph{arguments}, constructed using these atomic facts, and the model's task is to extract the minimal set of facts within a $\mathcal{KB}$ to produce a valid argument that proves the query hypothesis.

The type of systematic generalization ML addresses involves applying the seven fixed syllogistic inferences to new, unseen sets of atomic facts. This is central to logical reasoning because logical rules are, by definition, formal: conclusions follow from premises based solely on the \emph{structure} of the arguments, regardless of their specific content. Thus, successfully applying an inference type to a novel, unseen $\mathcal{KB}$ requires the model to recognize and instantiate the same formal structure with different premises. This generalization also includes variations in the number of atomic facts needed to instantiate an argument. Specifically, handling $A$-chains of varying lengths requires applying the learned inference patterns to longer or shorter instances of the same formal type.

\paragraph{Episodes organization.} To induce meta-learning of inference types, ML uses a set of episodes where each episode $\mathcal{T} =  (\mathcal{KB}, S^{\mathrm{supp}}, x^{\mathrm{query}}, y^{\mathrm{query}})$. Here, $\mathcal{KB}$ is a knowledge base, $S^{\mathrm{supp}}$ is a set of study valid hypothesis-premises pairs, $x^{\mathrm{query}}$ is a valid query hypothesis, and $y^{\mathrm{query}}$ is the minimal set of premises, in the KB, entailing $x^{\mathrm{query}}$. Figure \ref{fig:fig_1} shows a full ML episode using indexed variables in place of pseudowords for improved readability. Importantly, we consider study examples with inferences of the same type as the query. The number of study examples we set, i.e. valid hypothesis–premise pairs, is three. 
In their textual translation, we add the special tags \texttt{<STUDY>} to indicate the beginning of the sequence of study examples. During ML fine-tuning, models are trained to minimize the cross-entropy loss of the tokens in $y^{\mathrm{query}}$ given the input tokens from the context $(\mathcal{KB}, S^{\mathrm{supp}}, x^{\mathrm{query}})$. 
At inference time, ML models are provided as context $(\mathcal{KB}, S^{\mathrm{supp}}, x^{\mathrm{query}})$, and are tasked to generate $y^{\mathrm{query}}$.

\paragraph{Baseline.} Similarly to \citet{lake-baroni2023}, we consider a baseline where models are fine-tuned only on single input-output pairs $(x^{\mathrm{query}}, y^{\mathrm{query}})$ preceded by a $\mathcal{KB}$. The baseline is fine-tuned to minimize the cross-entropy loss of the tokens in $y^{\mathrm{query}}$ given the input tokens from the context $(\mathcal{KB}, x^{\mathrm{query}})$. At inference time, we tested the baseline both with and without study examples from $S^{\mathrm{supp}}$ in the context. Across all experiments, we report the results for whichever setting performed best, with a direct comparison of both configurations provided in Appendix~\ref{app:baseline-comparison}. In the tables, we use subscript notation (e.g., Baseline$_{\mathrm{+S}}$ or Baseline$_{\mathrm{-S}}$) to indicate whether the baseline was tested with or without study examples, respectively. To ensure a fair comparison between the meta-learning model and the baseline, we ensured that both models were fine-tuned on the exact same aggregate set of unique hypothesis-premises pairs. Specifically, the baseline was fine-tuned using a set $\mathcal{D}_{\text{baseline}}$ consisting of $(x^{\mathrm{query}}, y^{\mathrm{query}})$ unique pairs. For the meta-learning approach, the corresponding set of all unique hypothesis-premises pairs encountered across all $N$ episodes $\mathcal{T}_i = (\mathcal{KB}_i, S^{\mathrm{supp}}_i, x^{\mathrm{query}}_i, y^{\mathrm{query}}_i)$ is given by $\mathcal{D}_{\text{meta}} = \bigcup_{i=1}^N ( S^{\mathrm{supp}}_i \cup \{ (x^{\mathrm{query}}_i, y^{\mathrm{query}}_i) \} )$. We verified that $\mathcal{D}_{\text{baseline}} = \mathcal{D}_{\text{meta}}$. Moreover, since the meta-learning model processes more hypothesis-premises pairs within each training episode (due to $S^{\mathrm{supp}}_i$), we counterbalanced this by training the baseline model for a proportionally larger number of epochs.\footnote{Further details on number of epochs for each approach are provided in Appendix \ref{app:training-details}.}

\paragraph{Practical considerations.} ML training episodes involve longer input sequences, which increase memory demands; to accommodate this, we applied quantization to the ML model so that all models can be trained and evaluated on a single 80GB A100 GPU.\footnote{See Appendix~\ref{app:training-details} for full details on weight precisions used in the experiments.}

\section{Experimental Setup}

\subsection{Models and Evaluation}
We run experiments using the Qwen 2.5 family of decoder-only LLMs \citep{qwen2025qwen2.5}. More specifically, we test three sizes: 1.5B, 3B, and 7B parameters. This family of models is selected because it allows us to experiment with varying small sizes (ranging from 1.5 to 7 billion parameters) and achieves a better size vs. performance trade-off compared to other open weights model families.

In addition to the Qwen 2.5 family, we also evaluate the closed-source LLM GPT-4o \citep{gpt4o} and the Large Reasoning Model (LRM) o3-mini \citep{o3mini} on the logical premise selection task.\footnote{Note that LRMs are also LLMs, but post-trained to generate longer intermediate chains of thought, improving performance on complex reasoning tasks~\citep{xu2025lrmsurvey}.}
We evaluate these latter models both in a zero-shot setting and in a few-shot setting, using the $S^{\mathrm{supp}}$ study pairs as examples.\footnote{See the API details and the exact prompts used to evaluate closed models in Appendix \ref{app:closed_models}.}

\begin{table*}[t]
\centering
\begin{tabular}{c l l c c c}
\toprule
 & \textbf{Model} & \textbf{Method} & \textbf{All} & \textbf{Short} & \textbf{Long} \\
\midrule
\multirow{6}{*}{\rotatebox[origin=c]{90}{Fine-tuning}}
 & \multirow{2}{*}{Qwen-2.5 1.5B} 
              & ML           & 93.11 ± 0.61     & 94.28 ± 0.61     & 91.76 ± 0.27     \\
 &            & $\mathrm{Baseline}_{-S}$      & 85.56 ± 1.24     & 91.42 ± 0.82     & 80.56 ± 1.78     \\
\cmidrule(lr){2-6}
 & \multirow{2}{*}{Qwen-2.5 3B} 
              & ML           & 96.16 ± 0.44     & 96.24 ± 0.56     & 95.55 ± 0.43     \\
 &            & $\mathrm{Baseline}_{-S}$       & 93.03 ± 1.15     & 95.34 ± 1.18     & 90.92 ± 1.27     \\
\cmidrule(lr){2-6}
 & \multirow{2}{*}{Qwen-2.5 7B} 
              & ML           & 98.13 ± 0.98     & 98.26 ± 0.82     & 97.69 ± 1.40     \\
 &            & $\mathrm{Baseline}_{-S}$       & 95.76 ± 1.10     & 97.27 ± 1.22     & 94.13 ± 0.90     \\
\midrule
\multirow{4}{*}{\rotatebox[origin=c]{90}{Prompting}}
 & \multirow{2}{*}{GPT-4o} 
              & Few-shot       & 39.76            & 52.91            & 33.51            \\
 &            & Zero-shot         & 15.90            & 28.97            &  9.89            \\
\cmidrule(lr){2-6}
 & \multirow{2}{*}{o3-mini} 
              & Few-shot       & 88.45            & 87.91            & 88.51            \\
 &            & Zero-shot         & 67.98            & 73.29            & 64.54            \\
\bottomrule
\end{tabular}
\caption{\textbf{Core generalization.}  
Accuracy (mean ± std) on test inferences across all type‐length combinations (All), plus breakdown into the five shortest (Short) and longest (Long) inferences for each of the seven types of inference. Fine-tuned Qwen models use ML vs.\ Baseline; GPT-4o and o3-mini use few-shot vs.\ zero-shot prompting.}
\label{tab:core}
\end{table*}

We design experiments to evaluate the ability of ML to teach pretrained small LMs to systematically apply inferences to new, unseen sets of premises~\textemdash{}that is, to reason in a formal way by recognizing and instantiating the same underlying structure independently of the $\mathcal{KB}$s' content. 

\paragraph{Generalization.} In the first experiment, models are evaluated on their ability to generalize to unseen $\mathcal{KB}s$, while all inference lengths are seen. The training and testing sets contain inferences of all lengths for each of the seven types.
Since this is the simplest form of systematic application of syllogistic inference, we refer to it as \textbf{core generalization}. 

We then consider two more challenging generalizations involving inferences of \textbf{unseen length}. As illustrated in Figure~\ref{fig:generalization}, we examine the case of generalizing to \textbf{longer inferences} when the model has only learned from shorter ones (as studied in~\citealt{saparov2023ood}), and vice versa~\textemdash{}generalizing to \textbf{shorter inferences} after seeing only longer ones. In the logic literature, they are respectively known as recursiveness and compositionality~\citep{guzman2024syllogistic}.
To test this, we train exclusively on inferences whose lengths $x$ are $\sigma(t) \leq x \leq \mu(t) - 5$, and test on the five longest inferences for each type, i.e., those whose length is $\mu(t) - 5 < x \leq \mu(t)$. 
In the second case, we train on inferences with length $\sigma(t) + 5 \leq x \leq \mu(t)$, and test only on the five shortest inference lengths for each type, i.e., those with length $\sigma(t) \leq x < \sigma(t) + 5$. An intuitive representation of these generalizations is provided in Figure \ref{fig:generalization}. Notably, we consider two varying types of study examples $S^\mathrm{supp}$: the \textbf{aligned} and \textbf{disaligned} sets of study examples, in which each $(x^\mathrm{supp}, y^\mathrm{supp})$ either falls within or outside the range of inference lengths used for testing, respectively.\footnote{More precisely, \textbf{aligned} and \textbf{disaligned} depend on whether we are evaluating models on unseen shorter or longer inferences. For longer inferences, \textbf{disaligned} includes inferences with lengths $\sigma(t) \leq x \leq \mu(t) - 5$, and  \textbf{aligned} includes those with lengths $\mu(t) - 5 < x \leq \mu(t)$. For shorter ones, instead, \textbf{aligned} includes inferences with lengths $\sigma(t) \leq x < \sigma(t) + 5$, and  \textbf{disaligned} includes those with lengths $\sigma(t) + 5 \leq x \leq \mu(t)$.} 

Figure \ref{fig:data-stats}, in the Appendix, shows all inference type-length combinations within training and test split in the core and in the length generalization settings. 
These datasets contain 1,000 and 100 datapoints for each training and testing type–length combination, respectively. To further investigate the performance of ML in a \textbf{limited data regime}, we also consider the case where only 100 datapoints are available for each training type–length combination.

\subsection{Prediction Accuracy}
We consider a model prediction to be correct if the set of premises extracted from the generated text matches the ground truth set of \emph{minimal} premises. Using this criterion, we measure accuracy both aggregated, i.e., across an entire test set, and decomposed by each test type-length combination. All models (1.5B, 3B, and 7B) are fine-tuned three times and with different random seeds, thus we report mean and standard deviation of each accuracy. 

\section{Results}

\begin{table*}[ht]
\centering
\begin{tabular}{l l c c c c}
\toprule
\multirow{2}{*}{\textbf{Model}} & \multirow{2}{*}{\textbf{Method}} & \multicolumn{2}{c}{\textbf{Short $\rightarrow$ Long}} & \multicolumn{2}{c}{\textbf{Long $\rightarrow$ Short}} \\
\cmidrule(lr){3-4} \cmidrule(lr){5-6}
                               &                                & \textbf{Disaligned} & \textbf{Aligned} & \textbf{Disaligned} & \textbf{Aligned} \\
\midrule
\multirow{2}{*}{Qwen-2.5 1.5B} 
              & ML           & 76.42 ± 2.95 & 91.75 ± 1.10 & 70.94 ± 2.27 & 71.13 ± 1.83 \\
              & $\mathrm{Baseline}_{-S}$      & 63.53 ± 1.16 & 63.53 ± 1.16 & 56.67 ± 1.22 & 56.67 ± 1.22 \\
\cmidrule(lr){2-6}
\multirow{2}{*}{Qwen-2.5 3B} 
              & ML           & 87.61 ± 1.97 & 95.86 ± 0.70 & 77.19 ± 3.53 & 78.53 ± 1.71 \\
              & $\mathrm{Baseline}_{-S}$      & 76.78 ± 1.63 & 76.78 ± 1.63 & 71.88 ± 1.49 & 71.88 ± 1.49 \\
\cmidrule(lr){2-6}
\multirow{2}{*}{Qwen-2.5 7B} 
              & ML           & 90.03 ± 1.09 & 96.84 ± 0.15 & 76.23 ± 2.91 & 83.41 ± 1.63 \\
              & $\mathrm{Baseline}_{-S}$      & 80.76 ± 2.65 & 80.76 ± 2.65 & 71.08 ± 1.55 & 71.08 ± 1.55 \\
\bottomrule
\end{tabular}
\caption{\textbf{Generalization to unseen lengths.} Accuracy (mean ± std) of meta-learning and baseline models when trained on short inferences and tested on longer ones or vice versa.
In both cases, we compare the settings in which the inferences in the study examples either falls within (Aligned) or outside (Disaligned) the range of inference lengths used for testing. $\mathrm{Baseline}_{-S}$ models have no study examples, hence such difference does not hold for them.}
\label{tab:rec_comp_gen}
\end{table*}

\subsection{Core Generalization}
\label{sec:core}

We first examine the performance of meta-learning versus the baseline on core generalization (Table \ref{tab:core}), with models trained and tested on all inference type-length combinations. The ``Short'' and ``Long'' columns report aggregated accuracy on the sets of the five shortest and longest inferences, respectively, for each type. We hypothesize that longer inferences are harder because, to be correct, models must select \emph{all} premises belonging to a larger minimal set of premises.

Across all Qwen-2.5 model sizes (1.5B, 3B, 7B), the meta-learning approach consistently yields higher accuracy than the baseline. Performance improves with model scale in both approaches. For example, ML accuracy increases from 93.11\% (1.5B) to 98.13\% (7B) on all type‐length combinations, with accuracy on shortest inferences rising from 94.28\% to 98.26\%, and on the longest ones increasing from 91.76\% to 97.69\%. In contrast, baseline performance rises more slowly~\textemdash{}from 85.56\% (1.5B) to 95.76\% (7B)~\textemdash{}and shows a wider drop on the longest inferences, falling as low as 80.56\% for the smallest model. Notably, the performance gap between ML and the baseline narrows as model size increases, suggesting that larger models achieve better core generalization even without meta-learning. It is worth noting that with limited data, ML's advantage over the baseline becomes much wider at all sizes, as shown in Appendix \ref{app:limited-data}.

The closed-source models GPT-4o and o3-mini still underperform compared to Qwen-2.5 models fine-tuned with ML. Both models perform poorly in the zero-shot setting but improve with few-shot prompting: GPT-4o reaches 39.76\% on all type‐length combinations (with 52.91\% on shortest and 33.51\% on longest inferences), while o3-mini performs substantially better (88.45\% all combination, 87.91\% on shorters, and 88.51\% on longest). As expected, performance on the longest inferences is worse than that on the shortest ones for GPT-4o, while o3-mini maintains a more robust performance across inference lengths.

For completeness, we also include the results of Qwen-2.5 models in a few-shot prompting setting in Appendix \ref{app:prompt_model}. These results demonstrate that the base model, prior to fine-tuning, is indeed largely incapable of performing the task.

\subsection{Length Generalization}
Table \ref{tab:rec_comp_gen} shows that ML models consistently outperform baseline models in generalizing to both longer and shorter inferences than those seen during training. In core generalization, we observed that longer inferences are more challenging than shorter ones. Instead, in the case of unseen lengths, an interesting and somewhat counterintuitive pattern emerges: it is generally easier for models to generalize to longer inferences than to shorter ones. This is true across all model sizes and in both approaches; for instance, the largest model, Qwen-2.5 7B, achieved 90.03\% accuracy on longer inferences (disaligned) compared to 76.23\% on shorter ones (disaligned). 

Aligning study example lengths with the test condition (aligned) proves moderately to highly effective for unseen short inferences across all ML model sizes. For example, Qwen-2.5 1.5B improved from 76.42\% to 91.75\%, and Qwen-2.5 3B improved from 87.61\% to 95.86\%. For unseen long inferences, this alignment is moderately effective in larger models: Qwen-2.5 7B improved from 76.23\% to 83.41\%, while the 1.5B and 3B models showed smaller gains (70.94\% to 71.13\% and 77.19\% to 78.53\%, respectively). These results indicate that ML enables models in the aligned condition to exploit abstract patterns in the study examples (unseen inference lengths), allowing them to more effectively answer query hypotheses requiring length generalization.

Again, ML's better performance in length generalization is especially noticeable with limited training data, where the difference between ML and baseline models grows significantly (see Appendix \ref{app:limited-data} for more details).

\begin{table*}[t]
  \centering
  \begin{tabular}{ll|cc|cc|c}
    \toprule
     & \textbf{Method} & \textbf{NVM} [\%] & \textbf{Avg. NVM} & \textbf{MAP} [\%] & \textbf{Avg. MAP} & \textbf{HP} [\%]  \\
    \midrule
    \multirow{3}{*}{L $\rightarrow$ S} & ML (aligned) & 42.94 & 4.9  & 36.68 & 2.1  & 57.5 \\
                                     & ML (disaligned)  & 28.31 & 3.72 & 52.81 & 1.76 & 66.06 \\
                                     & $\mathrm{Baseline}_{-S}$ & 28.21 & 6.19 & 23.38 & 2.1  & 72.78 \\
                                     
    \midrule
    \multirow{3}{*}{S $\rightarrow$ L}    & ML (aligned) & 9.76  & 1.66 & 87.54 & 5.08 & 60.94 \\
                                     & ML (disaligned) & 14.14 & 6.14 & 81.82 & 3.65 & 35.35 \\
                                     & $\mathrm{Baseline}_{-S}$ & 3.87  & 2.36 & 89.79 & 6.66 & 66.9 \\
                                     
    \bottomrule
  \end{tabular}
  \caption{\textbf{Error analysis.} Error analysis comparing ML and baseline on long to short (L $\rightarrow$ S) and short to long (S $\rightarrow$ L) generalization. The table shows percentages and averages for non-minimal valid sets of premises (NVM) and missing necessary $A$ premises (MAP), and the percentage of hallucinated premises (HP).}
  \label{tab:error_analysis}
\end{table*}

\section{Error Analysis}

Beyond simply measuring the accuracy of ML and the baseline, we additionally focus on two main types of errors models make when evaluated on unseen lengths. First, among all errors, we consider the proportion of \emph{non-minimal valid set of premises} (NVM). This means that the correct minimal set was generated by the model, but together with unnecessary premises; for this case, we also measure how many unnecessary premises, on average, the models generate. Note that a very low number of unnecessary premises indicates that a model is indeed close to the correct solution, whereas including many unnecessary premises, with the extreme case of outputting the full knowledge base, would indicate that models learned to exploit a trivial solution instead of internalizing the abstract inference pattern beyond the length seen during training. Alternatively, models may fail to provide the complete $A$-chain within the correct minimal set of premises, meaning that at least one \emph{necessary $A$ premise is missing} (MAP); here, we also track the average number of missing necessary $A$-formulas in erroneous answers. NVM and MAP are mutually exclusive. Furthermore, we consider an additional type of error that can occur simultaneously with either NVM or MAP: models may \emph{hallucinate} premises~\textemdash{}referred to as \emph{hallucinated premises} (HP)~\textemdash{}and output a formula that is not contained in the $\mathcal{KB}$.

Table \ref{tab:error_analysis} presents the error analysis for Qwen-2.5 7B\footnote{Each model was fine-tuned three times with different random seeds, we selected the best model for each approach for this analysis.} on the challenging length generalization settings.\footnote{See Appendix \ref{app:additional_errors} for further error analysis results.} HP is a common error type across both settings (often $>$50\%). The baseline model has the highest HP rate in long to short (72.78\%), while ML models are generally better.

When generalizing to shorter inferences, a substantial portion of errors (28-43\%) are NVM, indicating models indeed find logical solutions but include unnecessary premises. In this context, a lower number of unnecessary premises is better, as it is closer to the minimal set. The baseline model adds the most unnecessary premises (6.19 average), compared to ML (disaligned) (4.9) and ML (aligned) (3.72). 

For generalizations to longer inferences, errors show different patterns, with few NVM errors (4-14\%) and predominantly MAP errors (81-90\%). The average number of missing premises is higher in short to long (3.65-6.66) than in long to short (1.76-2.1), suggesting models struggle to provide the complete set of premises when evaluated on longer inferences than seen during training. The baseline model struggles most with longer inferences, with a high MAP error rate (89.79\%) and a large number of missing premises (6.66) contributing to its lower accuracy compared to ML.

Overall, the fact that both ML models and the baseline have a higher percentage of NVM errors when generalizing to shorter inferences and, conversely, a higher percentage of MAP errors when generalizing to longer inferences shows that these models still exhibit some bias towards the lengths seen during training, reflected in the correlation between training length and error types at test time.

\section{Related Work}
\subsection{LLMs' Logical Capabilities}
 LLMs often struggle with OOD logical generalization \citep{clark2020soft,saparov2023ood,guzman2024syllogistic}, multi-step inference \citep{creswell2023selectioninference}, and consistency across formal reasoning patterns \citep{parmar2024logicbench, hong2024closer}. Neuro-symbolic methods address these gaps by integrating logic modules or symbolic solvers, improving both performance and interpretability \citep{pan2023logic, olausson2023linc, kambhampati2024llmmodulo}. In a different direction, LRMs have shown strong gains in reasoning and planning tasks \citep{xu2025lrmsurvey}. Our proposed meta-learning approach offers a complementary alternative by enabling LLMs to systematically learn logical relations without relying on symbolic modules.
 
 Many benchmarks, such as LogiQA \citep{liu2020logiqa}, ProofWriter \citep{tafjord2021proofwriter}, FOLIO \citep{han2024folio}, and ProntoQA \citep{saparov2023ood}, have been proposed to measure logical reasoning in LLMs. However, we relied on syllogistic logic since existing datasets either do not control for specific types of inferences or only include a limited set of very basic inference rules, e.g., modus ponens. Moreover, most of these datasets do not allow us to vary the complexity (length) of inferences or the set of premises forming knowledge bases from which inferences can be drawn.

\subsection{Meta-learning}
Meta-learning enables models to rapidly adapt to new tasks by leveraging prior experiences across tasks \citep{thrun1998meta, hospedales2022meta}. Foundational approaches include memory-augmented neural networks \citep{santoro2016momorymeta}, prototypical networks \citep{snell2017prototipical}, and model-agnostic meta-learning (MAML) \citep{finn2017meta}. In the context of LLMs, meta-learning has been explored through techniques such as meta-in-context learning \citep{codaforno2023meta}, in-context tuning \citep{chen-etal-2022-meta}, and MetaICL \citep{min-etal-2022-metaicl}, which either train for or exploit the in-context learning abilities of models to adapt to new tasks using few-shot examples. 
Our proposed method draws inspiration from the MSL framework \citep{irie2024}, which we adapt and extend to solve the logical premise selection task.
Within the meta-learning literature, \citet{lake-baroni2023} is most closely related to our work. While they explore how meta-learning induces systematic generalization by composing simple primitive functions with few arguments, we instead focus on fixed, more complex abstract rules rooted in syllogistic logic, featuring a varying number of premises from which an inference can be drawn.

\section{Conclusion}

In this work, we have shown that ML is a promising approach to enhance the logical capabilities of small LMs, at least with respect to the syllogistic fragment. We applied few-shot meta-learning as a fine-tuning approach to improve the deductive reasoning of models, explicitly targeting the logical premise selection task. Our results show that ML significantly enhances logical generalization compared to the baseline, especially in small-scale and low-data scenarios. Remarkably, our fine-tuned small models outperform state-of-the-art LLMs on our syllogistic reasoning task. This demonstrates the potential of ML to advance the development of more robust LLM reasoners.

In the future, we should investigate not only systematic generalization using fixed inference rules, as we have done here, but also extend our research to learning the composition of multiple logical inferences. This approach aligns with ideas proposed in other meta-learning research, such as Meta-Learning for Compositionality \citep{lake-baroni2023}. Additionally, we should examine increasingly complex formal fragments of language, where the interactions among various inference-building blocks and reasoning forms become more intricate, and assess the effectiveness of ML in helping LLMs to generalize in such contexts. Furthermore, having shown that in principle ML may enhance logical generalization, we want to next study the effectiveness of ML in more naturalistic or real-world reasoning scenarios.

\section{Limitations}

Despite demonstrating meaningful progress in enhancing the logical reasoning capabilities of language models through the ML approach, this study has several limitations that future research could address.

\paragraph{Model selection.} The evaluation primarily targets small to mid-sized language models (1.5B to 7B parameters). This focus leaves open the question of whether the observed improvements from ML generalize to larger-scale models.

\paragraph{Meta-learning trade-offs.} The gains in reasoning ability achieved by ML come with associated costs. The meta-learning strategy adopted involves incorporating multiple study examples into the input context during fine-tuning. This leads to longer input sequences, which in turn increase memory usage and computational demands compared to standard fine-tuning approaches.

\paragraph{Episode construction variability.} In this work, we fixed several meta-learning parameters that could be further explored in future studies. First, we set the number of study examples in each episode to three, primarily due to the memory constraints imposed by longer sequences in the ML setting. Second, our generalization experiments focused on length generalization, so the support set was restricted to inferences of the same type as the query. Future work could examine alternative support set configurations, for example investigating inter-inference transfer, that is, whether in-context knowledge of one inference type can be leveraged to solve queries of a different type.

\paragraph{Focus on a logic fragment.} This work is constrained to the syllogistic fragment of first-order logic. Future research should investigate whether our conclusions extend to more expressive logical systems or to real-world scenarios where reasoning tasks are less structured. The future work should also evaluate meta-learning trade-offs for such more complex logical systems. However, syllogistic logic is a restricted domain that allows for precise control over variables such as the type of inference considered, inference length, and the structure of knowledge bases. In the context of this study, it serves as a valuable testbed for investigating logical generalization in LLMs.

\section{Acknowledgement}
This work was partially funded by the National Science Centre, Poland under the OPUS call [grant 2020/37/B/HS1/04220]. We gratefully acknowledge Polish high-performance computing infrastructure PLGrid (HPC Center: ACK Cyfronet AGH) for providing computer facilities and support within computational grant no. PLG/2025/018321.

\bibliography{custom}

\appendix

\section{Formal Semantics and Syllogistic Inference Patterns}
\label{app:semantics}
In this section, we formally define the semantics of syllogistic logic by translating syllogistic formulas into first-order logic. 
We also specify a consistent set of such formulas and formalize a valid inference within this framework.
Let $\mathcal{A}=\{a,b,c, \ldots\}$ be a set of atomic terms, and let $\mathcal{R}=\{R,S,T, \ldots\}$ be a set of unary relational symbols. We bijectively assign to every atomic term $a \in \mathcal{A}$ a relational symbol $R_a \in \mathcal{R}$, and interpret syllogistic formulas as first-order logic sentences: $Aab$ as $\forall x\, [R_a(x) \to R_b(x)]$, $Eab$ as $\forall x\, [R_a(x) \to \neg R_b(x)]$, $Iab$ as $\exists x\, [R_a(x) \land R_b(x)]$, and $Oab$ as $\exists x\, [R_a(x) \land \neg R_b(x)]$. We say that a set $\mathcal{F}$ of syllogistic formulas is \emph{consistent} if there exists a structure $M$ in signature $\mathcal{R}$ such that every relation $R^M$ is non-empty, and the interpretation of every sentence in $\mathcal{F}$ holds in $M$, denoted by $M \vDash \mathcal{F}$. For a syllogistic formula $F$, the pair $(\mathcal{F},F)$ is an \emph{inference}, denoted by $\mathcal{F} \vDash F$, if $M \vDash \{F\}$, whenever $M \vDash \mathcal{F}$ for a structure $M$ in signature $\mathcal{R}$.

\section{Dataset}
\label{app:dataset}

\begin{table}[t!]
    \small
    \centering
    \begin{tabular}{cl}
        \toprule
        \textbf{Type} & \textbf{Inference} \\  
        \midrule
        1 & $\{Aa - b, Ac - d, Oad\} \vDash Obc$ \\  
        2 & $\{Aa - b\} \vDash Aab$ \\  
        3 & $\{Aa - b, Ac - d, Aa - e, Ede\} \vDash Obc$ \\  
        4 & $\{Aa - b, Aa - c\} \vDash Ibc$ \\  
        5 & $\{Aa - b, Ac - d, Ae - f, Iae, Edf\} \vDash Obc$ \\  
        6 & $\{Aa - b, Ac - d, Ebd\} \vDash Eac$ \\  
        7 & $\{Aa - b, Ac - d, Iac\} \vDash Ibd$ \\  
        \bottomrule
    \end{tabular}
    \caption{\textbf{Syllogistic inference types.} Each row shows a distinct logical inference pattern. Notation follows traditional categorical logic: $ Aab $ denotes a universal affirmative ("All $ a $ are $ b $"), $ Eab $ a universal negative ("No $ a $ are $ b $"), $ Iac $ a existential affirmative ("Some $ a $ are $ c $"), and $ Oad $ a existential negative ("Some $ a $ are not $ d $"). Formulas of the form $ Aa-b $ denote a sequence of $n$ $A$-formulas relating $ a $ and $ b $.}
    \label{tab:inferences}
\end{table}

\begin{table*}[ht]
    \centering
    \begin{tabular}{llccc}
    \toprule
    \textbf{Experiment} & \textbf{Split} & \textbf{Size} & \textbf{\# KBs} & \textbf{\# Premises (Min--Max)} \\
    \midrule
    \multirow{3}{*}{Core Generalization} 
        & Train      & 97,000 & 100 & 26--35 \\
        & Validation &   485  &  15 & 26--36 \\
        & Test       &  9,700 & 200 & 26--38 \\
    \midrule
    \multirow{3}{*}{Short $\rightarrow$ Long} 
        & Train      & 62,000 & 100 & 26--35 \\
        & Validation &   310  &  15 & 26--36 \\
        & Test       &  3,500 & 194 & 26--38 \\
    \midrule
    \multirow{3}{*}{Long $\rightarrow$ Short} 
        & Train      & 62,000 & 100 & 26--35 \\
        & Validation &   310  &  15 & 26--36 \\
        & Test       &  3,500 & 200 & 26--38 \\
    \bottomrule
    \end{tabular}
    \caption{\textbf{Dataset statistics across experiments.} For each experiment and split, the table reports the number of unique query hypothesis-premises pairs (Size), the number of $\mathcal{KB}$s from which the pairs are generated (\# KBs), and the range of total premises within $\mathcal{KB}$s (\# Premises). In the additional experiment with limited training data, the total training size is reduced by a factor of ten.}
    \label{tab:more-data-stats}
\end{table*}

\subsection{$\mathcal{KB}$s' Generation}
\label{app:kb_generation}
Knowledge bases can be modeled as edge-labeled graphs, in which nodes correspond to atomic terms and edges are labeled with quantifiers. 
Our graph generation algorithm comprises two principal stages: (1) We first construct all \emph{A-chains} of the knowledge base, 
which is used as its structural backbone, by generating disjoint trees---directed acyclic graphs that ensure a unique path exists between any pair of nodes. 
(2) Subsequently, we incorporate additional label edges corresponding to $E$, $I$, and $O$ formulas, while maintaining the overall consistency of the knowledge base.

To construct all possible valid syllogisms from each artificially generated knowledge base, we employ antillogisms---minimal inconsistent set of syllogistic formulas.
For example, consider the set $\{Aab, Aac, Ebc\}$, which forms an antilogism. By negating the inconsistent formula $Ebc$, we obtain a valid inference 
in which the remaining formulas $\{Aab, Aac\}$ entail its negation, i.e., $\{Aab, Aac\} \vDash Ibc$. This corresponds to an inference of type 4. 
More generally, any syllogism can be derived from an antilogism of the form $\mathcal{F} \cup \{\neg F\}$ by inferring the conclusion $F$ from the consistent set $\mathcal{F}$, 
that is, $\mathcal{F} \vDash F$. This result was formally established by \citep{smiley1973}, who also demonstrated that there exist only three distinct types of antilogisms. 
Furthermore, as shown by \citep{guzman2024syllogistic}, all valid syllogistic inferences can be systematically derived from these three canonical forms of antilogism 
(see Table \ref{tab:inferences}).

\subsection{$\mathcal{KB}$s' Visualization}
\label{app:kb_visualization}
To provide an intuitive understanding of the various types of inferences and their derivation from the knowledge bases employed in our framework, 
we represent syllogistic formulas as graphs. These graphs encompass the knowledge base, the corresponding hypothesis, 
and the minimal inference---defined as the smallest subset of premises required to derive the hypothesis.

\begin{figure*}[htbp]
  \centering

  % Titles above images
  \begin{minipage}{\textwidth}
    \centering
    \textbf{KB with Query Hypothesis and Type 1 Inference:} \\
    \vspace{1em}
    \includegraphics[width=0.8\textwidth]{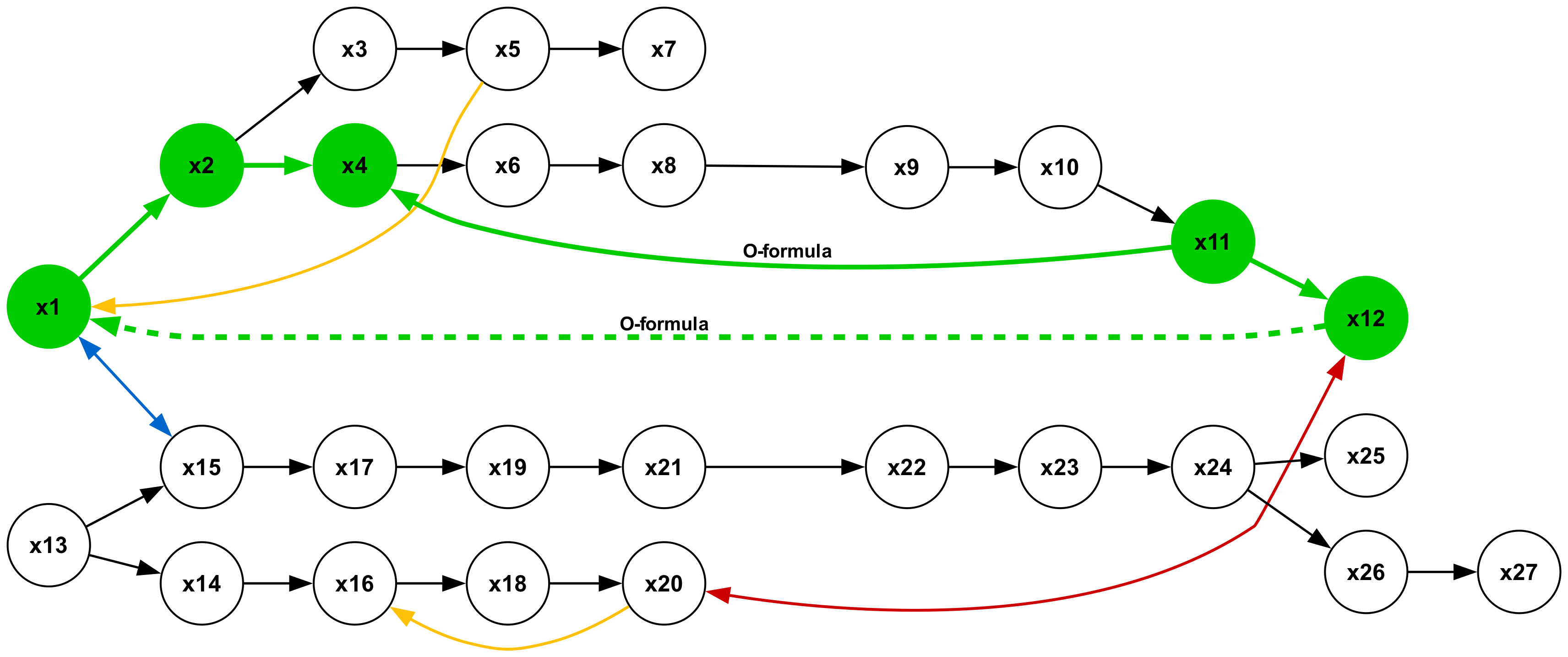}
  \end{minipage}

  % Linguistic description
  \vspace{1em}
  \textbf{Textual Translation:}
  \begin{tcolorbox}[
      colback=lightgray!20,
      fonttitle=\bfseries,
      fontupper=\small\ttfamily,
      sharp corners=southwest,
      rounded corners,
      width=\textwidth,
      boxrule=0pt,
      toptitle=0.1mm,
      bottomtitle=0.1mm
    ]
    \textbf{knowledge base:} 
    \textbf{\textcolor{darkgreen}{All x1 are x2}},
    All x2 are x3,
    \textbf{\textcolor{darkgreen}{All x2 are x4}},
    All x3 are x5,
    All x4 are x6,
    All x5 are x7,
    All x6 are x8, 
    All x8 are x9, 
    All x9 are x10,
    All x10 are x11,
    \textbf{\textcolor{darkgreen}{All x11 are x12}},
    All x13 are x14,
    All x13 are x15,
    All x14 are x16,
    All x15 are x17,
    All x16 are x18,
    All x17 are x19,
    All x18 are x20,
    All x19 are x21,
    All x21 are x22,
    All x22 are x23,
    All x23 are x24,
    All x24 are x25,
    All x24 are x26,
    All x26 are x27,
    No x20 are x12,
    Some x15 are x1,
    \textbf{\textcolor{darkgreen}{Some x11 are not x4}},
    Some x5 are not x1,
    Some x20 are not x16 \\[1pt]
    
    \textbf{hypothesis:} Some x12 are not x1 \\[1pt]
    
    \textbf{premises:} 
    \textbf{\textcolor{darkgreen}{All x1 are x2}},
    \textbf{\textcolor{darkgreen}{All x2 are x4}}, 
    \textbf{\textcolor{darkgreen}{All x11 are x12}},
    \textbf{\textcolor{darkgreen}{Some x11 are not x4}}
  \end{tcolorbox}
  
  \caption{\textbf{Type 1 syllogistic inference on graphs}. Visualization of a type 1 syllogistic inference using a graph representation of an example $\mathcal{KB}$, alongside the corresponding textual translation. In the graph (top), nodes represent predicates. Black edges indicate A-formulas (``All As are Bs''), blue edges indicate I-formulas (``Some As are Bs''), red edges indicate E-formulas (``No As are Bs''), and yellow edges indicate O-formulas (``Some As are not Bs''). The query hypothesis is represented by a dashed green edge, and the edges that prove the hypothesis are highlighted in green. The text translation illustrates how the abstract graph representation is converted into a text format suitable for LLM processing by applying fixed templates that represent logical formulas.}

  \label{fig:type_1}
\end{figure*}

\begin{figure*}[htbp]
  \centering

  % Titles above images
  \begin{minipage}{\textwidth}
    \centering
    \textbf{KB with Query Hypothesis and Type 2 Inference:} \\
    \vspace{1em}
    \includegraphics[width=0.8\textwidth]{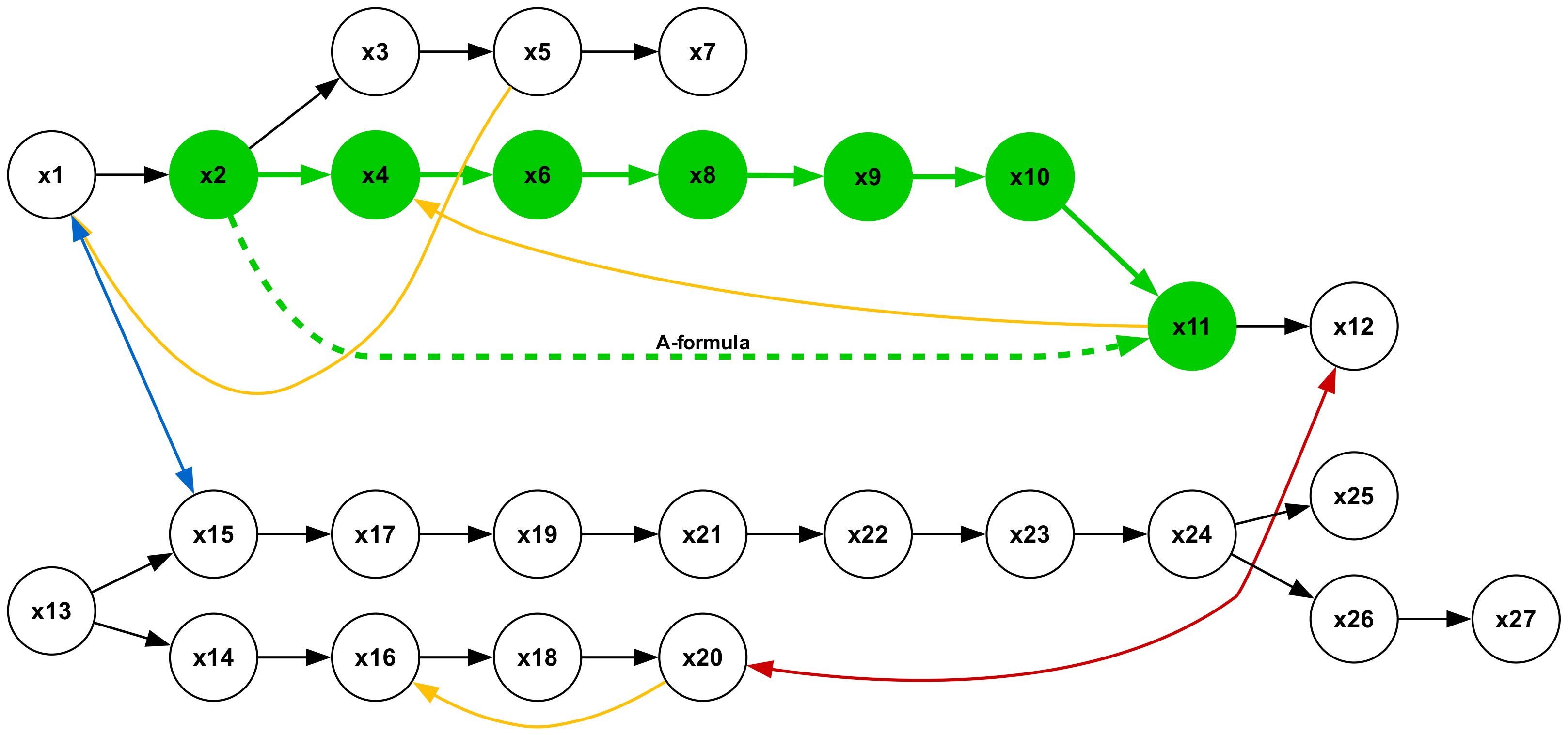}
  \end{minipage}

  % Linguistic description
  \vspace{1em}
  \textbf{Textual Translation:}
  \begin{tcolorbox}[
      colback=lightgray!20,
      fonttitle=\bfseries,
      fontupper=\small\ttfamily,
      sharp corners=southwest,
      rounded corners,
      width=\textwidth,
      boxrule=0pt,
      toptitle=0.1mm,
      bottomtitle=0.1mm
    ]
    \textbf{knowledge base:} 
    All x1 are x2,
    All x2 are x3,
    \textbf{\textcolor{darkgreen}{All x2 are x4}},
    All x3 are x5,
    \textbf{\textcolor{darkgreen}{All x4 are x6}},
    All x5 are x7,
    \textbf{\textcolor{darkgreen}{All x6 are x8}}, 
    \textbf{\textcolor{darkgreen}{All x8 are x9}}, 
    \textbf{\textcolor{darkgreen}{All x9 are x10}},
    \textbf{\textcolor{darkgreen}{All x10 are x11}},
    All x11 are x12,
    All x13 are x14,
    All x13 are x15,
    All x14 are x16,
    All x15 are x17,
    All x16 are x18,
    All x17 are x19,
    All x18 are x20,
    All x19 are x21,
    All x21 are x22,
    All x22 are x23,
    All x23 are x24,
    All x24 are x25,
    All x24 are x26,
    All x26 are x27,
    No x20 are x12,
    Some x15 are x1,
    Some x11 are not x4,
    Some x5 are not x1,
    Some x20 are not x16 \\[1pt]
    
    \textbf{hypothesis:} All x2 are x11 \\[1pt]
    
    \textbf{premises:} 
    \textbf{\textcolor{darkgreen}{All x2 are x4}}, 
    \textbf{\textcolor{darkgreen}{All x4 are x6}}, 
    \textbf{\textcolor{darkgreen}{All x6 are x8}}, 
    \textbf{\textcolor{darkgreen}{All x8 are x9}}, 
    \textbf{\textcolor{darkgreen}{All x9 are x10}}, 
    \textbf{\textcolor{darkgreen}{All x10 are x11}}
  \end{tcolorbox}
  
  \caption{\textbf{Type 2 syllogistic inference on graphs}. Visualization of a type 2 syllogistic inference using a graph representation of an example $\mathcal{KB}$, alongside the corresponding textual translation. In the graph (top), nodes represent predicates. Black edges indicate A-formulas (``All As are Bs''), blue edges indicate I-formulas (``Some As are Bs''), red edges indicate E-formulas (``No As are Bs''), and yellow edges indicate O-formulas (``Some As are not Bs''). The query hypothesis is represented by a dashed green edge, and the edges that prove the hypothesis are highlighted in green. The text translation illustrates how the abstract graph representation is converted into a text format suitable for LM processing by applying fixed templates that represent logical formulas.}

  \label{fig:task_visualization}
\end{figure*}

\begin{figure*}[htbp]
  \centering

  % Titles above images
  \begin{minipage}{\textwidth}
    \centering
    \textbf{KB with Query Hypothesis and Type 3 Inference:} \\
    \vspace{1em}
    \includegraphics[width=0.8\textwidth]{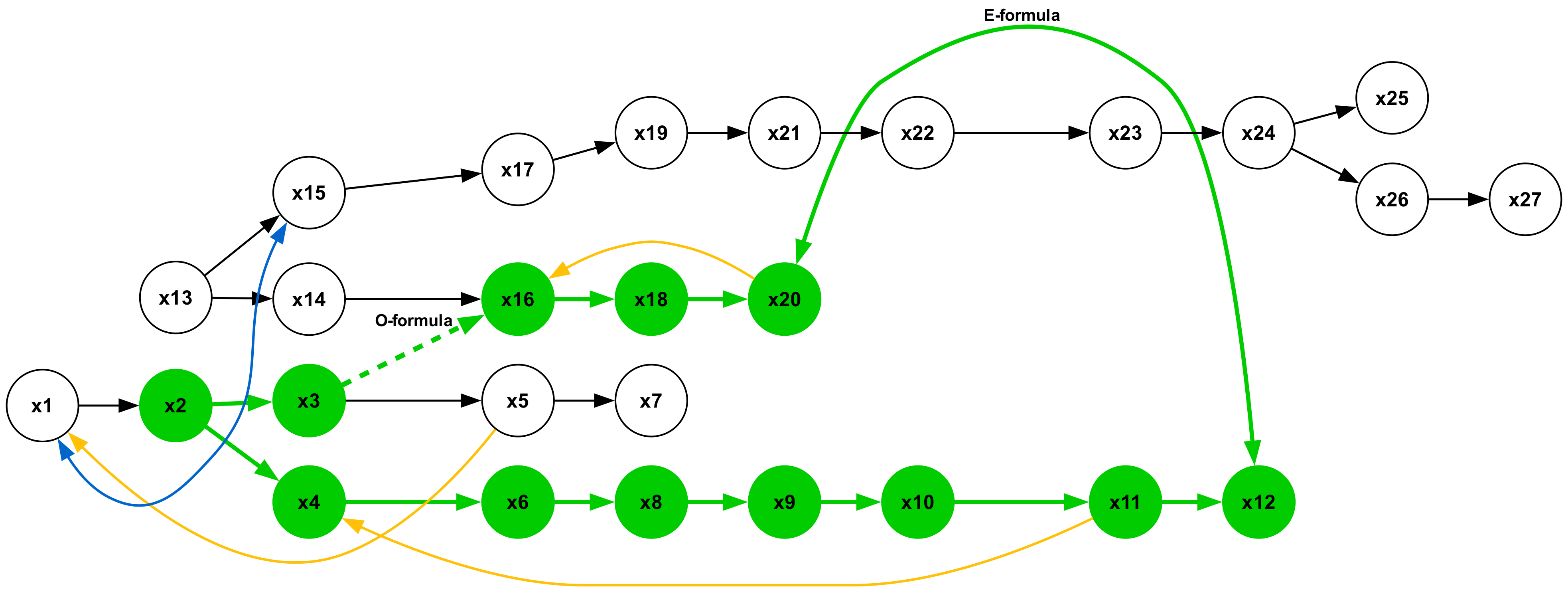}
  \end{minipage}

  % Linguistic description
  \vspace{1em}
  \textbf{Textual Translation:}
  \begin{tcolorbox}[
      colback=lightgray!20,
      fonttitle=\bfseries,
      fontupper=\small\ttfamily,
      sharp corners=southwest,
      rounded corners,
      width=\textwidth,
      boxrule=0pt,
      toptitle=0.1mm,
      bottomtitle=0.1mm
    ]
    \textbf{knowledge base:} 
    All x1 are x2,
    \textbf{\textcolor{darkgreen}{All x2 are x3}},
    \textbf{\textcolor{darkgreen}{All x2 are x4}},
    All x3 are x5,
    \textbf{\textcolor{darkgreen}{All x4 are x6}},
    All x5 are x7,
    \textbf{\textcolor{darkgreen}{All x6 are x8}}, 
    \textbf{\textcolor{darkgreen}{All x8 are x9}}, 
    \textbf{\textcolor{darkgreen}{All x9 are x10}},
    \textbf{\textcolor{darkgreen}{All x10 are x11}},
    \textbf{\textcolor{darkgreen}{All x11 are x12}},
    All x13 are x14,
    All x13 are x15,
    All x14 are x16,
    All x15 are x17,
    \textbf{\textcolor{darkgreen}{All x16 are x18}},
    All x17 are x19,
    \textbf{\textcolor{darkgreen}{All x18 are x20}},
    All x19 are x21,
    All x21 are x22,
    All x22 are x23,
    All x23 are x24,
    All x24 are x25,
    All x24 are x26,
    All x26 are x27,
    \textbf{\textcolor{darkgreen}{No x20 are x12}},
    Some x15 are x1,
    Some x11 are not x4,
    Some x5 are not x1,
    Some x20 are not x16 \\[1pt]
    
    \textbf{hypothesis:} Some x3 are not x16 \\[1pt]
    
    \textbf{premises:} 
    \textbf{\textcolor{darkgreen}{All x2 are x3}},
    \textbf{\textcolor{darkgreen}{All x2 are x4}}, 
    \textbf{\textcolor{darkgreen}{All x4 are x6}}, 
    \textbf{\textcolor{darkgreen}{All x6 are x8}}, 
    \textbf{\textcolor{darkgreen}{All x8 are x9}}, 
    \textbf{\textcolor{darkgreen}{All x9 are x10}}, 
    \textbf{\textcolor{darkgreen}{All x10 are x11}},
    \textbf{\textcolor{darkgreen}{All x11 are x12}},
    \textbf{\textcolor{darkgreen}{All x16 are x18}},
    \textbf{\textcolor{darkgreen}{All x18 are x20}},
    \textbf{\textcolor{darkgreen}{No x20 are x12}}
  \end{tcolorbox}
  
  \caption{\textbf{Type 3 syllogistic inference on graphs}. Visualization of a type 3 syllogistic inference using a graph representation of an example $\mathcal{KB}$, alongside the corresponding textual translation. In the graph (top), nodes represent predicates. Black edges indicate A-formulas (``All As are Bs''), blue edges indicate I-formulas (``Some As are Bs''), red edges indicate E-formulas (``No As are Bs''), and yellow edges indicate O-formulas (``Some As are not Bs''). The query hypothesis is represented by a dashed green edge, and the edges that prove the hypothesis are highlighted in green. The text translation illustrates how the abstract graph representation is converted into a text format suitable for LM processing by applying fixed templates that represent logical formulas.}

  \label{fig:type_3}
\end{figure*}

\begin{figure*}[htbp]
  \centering

  % Titles above images
  \begin{minipage}{\textwidth}
    \centering
    \textbf{KB with Query Hypothesis and Type 4 Inference:} \\
    \vspace{1em}
    \includegraphics[width=0.8\textwidth]{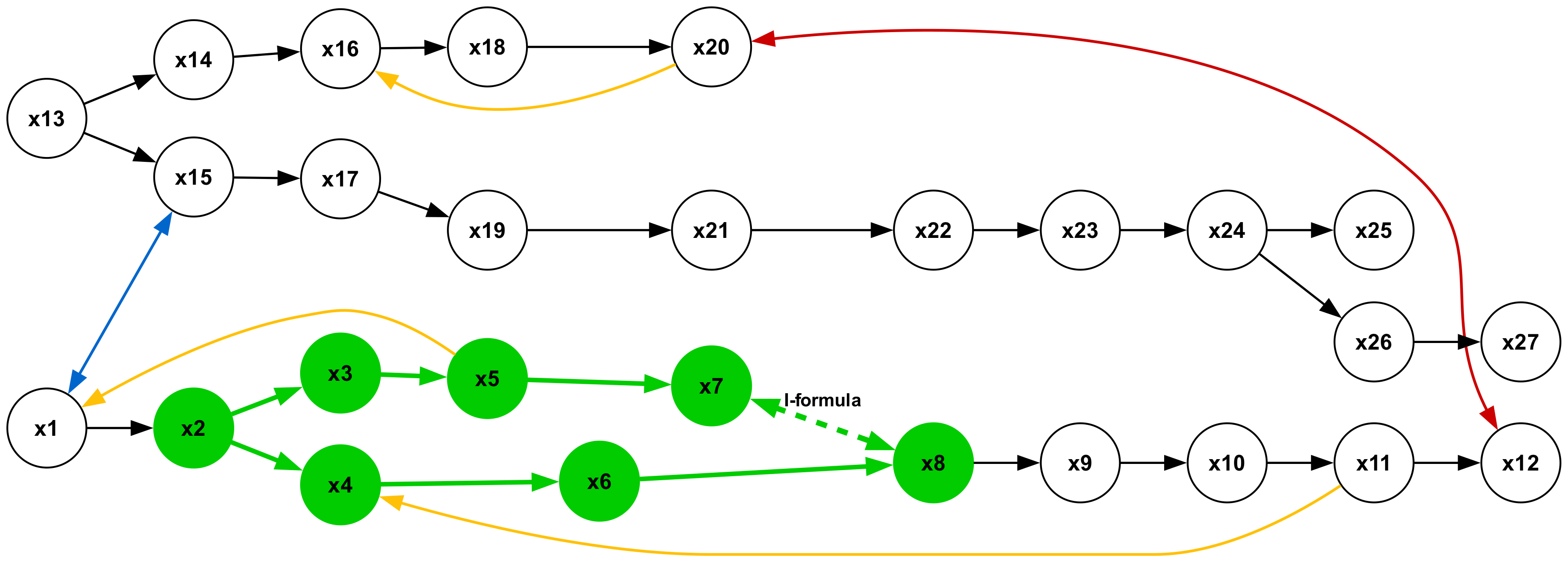}
  \end{minipage}

  % Linguistic description
  \vspace{1em}
  \textbf{Textual Translation:}
  \begin{tcolorbox}[
      colback=lightgray!20,
      fonttitle=\bfseries,
      fontupper=\small\ttfamily,
      sharp corners=southwest,
      rounded corners,
      width=\textwidth,
      boxrule=0pt,
      toptitle=0.1mm,
      bottomtitle=0.1mm
    ]
    \textbf{knowledge base:} 
    All x1 are x2,
    \textbf{\textcolor{darkgreen}{All x2 are x3}},
    \textbf{\textcolor{darkgreen}{All x2 are x4}},
    \textbf{\textcolor{darkgreen}{All x3 are x5}},
    \textbf{\textcolor{darkgreen}{All x4 are x6}},
    \textbf{\textcolor{darkgreen}{All x5 are x7}},
    \textbf{\textcolor{darkgreen}{All x6 are x8}}, 
    All x8 are x9, 
    All x9 are x10,
    All x10 are x11,
    All x11 are x12,
    All x13 are x14,
    All x13 are x15,
    All x14 are x16,
    All x15 are x17,
    All x16 are x18,
    All x17 are x19,
    All x18 are x20,
    All x19 are x21,
    All x21 are x22,
    All x22 are x23,
    All x23 are x24,
    All x24 are x25,
    All x24 are x26,
    All x26 are x27,
    No x20 are x12,
    Some x15 are x1,
    Some x11 are not x4,
    Some x5 are not x1,
    Some x20 are not x16 \\[1pt]
    
    \textbf{hypothesis:} Some x7 are x8 \\[1pt]
    
    \textbf{premises:} 
    \textbf{\textcolor{darkgreen}{All x2 are x4}}, 
    \textbf{\textcolor{darkgreen}{All x2 are x3}}, 
    \textbf{\textcolor{darkgreen}{All x4 are x6}}, 
    \textbf{\textcolor{darkgreen}{All x6 are x8}}, 
    \textbf{\textcolor{darkgreen}{All x3 are x5}},
    \textbf{\textcolor{darkgreen}{All x5 are x7}}
  \end{tcolorbox}
  
  \caption{\textbf{Type 4 syllogistic inference on graphs}. Visualization of a type 4 syllogistic inference using a graph representation of an example $\mathcal{KB}$, alongside the corresponding textual translation. In the graph (top), nodes represent predicates. Black edges indicate A-formulas (``All As are Bs''), blue edges indicate I-formulas (``Some As are Bs''), red edges indicate E-formulas (``No As are Bs''), and yellow edges indicate O-formulas (``Some As are not Bs''). The query hypothesis is represented by a dashed green edge, and the edges that prove the hypothesis are highlighted in green. The text translation illustrates how the abstract graph representation is converted into a text format suitable for LM processing by applying fixed templates that represent logical formulas.}

  \label{fig:type_4}
\end{figure*}

\begin{figure*}[htbp]
  \centering

  % Titles above images
  \begin{minipage}{\textwidth}
    \centering
    \textbf{KB with Query Hypothesis and Type 5 Inference:} \\
    \vspace{1em}
    \includegraphics[width=0.8\textwidth]{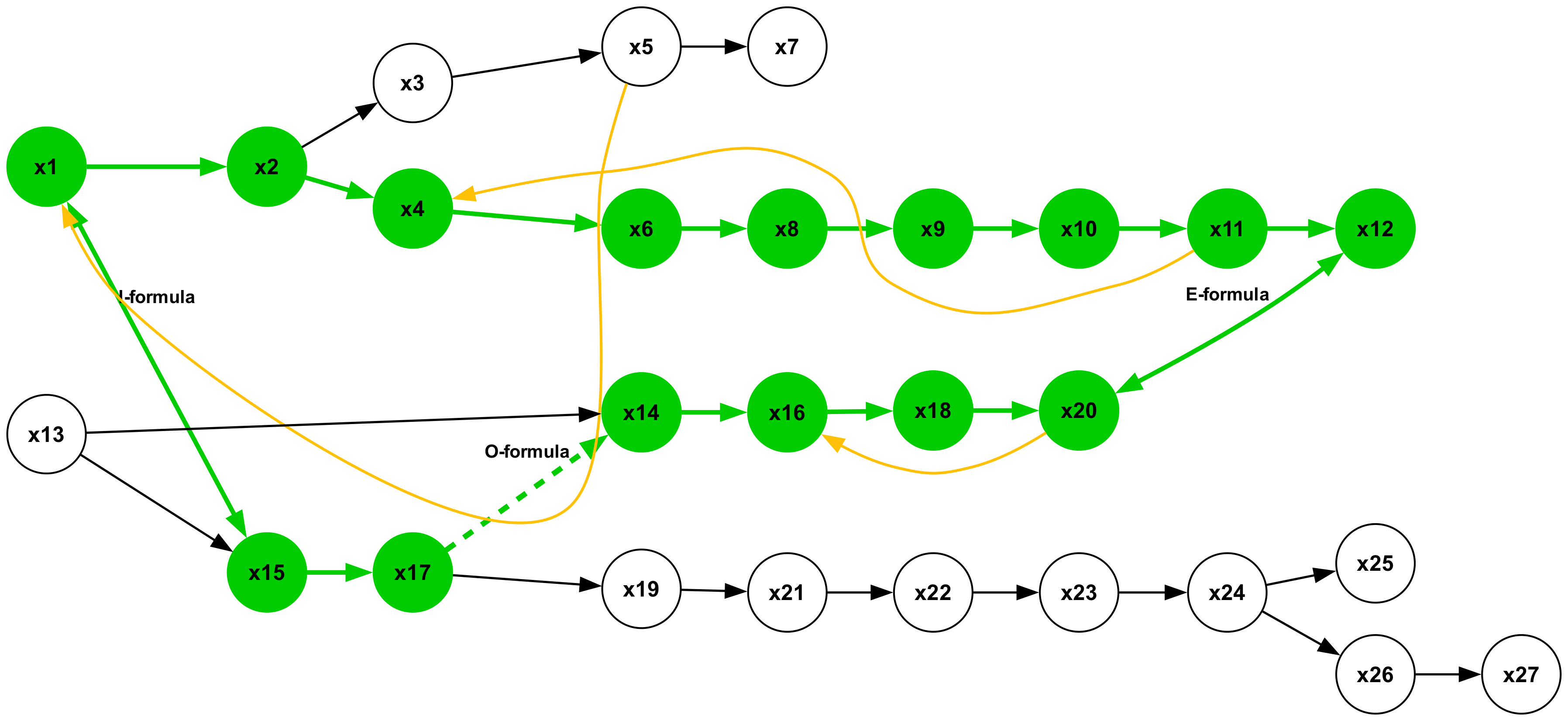}
  \end{minipage}

  % Linguistic description
  \vspace{1em}
  \textbf{Textual Translation:}
  \begin{tcolorbox}[
      colback=lightgray!20,
      fonttitle=\bfseries,
      fontupper=\small\ttfamily,
      sharp corners=southwest,
      rounded corners,
      width=\textwidth,
      boxrule=0pt,
      toptitle=0.1mm,
      bottomtitle=0.1mm
    ]
    \textbf{knowledge base:} 
    \textbf{\textcolor{darkgreen}{All x1 are x2}},
    All x2 are x3,
    \textbf{\textcolor{darkgreen}{All x2 are x4}},    
    \textbf{\textcolor{darkgreen}{All x4 are x6}},
    All x3 are x5,
    \textbf{\textcolor{darkgreen}{All x6 are x8}},
    All x5 are x7, 
    \textbf{\textcolor{darkgreen}{All x8 are x9}}, 
    \textbf{\textcolor{darkgreen}{All x9 are x10}},
    \textbf{\textcolor{darkgreen}{All x10 are x11}},
    \textbf{\textcolor{darkgreen}{All x11 are x12}},
    All x13 are x14,
    All x13 are x15,
    \textbf{\textcolor{darkgreen}{All x14 are x16}},
    \textbf{\textcolor{darkgreen}{All x15 are x17}},
    \textbf{\textcolor{darkgreen}{All x16 are x18}},
    All x17 are x19,
    \textbf{\textcolor{darkgreen}{All x18 are x20}},
    All x19 are x21,
    All x21 are x22,
    All x22 are x23,
    All x23 are x24,
    All x24 are x25,
    All x24 are x26,
    All x26 are x27,
    \textbf{\textcolor{darkgreen}{No x20 are x12}},
    \textbf{\textcolor{darkgreen}{Some x15 are x1}},
    Some x11 are not x4,
    Some x5 are not x1,
    Some x20 are not x16 \\[1pt]
    
    \textbf{hypothesis:} Some x17 are not x14 \\[1pt]
    
    \textbf{premises:} 
    \textbf{\textcolor{darkgreen}{All x1 are x2}},
    \textbf{\textcolor{darkgreen}{All x2 are x4}},
    \textbf{\textcolor{darkgreen}{All x4 are x6}},
    \textbf{\textcolor{darkgreen}{All x6 are x8}},
    \textbf{\textcolor{darkgreen}{All x8 are x9}}, 
    \textbf{\textcolor{darkgreen}{All x9 are x10}},
    \textbf{\textcolor{darkgreen}{All x10 are x11}},
    \textbf{\textcolor{darkgreen}{All x11 are x12}},
    \textbf{\textcolor{darkgreen}{All x14 are x16}},
    \textbf{\textcolor{darkgreen}{All x15 are x17}},
    \textbf{\textcolor{darkgreen}{All x16 are x18}},
    \textbf{\textcolor{darkgreen}{All x18 are x20}},    
    \textbf{\textcolor{darkgreen}{No x20 are x12}},
    \textbf{\textcolor{darkgreen}{Some x15 are x1}},
  \end{tcolorbox}
  
  \caption{\textbf{Type 5 syllogistic inference on graphs}. Visualization of a type 5 syllogistic inference using a graph representation of an example $\mathcal{KB}$, alongside the corresponding textual translation. In the graph (top), nodes represent predicates. Black edges indicate A-formulas (``All As are Bs''), blue edges indicate I-formulas (``Some As are Bs''), red edges indicate E-formulas (``No As are Bs''), and yellow edges indicate O-formulas (``Some As are not Bs''). The query hypothesis is represented by a dashed green edge, and the edges that prove the hypothesis are highlighted in green. The text translation illustrates how the abstract graph representation is converted into a text format suitable for LM processing by applying fixed templates that represent logical formulas.}

  \label{fig:type_5}
\end{figure*}

\begin{figure*}[htbp]
  \centering

  % Titles above images
  \begin{minipage}{\textwidth}
    \centering
    \textbf{KB with Query Hypothesis and Type 6 Inference:} \\
    \vspace{1em}
    \includegraphics[width=0.8\textwidth]{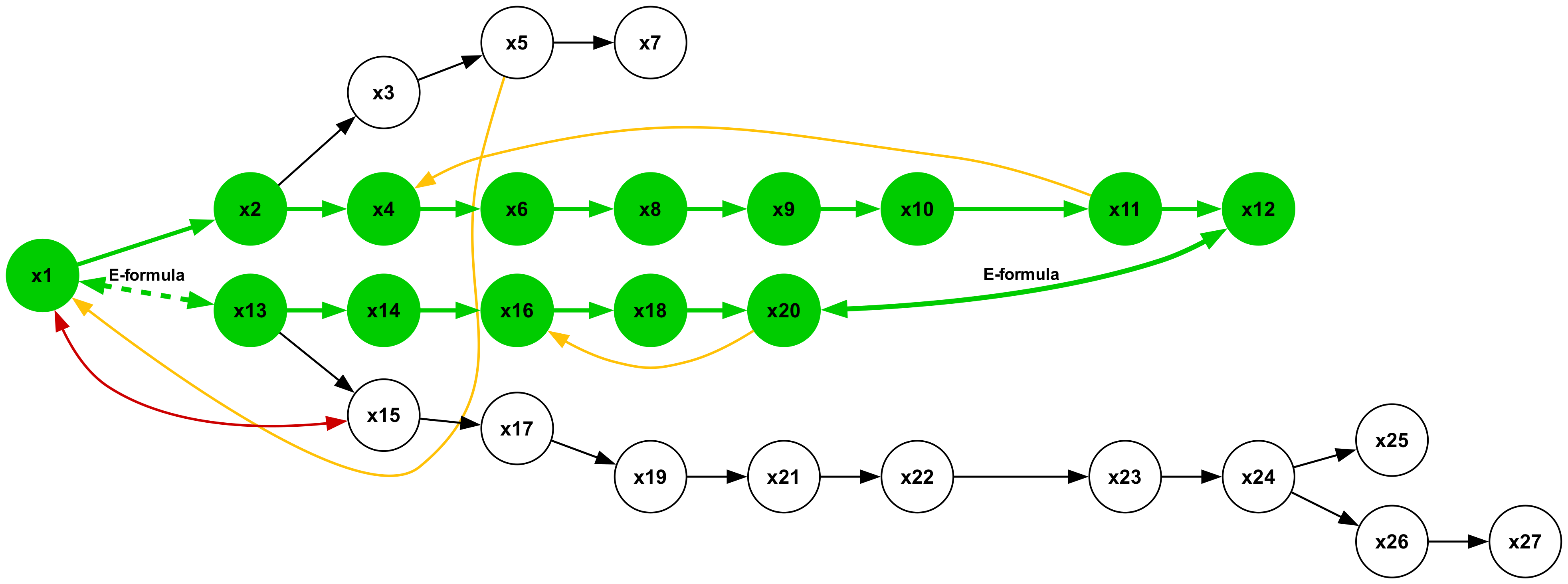}
  \end{minipage}

  % Linguistic description
  \vspace{1em}
  \textbf{Textual Translation:}
  \begin{tcolorbox}[
      colback=lightgray!20,
      fonttitle=\bfseries,
      fontupper=\small\ttfamily,
      sharp corners=southwest,
      rounded corners,
      width=\textwidth,
      boxrule=0pt,
      toptitle=0.1mm,
      bottomtitle=0.1mm
    ]
    \textbf{knowledge base:} 
    \textbf{\textcolor{darkgreen}{All x1 are x2}},
    All x2 are x3,
    \textbf{\textcolor{darkgreen}{All x2 are x4}},
    All x3 are x5,
    \textbf{\textcolor{darkgreen}{All x4 are x6}},
    All x5 are x7,
    \textbf{\textcolor{darkgreen}{All x6 are x8}}, 
    \textbf{\textcolor{darkgreen}{All x8 are x9}}, 
    \textbf{\textcolor{darkgreen}{All x9 are x10}},
    \textbf{\textcolor{darkgreen}{All x10 are x11}},
    \textbf{\textcolor{darkgreen}{All x11 are x12}},
    \textbf{\textcolor{darkgreen}{All x13 are x14}},
    All x13 are x15,
    \textbf{\textcolor{darkgreen}{All x14 are x16}},
    All x15 are x17,
    \textbf{\textcolor{darkgreen}{All x16 are x18}},
    All x17 are x19,
    \textbf{\textcolor{darkgreen}{All x18 are x20}},
    All x19 are x21,
    All x21 are x22,
    All x22 are x23,
    All x23 are x24,
    All x24 are x25,
    All x24 are x26,
    All x26 are x27,
    \textbf{\textcolor{darkgreen}{No x20 are x12}},
    Some x15 are x1,
    Some x11 are not x4,
    Some x5 are not x1,
    Some x20 are not x16 \\[1pt]
    
    \textbf{hypothesis:} No x1 are x13 \\[1pt]
    
    \textbf{premises:} 
    \textbf{\textcolor{darkgreen}{All x1 are x2}},
    \textbf{\textcolor{darkgreen}{All x2 are x4}},
    \textbf{\textcolor{darkgreen}{All x4 are x6}},
    \textbf{\textcolor{darkgreen}{All x6 are x8}}, 
    \textbf{\textcolor{darkgreen}{All x8 are x9}}, 
    \textbf{\textcolor{darkgreen}{All x9 are x10}},
    \textbf{\textcolor{darkgreen}{All x10 are x11}},
    \textbf{\textcolor{darkgreen}{All x11 are x12}},
    \textbf{\textcolor{darkgreen}{All x13 are x14}},
    \textbf{\textcolor{darkgreen}{All x14 are x16}},
    \textbf{\textcolor{darkgreen}{All x16 are x18}},
    \textbf{\textcolor{darkgreen}{All x18 are x20}},
    \textbf{\textcolor{darkgreen}{No x20 are x12}},
  \end{tcolorbox}
  
  \caption{\textbf{Type 6 syllogistic inference on graphs}. Visualization of a type 6 syllogistic inference using a graph representation of an example $\mathcal{KB}$, alongside the corresponding textual translation. In the graph (top), nodes represent predicates. Black edges indicate A-formulas (``All As are Bs''), blue edges indicate I-formulas (``Some As are Bs''), red edges indicate E-formulas (``No As are Bs''), and yellow edges indicate O-formulas (``Some As are not Bs''). The query hypothesis is represented by a dashed green edge, and the edges that prove the hypothesis are highlighted in green. The text translation illustrates how the abstract graph representation is converted into a text format suitable for LM processing by applying fixed templates that represent logical formulas.}

  \label{fig:type_6}
\end{figure*}

\begin{figure*}[htbp]
  \centering

  % Titles above images
  \begin{minipage}{\textwidth}
    \centering
    \textbf{KB with Query Hypothesis and Type 7 Inference:} \\
    \vspace{1em}
    \includegraphics[width=0.8\textwidth]{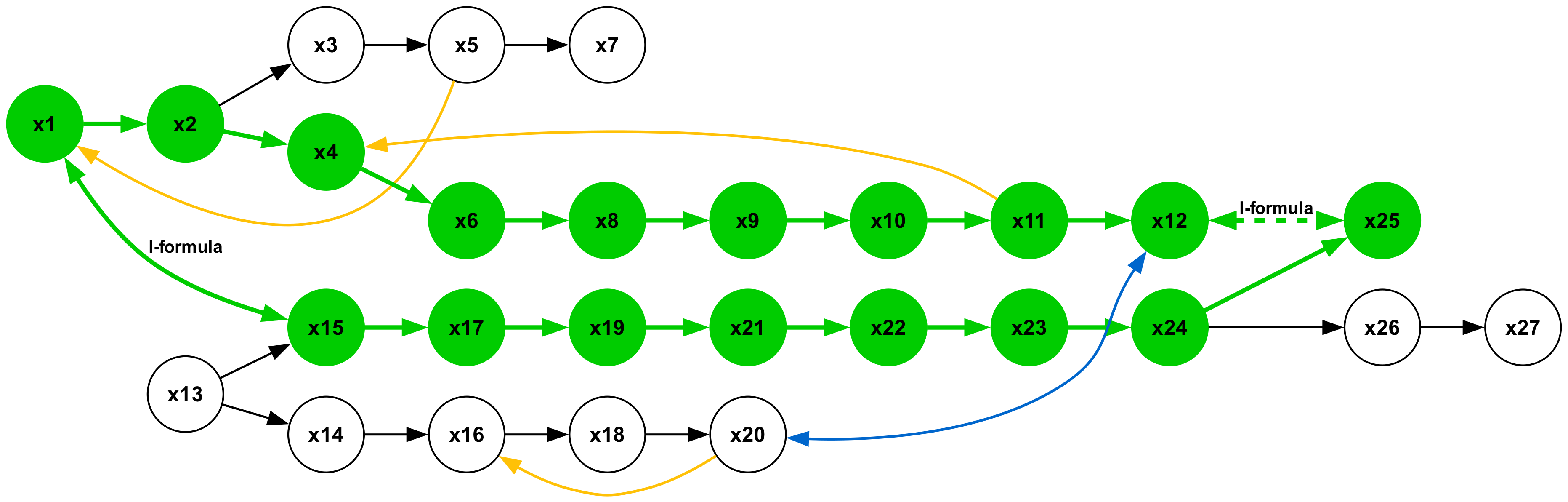}
  \end{minipage}

  % Linguistic description
  \vspace{1em}
  \textbf{Textual Translation:}
  \begin{tcolorbox}[
      colback=lightgray!20,
      fonttitle=\bfseries,
      fontupper=\small\ttfamily,
      sharp corners=southwest,
      rounded corners,
      width=\textwidth,
      boxrule=0pt,
      toptitle=0.1mm,
      bottomtitle=0.1mm
    ]
    \textbf{knowledge base:} 
    \textbf{\textcolor{darkgreen}{All x1 are x2}},
    All x2 are x3,
    \textbf{\textcolor{darkgreen}{All x2 are x4}},
    All x3 are x5,
    \textbf{\textcolor{darkgreen}{All x4 are x6}},
    All x5 are x7,
    \textbf{\textcolor{darkgreen}{All x6 are x8}}, 
    \textbf{\textcolor{darkgreen}{All x8 are x9}}, 
    \textbf{\textcolor{darkgreen}{All x9 are x10}},
    \textbf{\textcolor{darkgreen}{All x10 are x11}},
    \textbf{\textcolor{darkgreen}{All x11 are x12}},
    All x13 are x14,
    All x13 are x15,
    All x14 are x16,
    \textbf{\textcolor{darkgreen}{All x15 are x17}},
    All x16 are x18,
    \textbf{\textcolor{darkgreen}{All x17 are x19}},
    All x18 are x20,
    \textbf{\textcolor{darkgreen}{All x19 are x21}},
    \textbf{\textcolor{darkgreen}{All x21 are x22}},
    \textbf{\textcolor{darkgreen}{All x22 are x23}},
    \textbf{\textcolor{darkgreen}{All x23 are x24}},
    \textbf{\textcolor{darkgreen}{All x24 are x25}},
    All x24 are x26,
    All x26 are x27,
    No x20 are x12,
    \textbf{\textcolor{darkgreen}{Some x15 are x1}},
    Some x11 are not x4,
    Some x5 are not x1,
    Some x20 are not x16 \\[1pt]
    
    \textbf{hypothesis:} Some x25 are x12 \\[1pt]
    
    \textbf{premises:} 
    \textbf{\textcolor{darkgreen}{All x1 are x2}},
    \textbf{\textcolor{darkgreen}{All x2 are x4}}, 
    \textbf{\textcolor{darkgreen}{All x4 are x6}}, 
    \textbf{\textcolor{darkgreen}{All x6 are x8}}, 
    \textbf{\textcolor{darkgreen}{All x8 are x9}}, 
    \textbf{\textcolor{darkgreen}{All x9 are x10}}, 
    \textbf{\textcolor{darkgreen}{All x10 are x11}},
    \textbf{\textcolor{darkgreen}{All x11 are x12}},
    \textbf{\textcolor{darkgreen}{All x15 are x17}},
    \textbf{\textcolor{darkgreen}{All x17 are x19}},
    \textbf{\textcolor{darkgreen}{All x19 are x21}},
    \textbf{\textcolor{darkgreen}{All x21 are x22}},
    \textbf{\textcolor{darkgreen}{All x22 are x23}},
    \textbf{\textcolor{darkgreen}{All x23 are x24}},
    \textbf{\textcolor{darkgreen}{All x24 are x25}},
    \textbf{\textcolor{darkgreen}{Some x15 are x1}}
  \end{tcolorbox}
  
  \caption{\textbf{Type 7 syllogistic inference on graphs}. Visualization of a type 7 syllogistic inference using a graph representation of an example $\mathcal{KB}$, alongside the corresponding textual translation. In the graph (top), nodes represent predicates. Black edges indicate A-formulas (``All As are Bs''), blue edges indicate I-formulas (``Some As are Bs''), red edges indicate E-formulas (``No As are Bs''), and yellow edges indicate O-formulas (``Some As are not Bs''). The query hypothesis is represented by a dashed green edge, and the edges that prove the hypothesis are highlighted in green. The text translation illustrates how the abstract graph representation is converted into a text format suitable for LM processing by applying fixed templates that represent logical formulas.}

  \label{fig:type_7}
\end{figure*}

Figure~\ref{fig:task_visualization} illustrates a type 2 inference, characterized by a conclusion in the form of a universal affirmative ($A$-formula). 
The premises consist of a single sequence of $A$-formulas. This represents the most elementary form of syllogistic inference, whose structural pattern is embedded 
within all other types. Inferences of types 1, 3, and 5, which yield particular negative conclusions ($O$-formulas), are presented in Figures~\ref{fig:type_1},~\ref{fig:type_3}, and~\ref{fig:type_5}, respectively. 
Syllogisms corresponding to types 4 and 7, both concluding with particular affirmative statements ($I$-formulas), are shown in Figures~\ref{fig:type_4} and~\ref{fig:type_7}. 
Finally, the type 6 inference, which concludes with a universal negative ($E$-formula), is depicted in Figure~\ref{fig:type_6}.

\subsection{Term Vocabulary}
\label{app:vocabulary}
To train and evaluate our models, we artificially generated 5000 unique pseudowords by randomly concatenating two syllables selected from a set of approximately 
300 of the most commonly used English syllables. 
Although these pseudowords are semantically meaningless, they remain phonologically plausible and are generally pronounceable. 
On occasion, the generation process may yield actual English words.

Additionally, we constructed two substitution sets to support our lexical generalization evaluation (see Appendix~\ref{app:lexical-gen}). 
The first set comprises 5000 pseudowords generated using the Wuggy pseudoword generator \cite{keuleers2010wuggy}. 
We selected 500 English two-syllable nouns and, for each, produced 10 distinct pseudowords using Wuggy's default parameters. 
The second set consists of symbolic constants, each formed by the character “X” followed by an integers ranging from 1 to 5000.

\subsection{Data Statistics}
\label{app:data_stats}
\begin{figure*}[ht]
    \centering
    \begin{tabular}{cc}
    % Header row
    \textbf{Train/Val} & \textbf{Test} \\[1ex]
    % Overall row
    \includegraphics[width=0.45\linewidth]{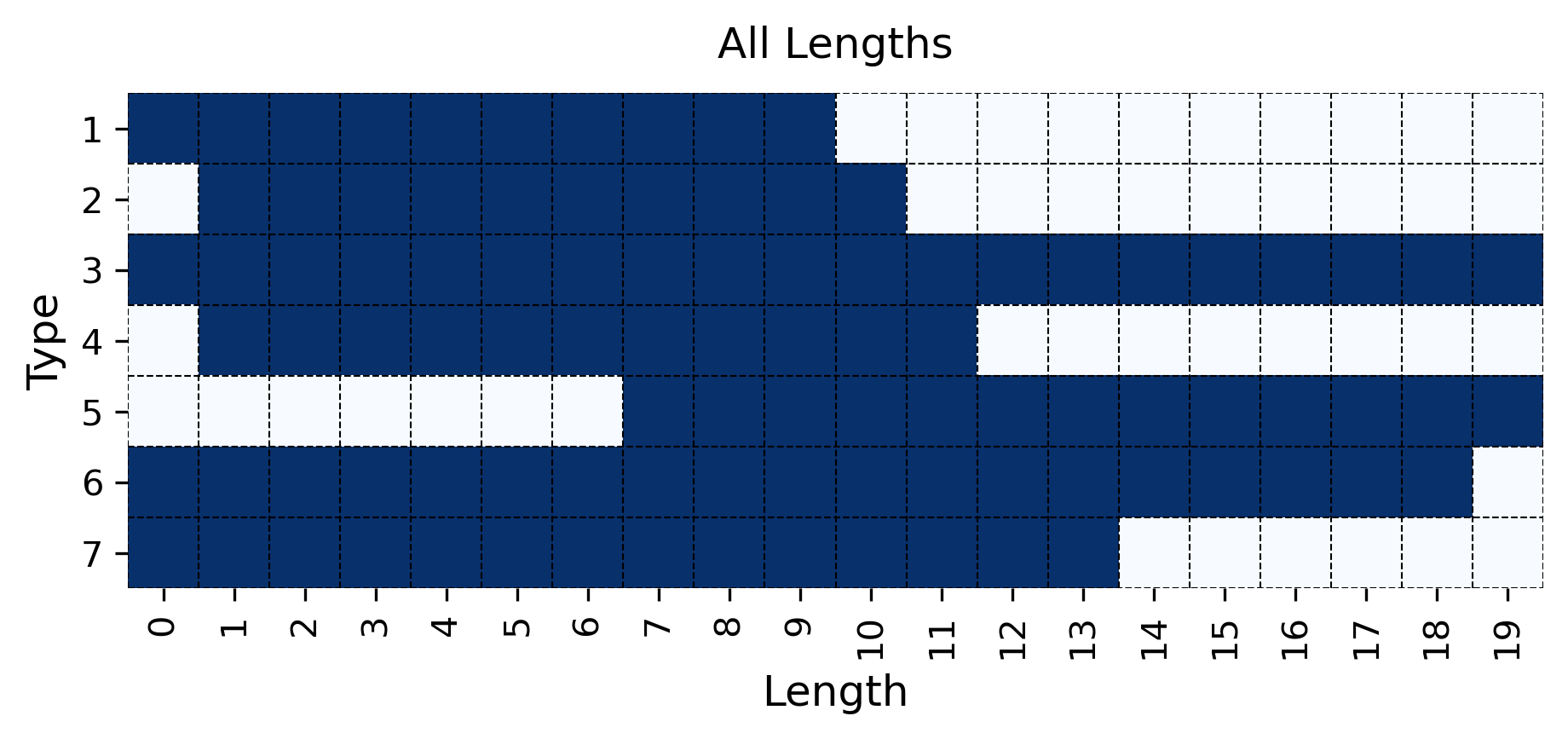} 
    & \includegraphics[width=0.45\linewidth]{figs/overall_trainval.png} \\[2ex]
    % Compositionality row
    \includegraphics[width=0.45\linewidth]{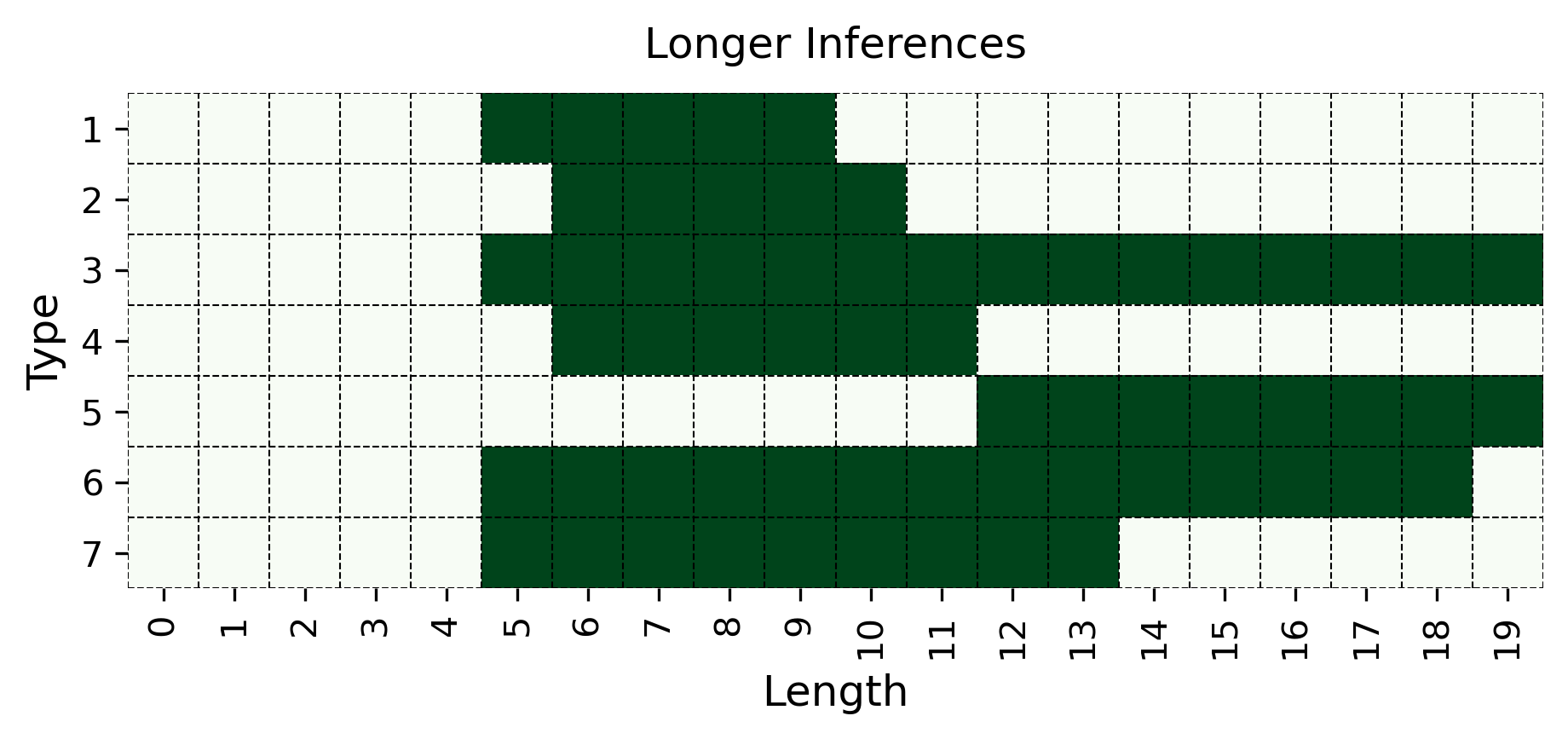} 
    & \includegraphics[width=0.45\linewidth]{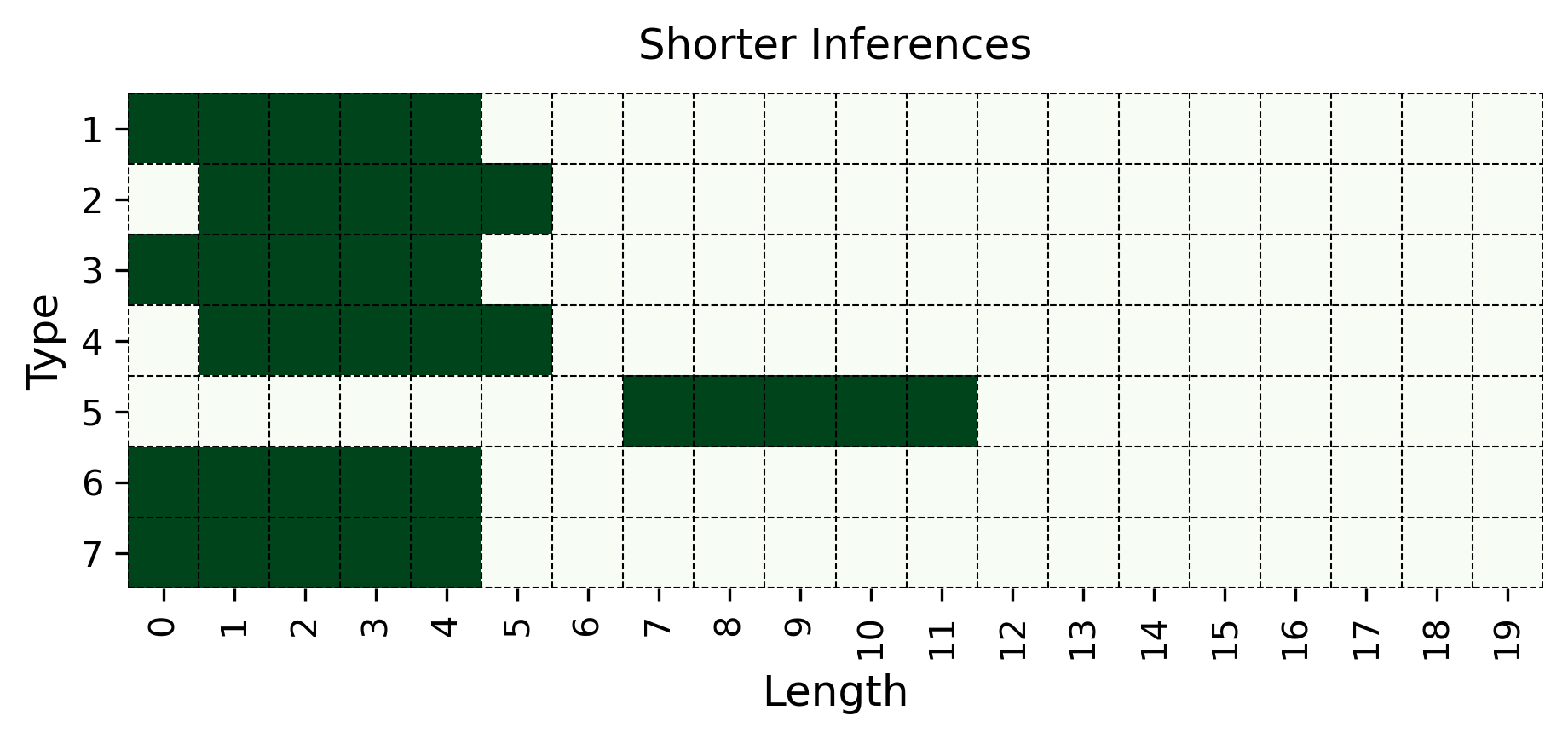} \\[2ex]
    % Recursiveness row
    \includegraphics[width=0.45\linewidth]{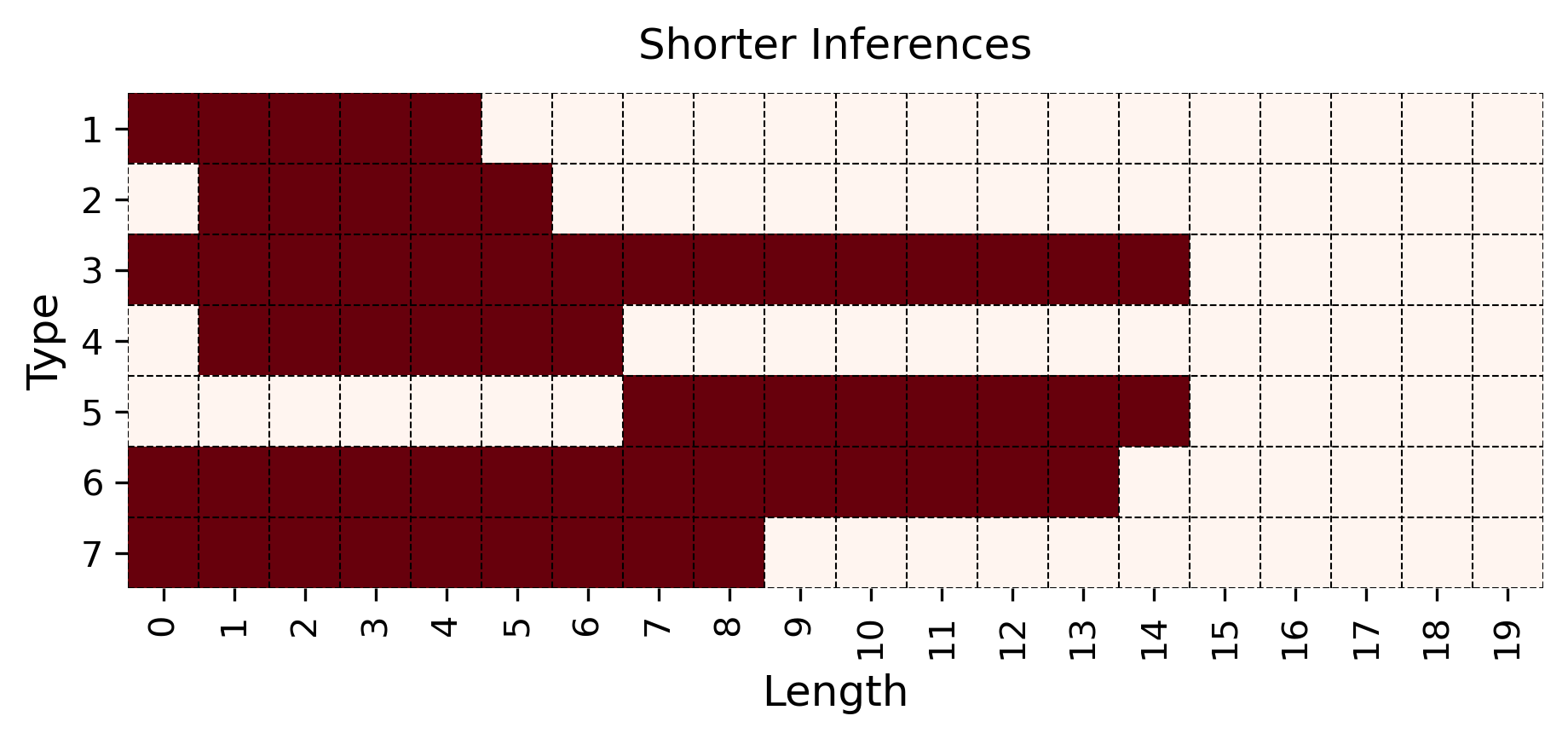} 
    & \includegraphics[width=0.45\linewidth]{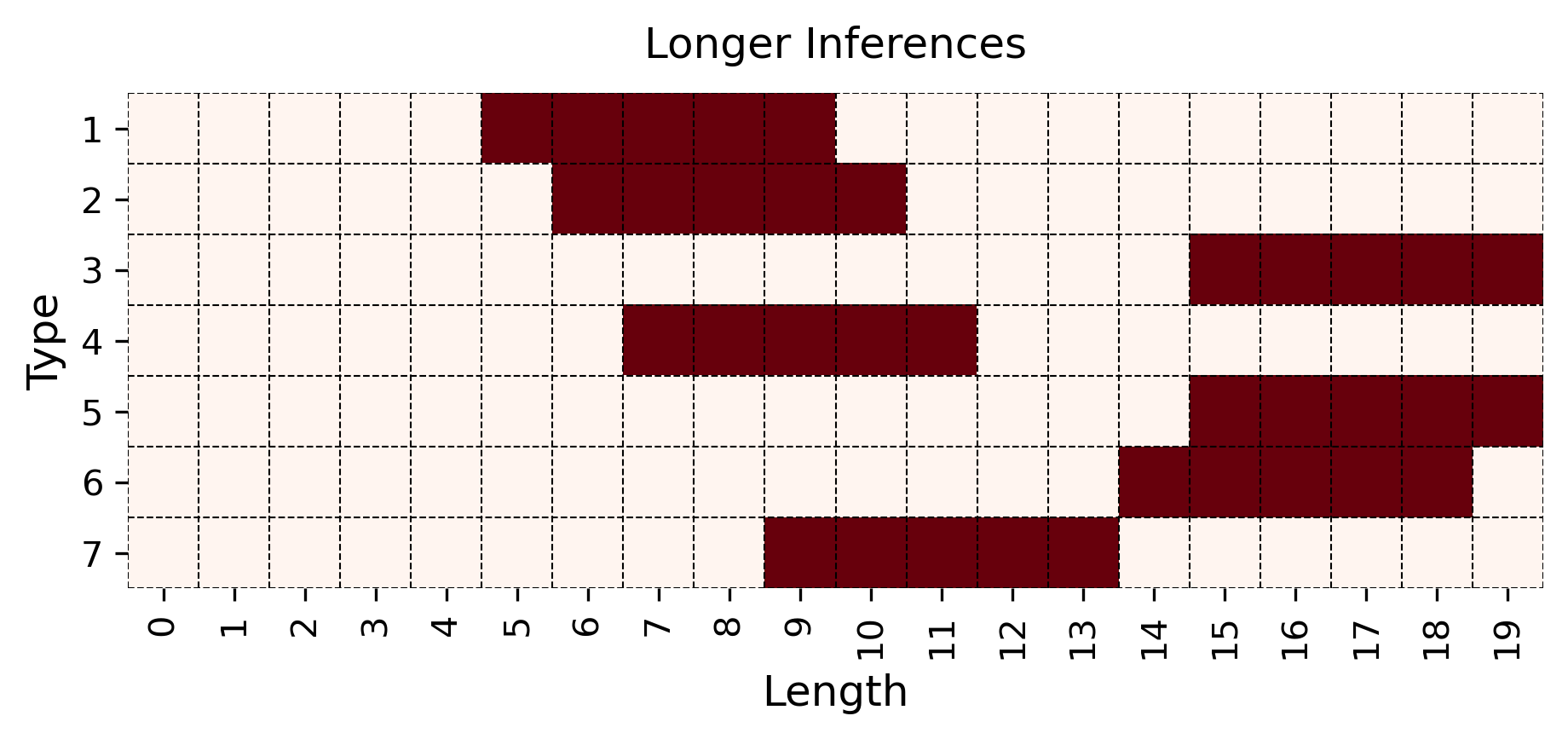} \\
    \end{tabular}
    \caption{\textbf{Combination of inference type and length within generated $\mathcal{KB}$s}. In each heatmap, rows represent Inference Types (1–7), while columns represent Lengths (0–19). The train, validation, and test splits use fixed values of 1000 or 100, 5, and 100 samples respectively for all non-zero entries (Colored). Entries with values equal to 0 indicate non-existing combinations of length and type within the split that is considered (White).}
    \label{fig:data-stats}
\end{figure*}

As described in Section \ref{sec:data}, we generated as many KBs as necessary to obtain at least 1000 training, 5 validation, and 100 test examples for each inference type and length combination in the range from 0 to 19 (see all the combinations in Figure \ref{fig:data-stats}). Table~\ref{tab:more-data-stats} summarizes dataset statistics for the core generalization experiment, as well as for the length generalization ones (``Short~$\rightarrow$~Long'' and ``Long~$\rightarrow$~Short''). For each experiment and split, the table provides the total number of examples, the number of  $\mathcal{KB}$s used to generate them, and the range of premises across $\mathcal{KB}$s. In the additional experiment with limited training data described in Appendix \ref{app:limited-data}, the total training size is reduced by a factor of ten in each setting.

\section{Experiment Details}
\label{app:exp-details}

\subsection{Implementation Details}
\label{app:implementation}
All experiments were conducted using the PyTorch and Hugging Face Transformers libraries. We used NVIDIA A100 80GB GPUs. Due to the relatively small size of the models used in the experiments, each fine-tuning run, both for ML and the baseline, was able to fit on a single GPU. We estimate a total compute usage of approximately 500 GPU hours across all experiments. Additionally, GitHub Copilot was used as an assistant tool for parts of the project’s source code development.

\begin{figure*}[ht]
\centering
\begin{tcolorbox}[colback=lightgray!20, boxrule=0pt, arc=2mm, width=0.9\linewidth]
\ttfamily % Monospaced font to resemble code
You are tasked with logical premise selection. Given: \\ 1. A knowledge base consisting of premises. \\ 2. A query hypothesis to solve, preceded by the token <QUERY>. \\ Your task is to identify the unique minimal set of premises from the knowledge base that logically proves the query hypothesis. Since the knowledge base is non-redundant, every valid hypothesis has exactly one minimal set of premises that proves it. \\ Provide your answer in exactly this format: \\ \#\#\# Answer: premise1, premise2, ..., premiseN
\end{tcolorbox}
\caption{\textbf{Zero-shot system prompt.} The zero-shot system prompt used with the closed models GPT-4o and o3-mini. The query hypothesis is subsequently provided as the first user interaction. We then extract the set of premises returned by the model using regular expressions.}
\label{fig:zero-shot-prompt}
\end{figure*}

\begin{figure*}[ht]
\centering
\begin{tcolorbox}[colback=lightgray!20, boxrule=0pt, arc=2mm, width=0.9\linewidth]
\ttfamily
You are tasked with logical premise selection. Given: \\ 1. A knowledge base consisting of premises. \\ 2. Example hypotheses along with their correct minimal premise sets, preceded by the token <STUDY>. \\ 3. A query hypothesis to solve, preceded by the token <QUERY>. \\ Your task is to identify the unique minimal set of premises from the knowledge base that logically proves the query hypothesis. Since the knowledge base is non-redundant, every valid hypothesis has exactly one minimal set of premises that proves it. \\ Examine the provided examples carefully to understand how to select the correct minimal set of premises. The examples demonstrate correct premise selections for various hypotheses. \\ Provide your answer in exactly this format: \\ \#\#\# Answer: premise1, premise2, ..., premiseN
\end{tcolorbox}
\caption{\textbf{Few-shot system prompt.} The Few-shot system prompt used with the closed models GPT-4o and o3-mini. The set of study examples provided as few-shot examples, along with the query hypothesis are provided as the first user interaction. We then extract the set of premises returned by the model using regular expressions.}
\label{fig:few-shot-prompt}
\end{figure*}

\subsection{Fine-tuning Details}
\label{app:training-details}

All models were fine-tuned using Low-Rank Adaptation (LoRA) \citep{hu2022lora} with a rank $r=64$, alpha value $\alpha=128$, and dropout probability $p=0.05$. The adaptation was applied to all attention and linear weight matrices, excluding the embedding and unembedding layers. Baseline models were loaded in \texttt{bfloat16} precision, while ML fine-tuned models employed QLoRA \citep{dettmers2023qlora} with 4-bit quantization to accommodate memory constraints from longer sequences. Despite the lower precision, the meta-learning models outperformed the baseline.  

Training hyperparameters included a learning rate of $5\times10^{-5}$, zero weight decay, and no learning rate warmup (steps=0, ratio=0.0). Batch sizes were 4 (training), 8 (validation), and 32 (testing). We used the AdamW optimizer \citep{kingma2017adam} with a linear learning rate scheduler. Although we experimented with a range of other hyperparameter configurations, we found this setup to be the most stable across tasks and random seeds. Baseline models were trained for 4 epochs, whereas meta-learning models were trained for only 1 epoch to account for differences in per-sample data exposure (see Section \ref{sec:mind}). We performed 10 validations per epoch and selected the model with the highest validation accuracy. Each fine-tuning run was repeated with three different random seeds: 1048, 512, and 1056.

\subsection{Closed Source Models}
\label{app:closed_models}

\paragraph{API details.}
We accessed OpenAI's closed-source models GPT-4o \citep{gpt4o} and o3-mini \citep{o3mini} through the Azure OpenAI Service's Batch API. The API version used was \texttt{2025-03-01-preview}, and the specific model versions were \texttt{gpt-4o-2024-08-06} and \texttt{o3-mini-2025-01-31}. The total cost of the experiments was approximately 250 USD. For both models, we employed the default API settings. In the case of o3-mini, this corresponds to a “medium” reasoning effort. We did not experiment with a high reasoning effort in order to limit API usage costs.

\paragraph{Prompts.} We provide the exact system prompts used in the experiments involving GPT-4o and o3-mini in both the zero-shot (Figure~\ref{fig:zero-shot-prompt}) and few-shot (Figure~\ref{fig:few-shot-prompt}) settings. In both cases, the system prompt instructs the models on how to perform the task and specifies the exact format of the answer they should provide. This format facilitates the extraction of the set of premises generated by the models. We then present the query hypothesis as the first user interaction. In the few-shot setting, example interactions are included in the user message prior to the query.

\begin{table*}[ht]
\centering
\begin{tabular}{l l c c c}
\toprule
\textbf{Model} & \textbf{Type} & \textbf{Core} & \textbf{Unseen Pseudowords} & \textbf{Unseen Constants} \\
\midrule
\multirow{2}{*}{Qwen-2.5 1.5B} 
              & ML           & 93.11 ± 0.61     & 93.15 ± 0.11    & 74.24 ± 1.07     \\
              & $\mathrm{Baseline}_{-S}$      & 85.56 ± 1.24     & 83.34 ± 1.90    & 38.49 ± 1.06     \\
\cmidrule(lr){2-5}
\multirow{2}{*}{Qwen-2.5 3B} 
              & ML           & 96.16 ± 0.44     & 96.09 ± 0.30    & 83.21 ± 1.19     \\
              & $\mathrm{Baseline}_{-S}$      & 93.03 ± 1.15     & 91.49 ± 0.68    & 53.12 ± 2.03     \\
\cmidrule(lr){2-5}
\multirow{2}{*}{Qwen-2.5 7B} 
              & ML           & 98.13 ± 0.98     & 98.03 ± 1.19    & 86.87 ± 0.31     \\
              & $\mathrm{Baseline}_{-S}$      & 95.76 ± 1.10     & 94.89 ± 1.55    & 57.81 ± 2.17     \\
\bottomrule
\end{tabular}
\caption{\textbf{Lexical generalization.} Accuracy (mean ± std) of ML and Baseline models in core generalization as in the main paper (Core) and with novel unseen terms (Unseen Pseudowords, Unseen Constants).}
\label{tab:lexical_gen}
\end{table*}

\begin{table*}[ht]
\centering
\begin{tabular}{l l c c c}
\toprule
\textbf{Model} & \textbf{Type} & \textbf{Core} & \textbf{Long $\rightarrow$ Short} & \textbf{Short $\rightarrow$ Long} \\
\midrule
\multirow{2}{*}{Qwen-2.5 1.5B} 
              & ML           & 76.67 ± 0.38     & 50.40 ± 3.45    & 45.81 ± 1.13    \\
              & $\mathrm{Baseline}_{-S}$      & 55.14 ± 0.53     & 29.37 ± 1.85    & 30.22 ± 1.52    \\
\cmidrule(lr){2-5}
\multirow{2}{*}{Qwen-2.5 3B} 
              & ML           & 84.68 ± 0.54     & 64.77 ± 0.73    & 53.95 ± 3.46    \\
              & $\mathrm{Baseline}_{-S}$      & 66.51 ± 0.19     & 43.66 ± 1.93    & 43.67 ± 2.05    \\
\cmidrule(lr){2-5}
\multirow{2}{*}{Qwen-2.5 7B} 
              & ML           & 88.01 ± 1.11     & 69.24 ± 9.79    & 60.90 ± 2.94    \\
              & $\mathrm{Baseline}_{-S}$      & 68.54 ± 2.25     & 45.27 ± 0.95    & 43.94 ± 2.82    \\
\bottomrule
\end{tabular}
\caption{\textbf{Generalization in limited data regime.} Accuracy (mean ± std) of meta-learning and baseline models trained and tested on all inference types and lengths (Core), as well as tested for longer or shorter inferences than those seen during training. The models are trained on only 100 examples for each combination of inference type and inference length.}
\label{tab:low_data}
\end{table*}

\section{Additional Results}
\label{app:additional_results}

\subsection{Accuracies by Type and Length}
\label{app:type_length}
In this section, we present the complete set of accuracies broken down by type and length for both ML and baseline models, as well as closed source models.

\begin{figure*}[htbp]
    \centering
    \includegraphics[width=0.8\linewidth]{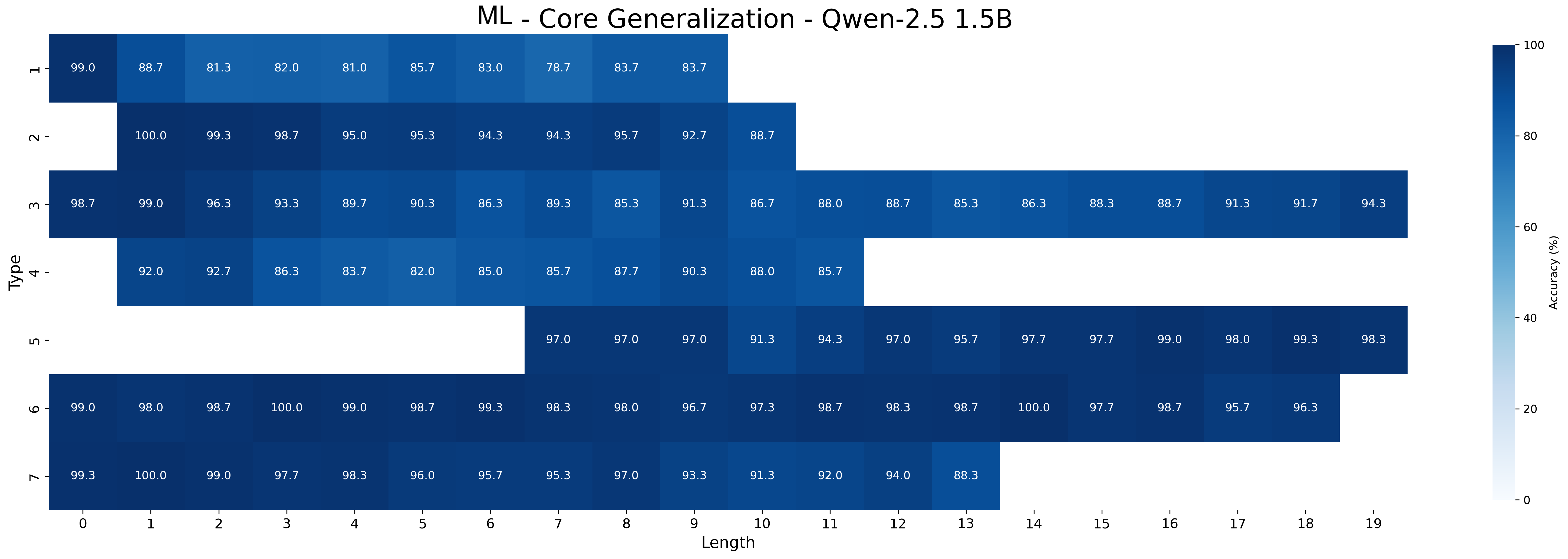} \\[10pt]
    \includegraphics[width=0.8\linewidth]{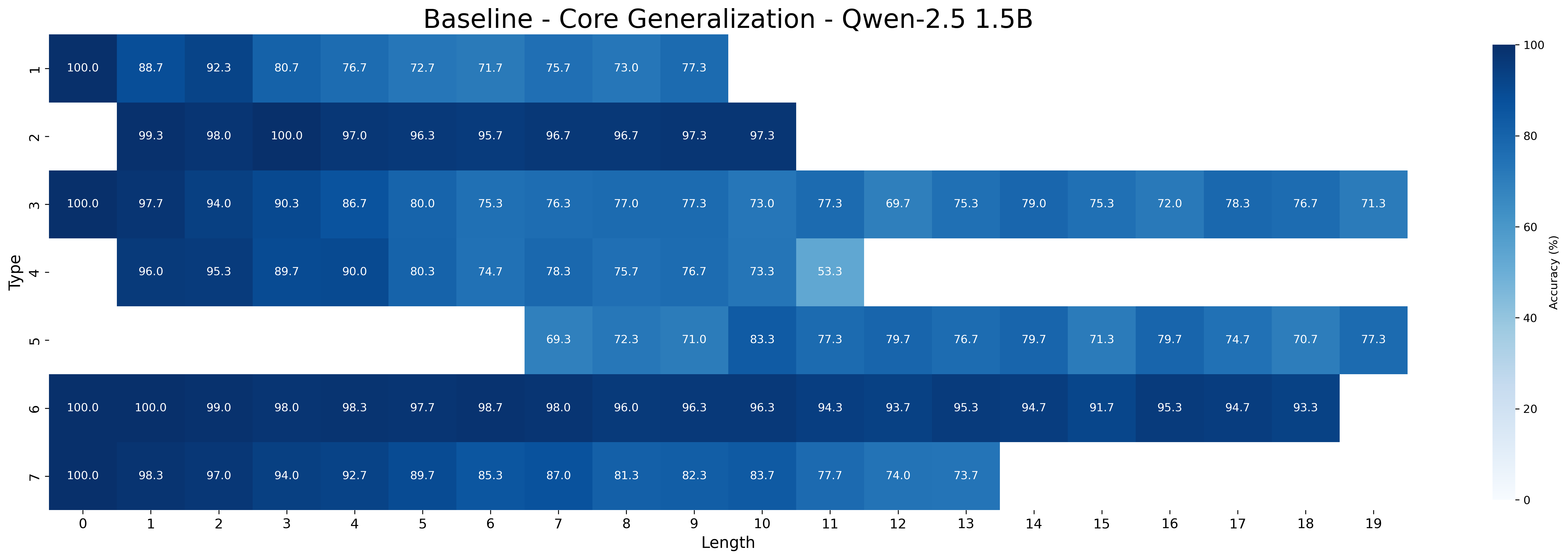}
    \caption{Accuracy of ML (Top) and Baseline (Bottom) Qwen-2.5 1.5B on core generalization decomposed by inference type and length.}
    \label{fig:qwen-1.5-core}
\end{figure*}

\begin{figure*}[htbp]
    \centering
    \includegraphics[width=0.8\linewidth]{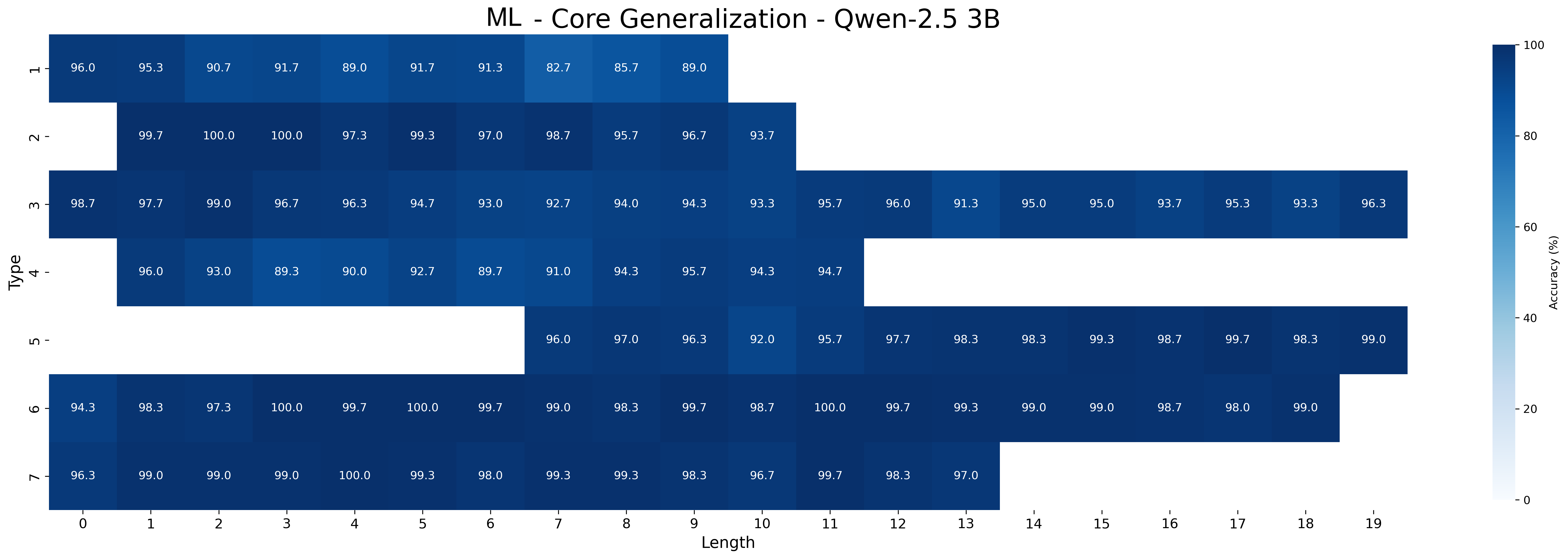} \\[10pt]
    \includegraphics[width=0.8\linewidth]{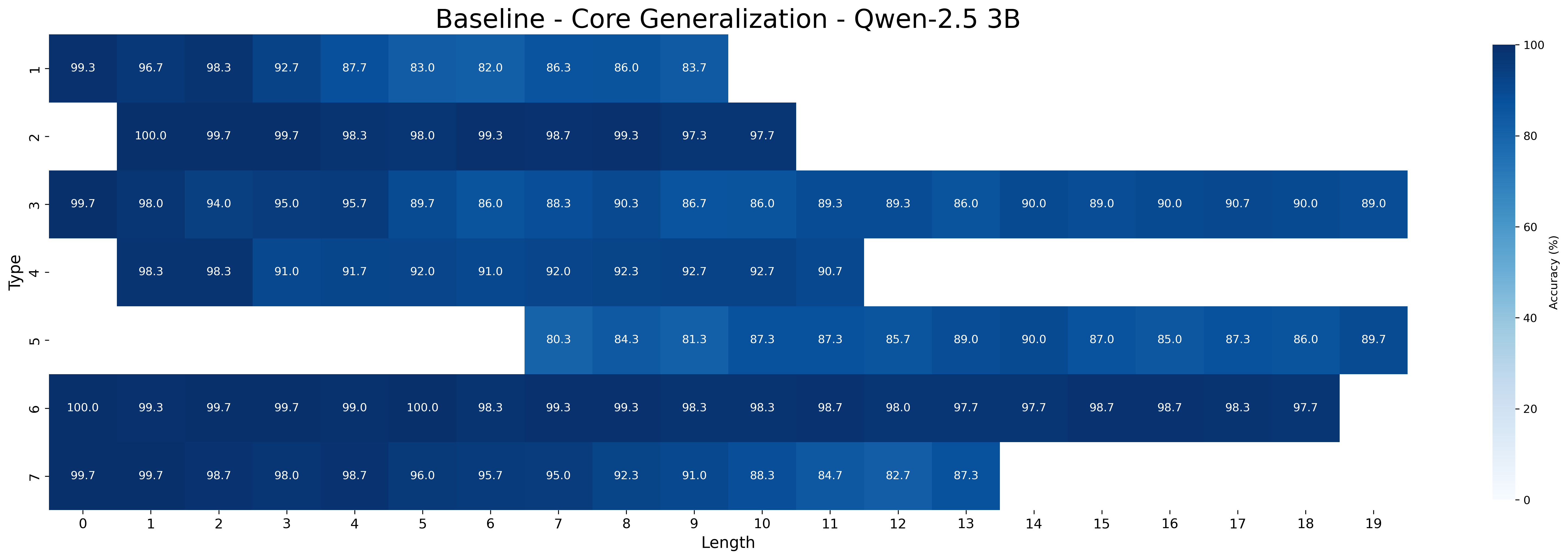}
    \caption{Accuracy of ML (Top) and Baseline (Bottom) Qwen-2.5 3B on core generalization decomposed by inference type and length.}
    \label{fig:qwen-3-core}
\end{figure*}

\begin{figure*}[htbp]
    \centering
    \includegraphics[width=0.8\linewidth]{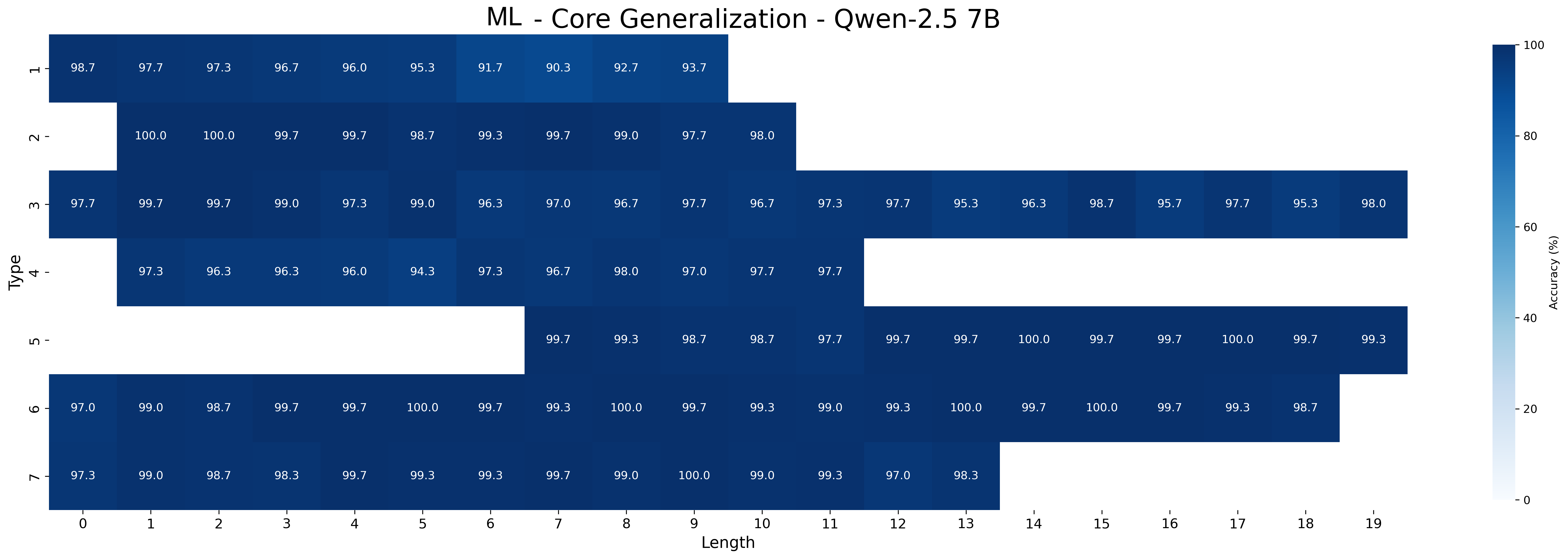} \\[10pt]
    \includegraphics[width=0.8\linewidth]{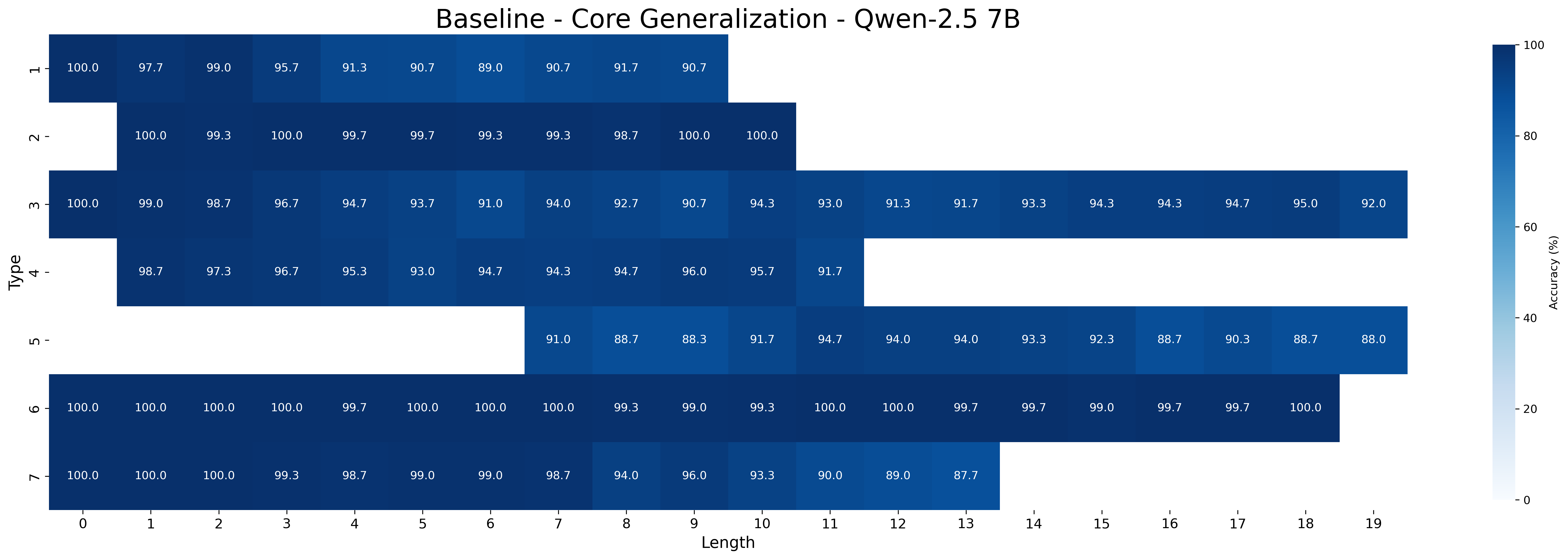}
    \caption{Accuracy of ML (Top) and Baseline (Bottom) Qwen-2.5 7B on core generalization decomposed by inference type and length.}
    \label{fig:qwen-7-core}
\end{figure*}

\begin{figure*}[htbp]
    \centering
    \includegraphics[width=0.45\linewidth]{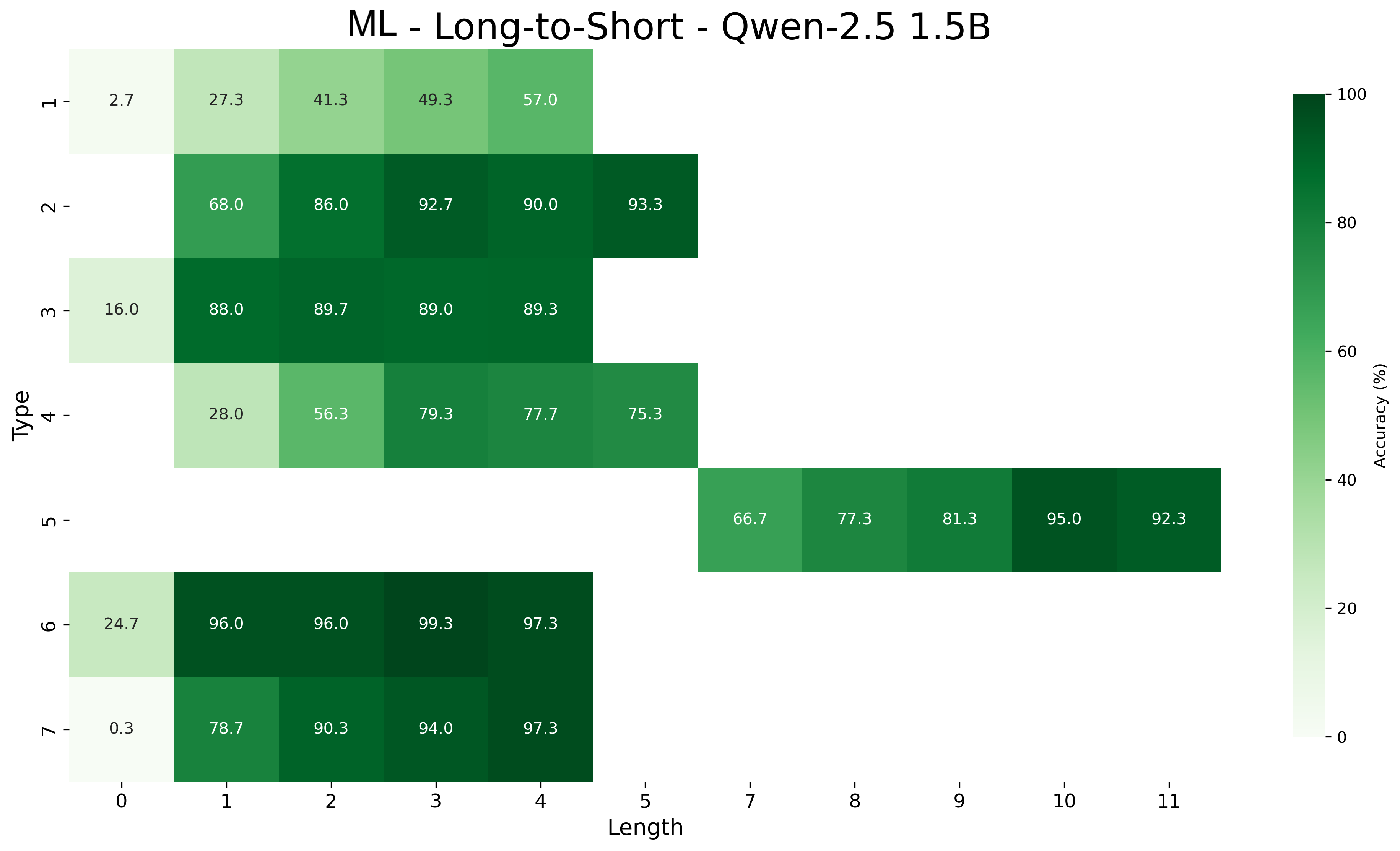}
    \hspace{3pt}
    \includegraphics[width=0.45\linewidth]{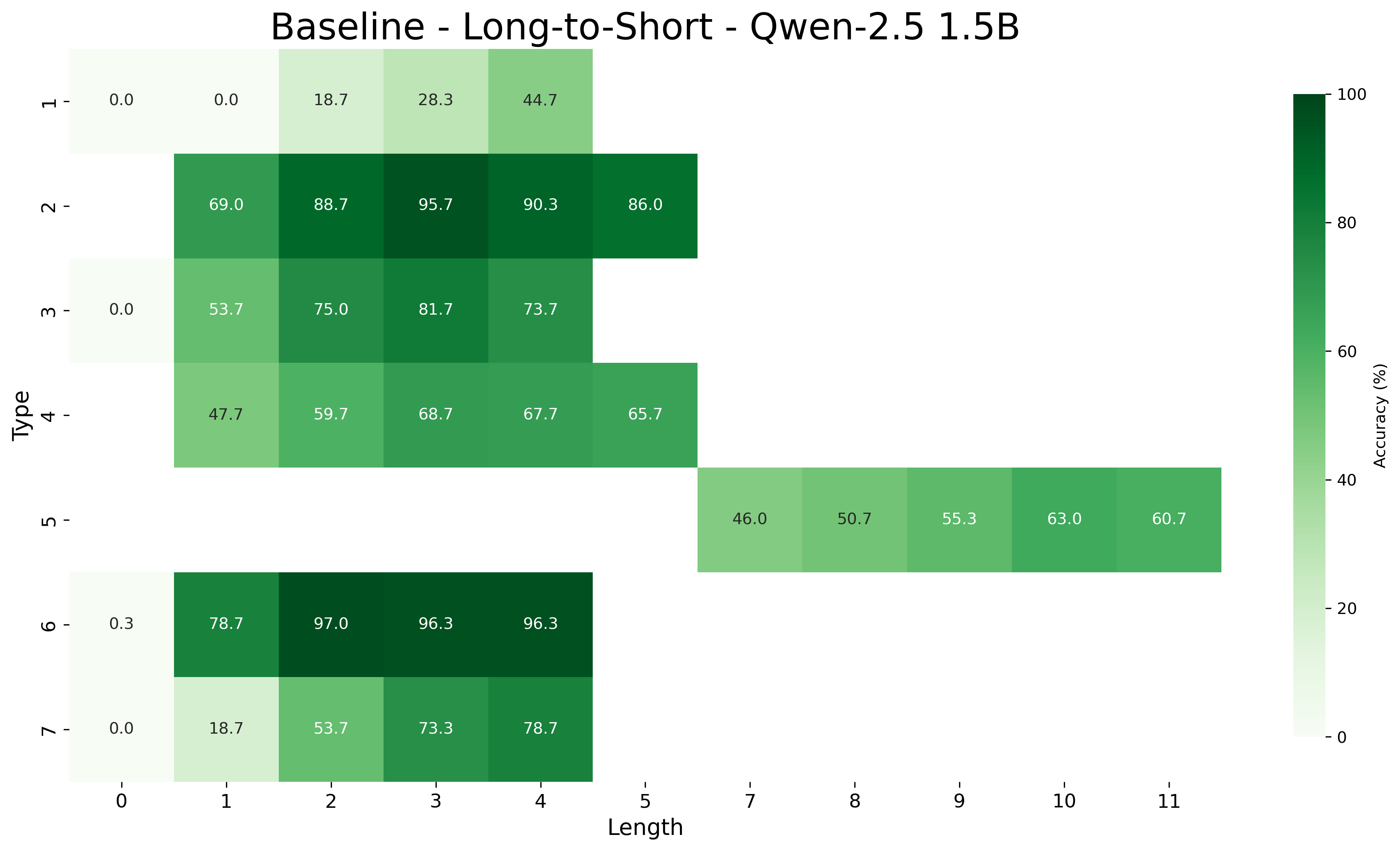}
    \caption{Accuracy of ML (Left) and Baseline (Right) Qwen-2.5 1.5B on long to short generalization decomposed by inference type and length.}
    \label{fig:qwen-1.5-long-to-short}
\end{figure*}

\begin{figure*}[htbp]
    \centering
    \includegraphics[width=0.45\linewidth]{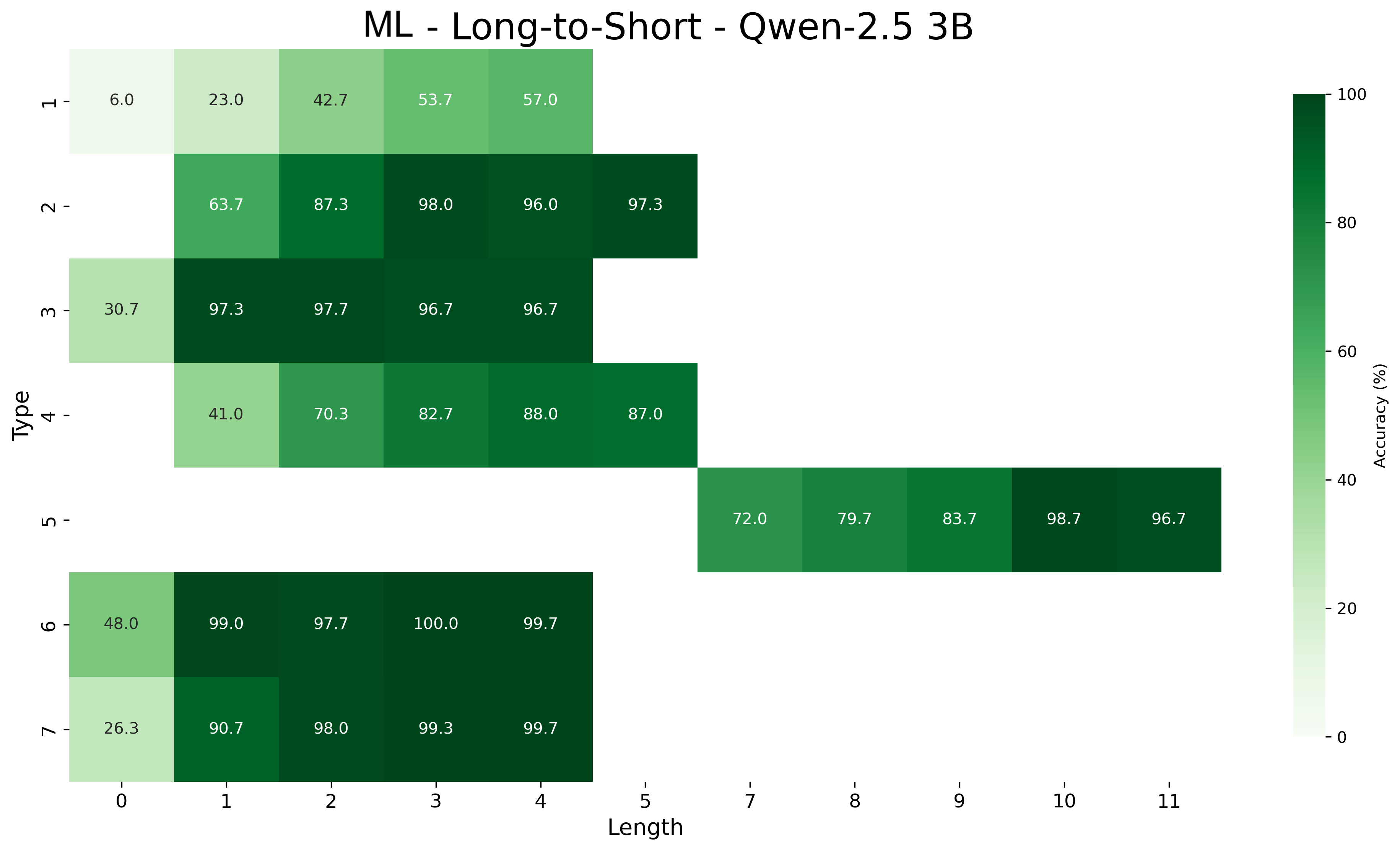}
    \hspace{3pt}
    \includegraphics[width=0.45\linewidth]{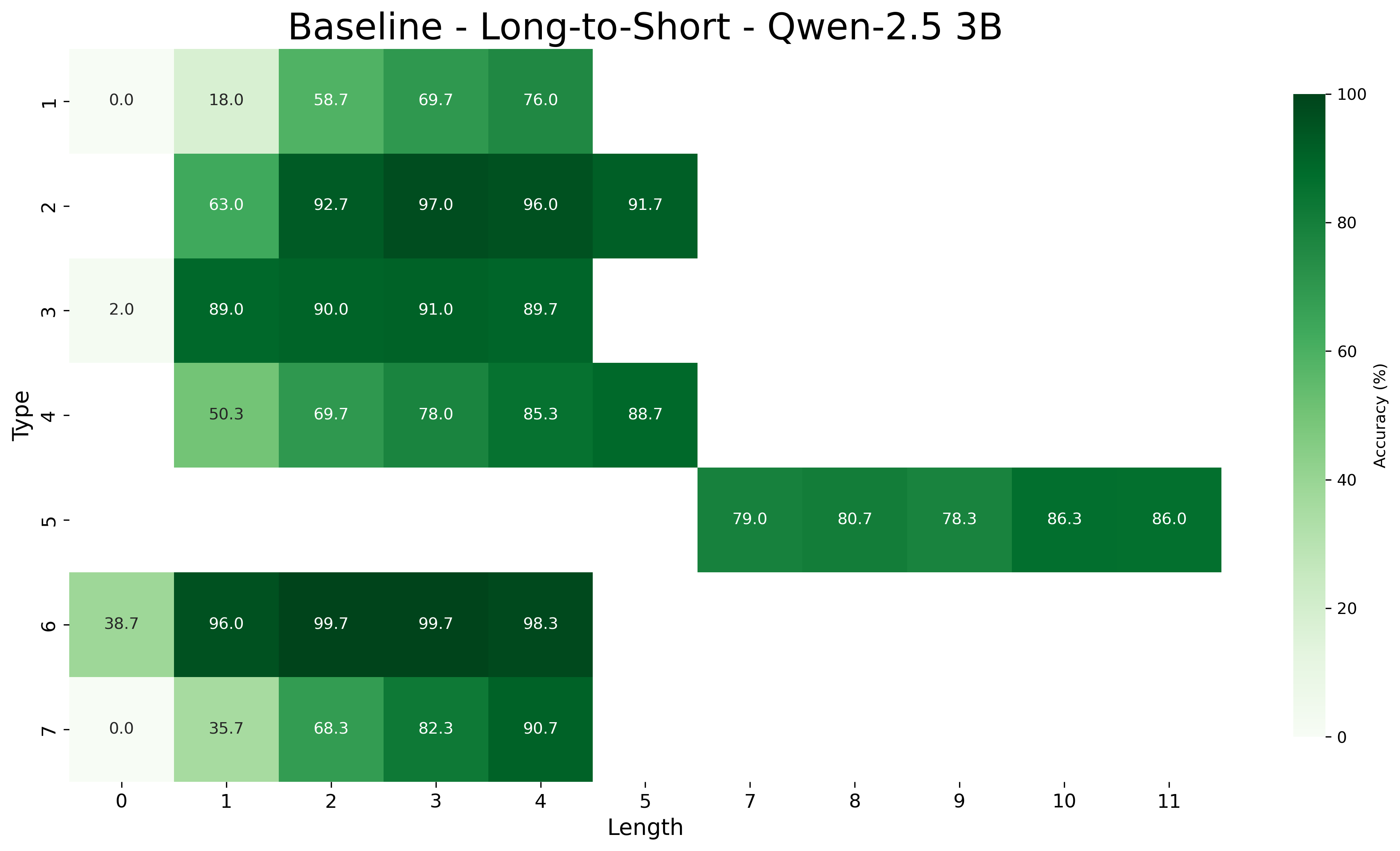}
    \caption{Accuracy of ML (Left) and Baseline (Right) Qwen-2.5 3B on long to short generalization decomposed by inference type and length.}
    \label{fig:qwen-3-long-to-short}
\end{figure*}

\begin{figure*}[htbp]
    \centering
    \includegraphics[width=0.45\linewidth]{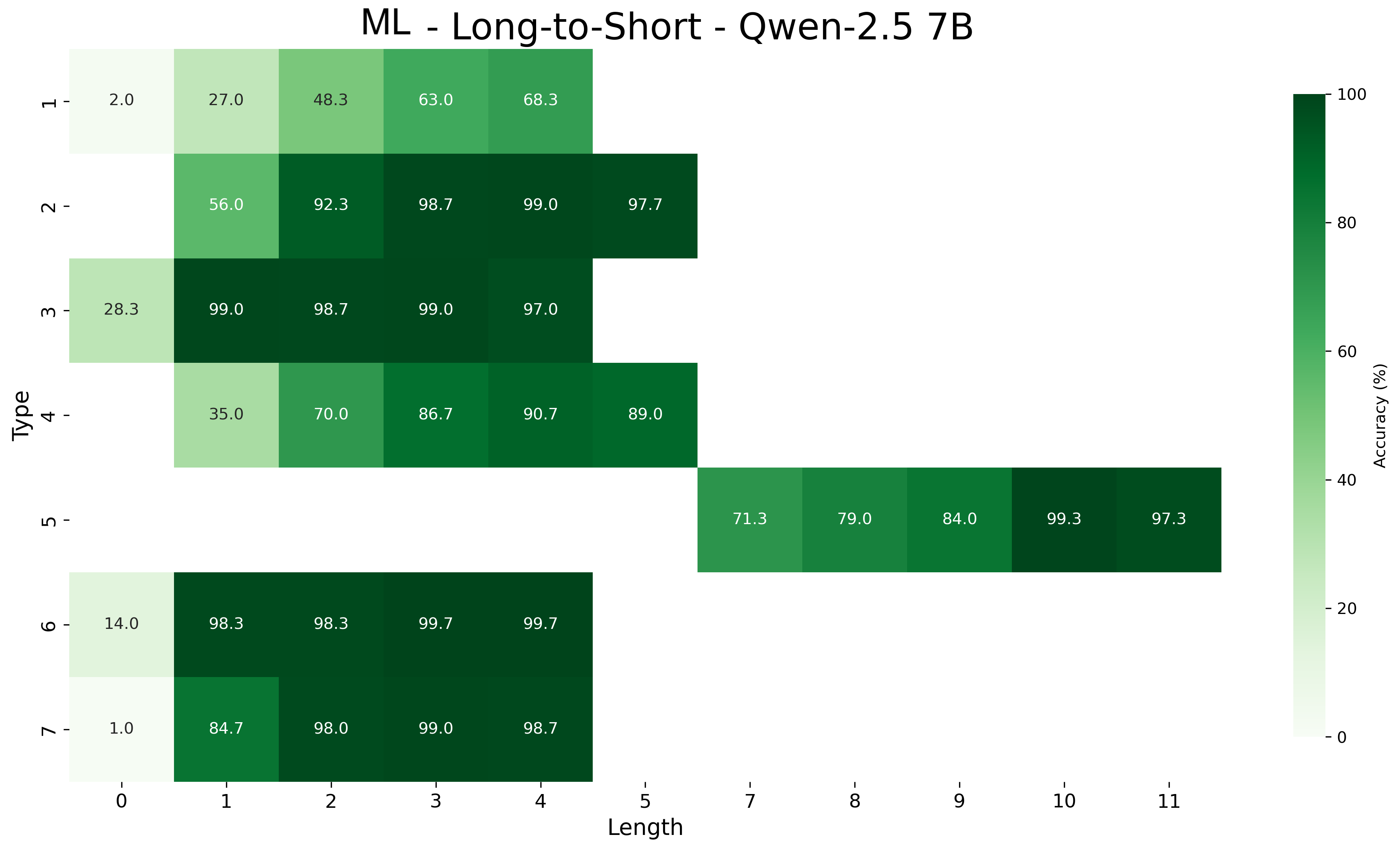}
    \hspace{3pt}
    \includegraphics[width=0.45\linewidth]{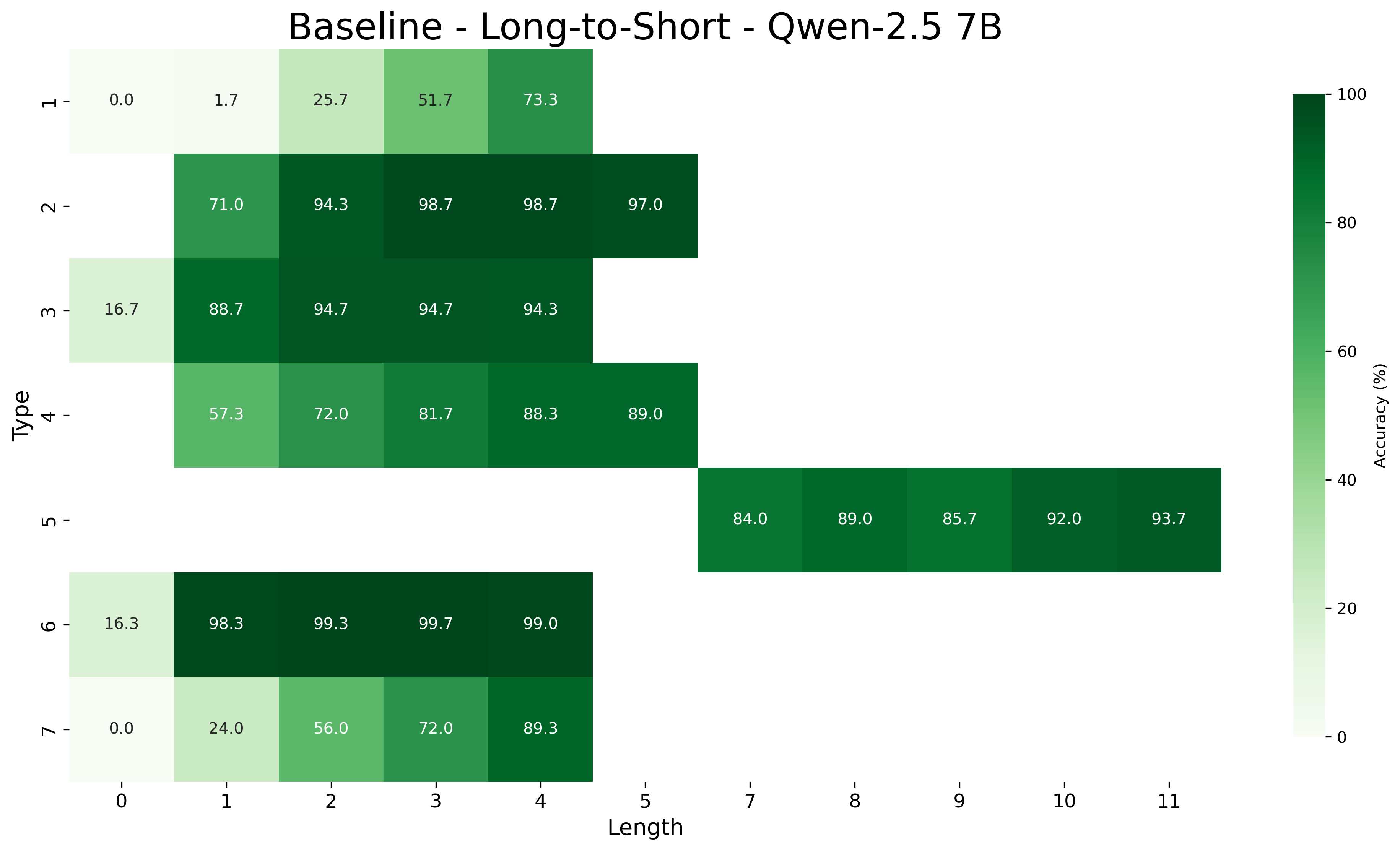}
    \caption{Accuracy of ML (Left) and Baseline (Right) Qwen-2.5 7B on long to short generalization decomposed by inference type and length.}
    \label{fig:qwen-7-long-to-short}
\end{figure*}

\begin{figure*}[htbp]
    \centering
    \includegraphics[width=0.45\linewidth]{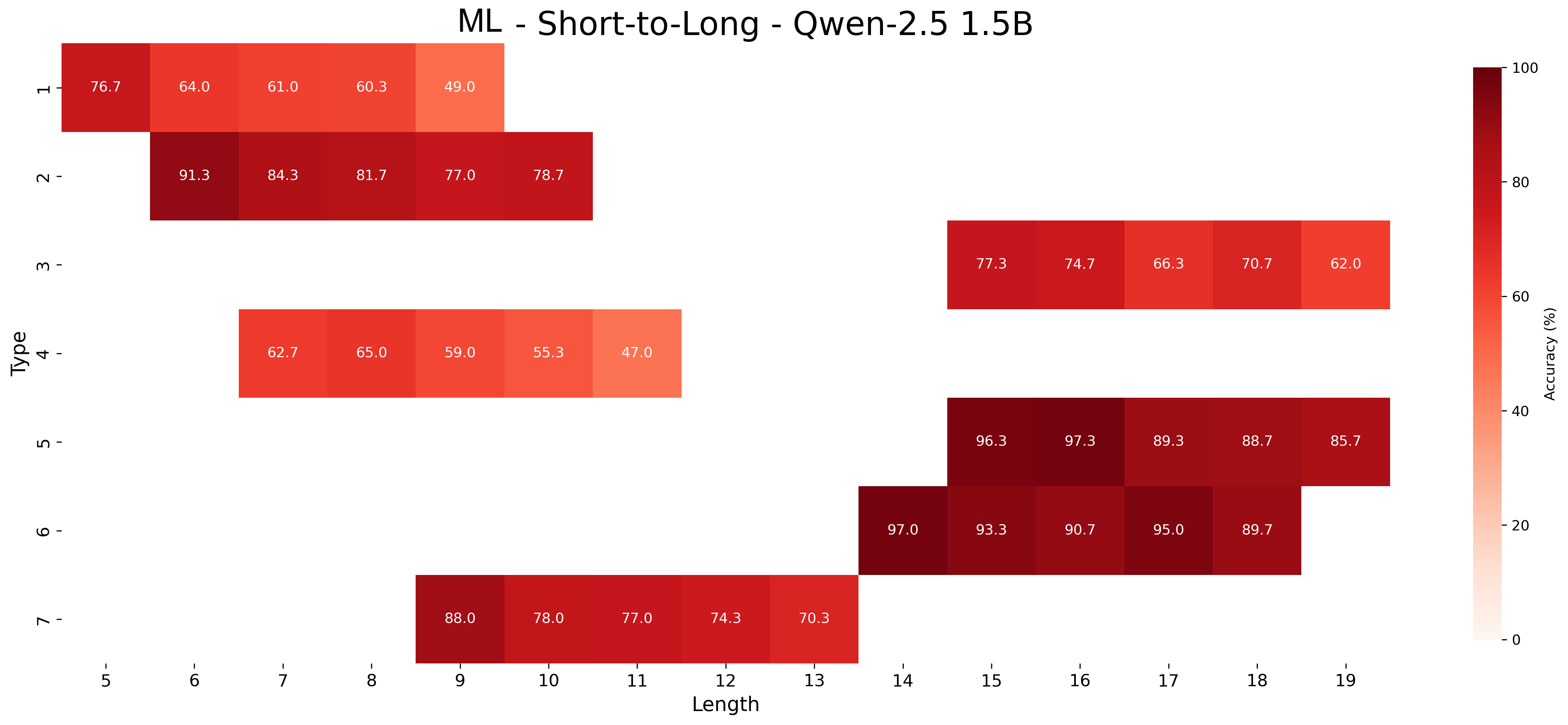}
    \hspace{3pt}
    \includegraphics[width=0.45\linewidth]{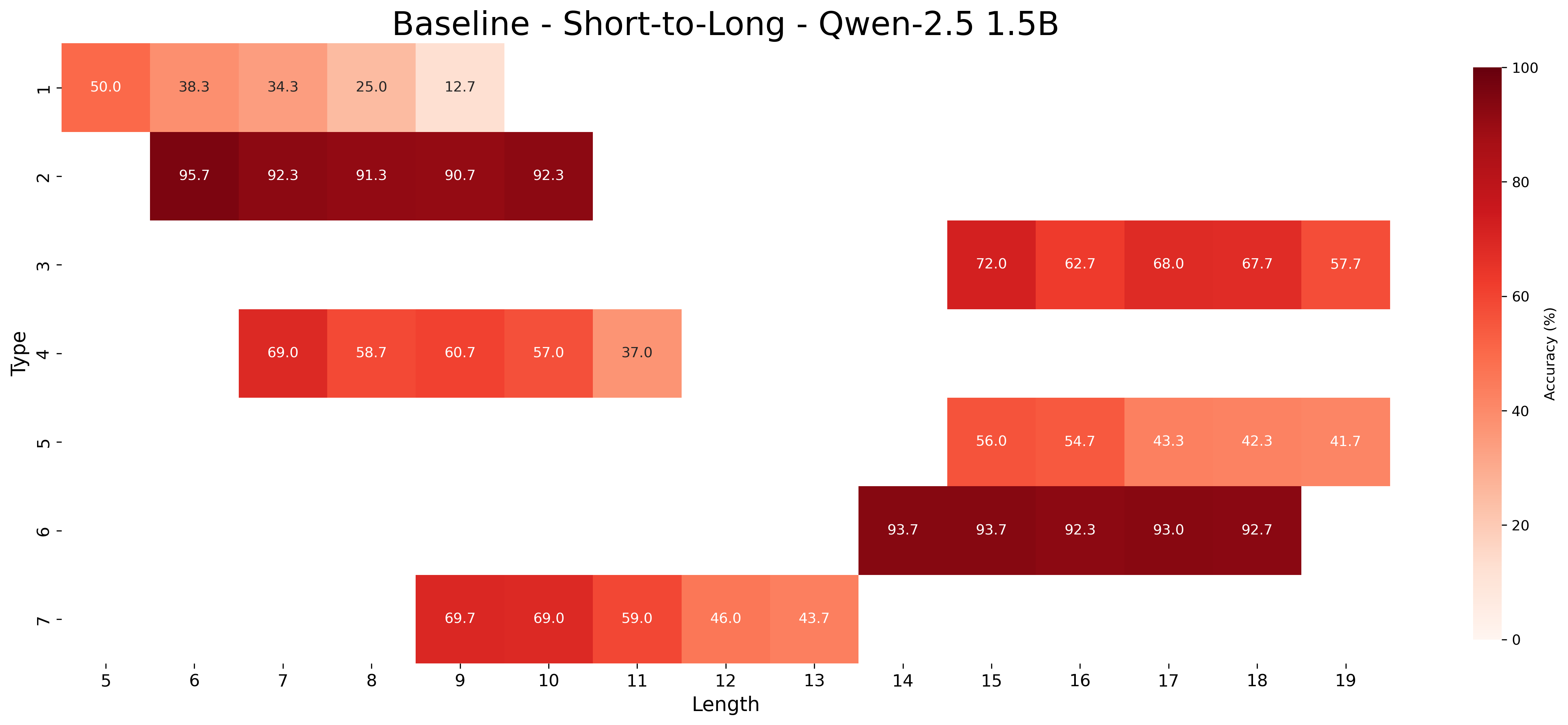}
    \caption{Accuracy of ML (Left) and Baseline (Right) Qwen-2.5 1.5B on short to long generalization decomposed by inference type and length.}
    \label{fig:qwen-1.5-short-to-long}
\end{figure*}

\begin{figure*}[htbp]
    \centering
    \includegraphics[width=0.45\linewidth]{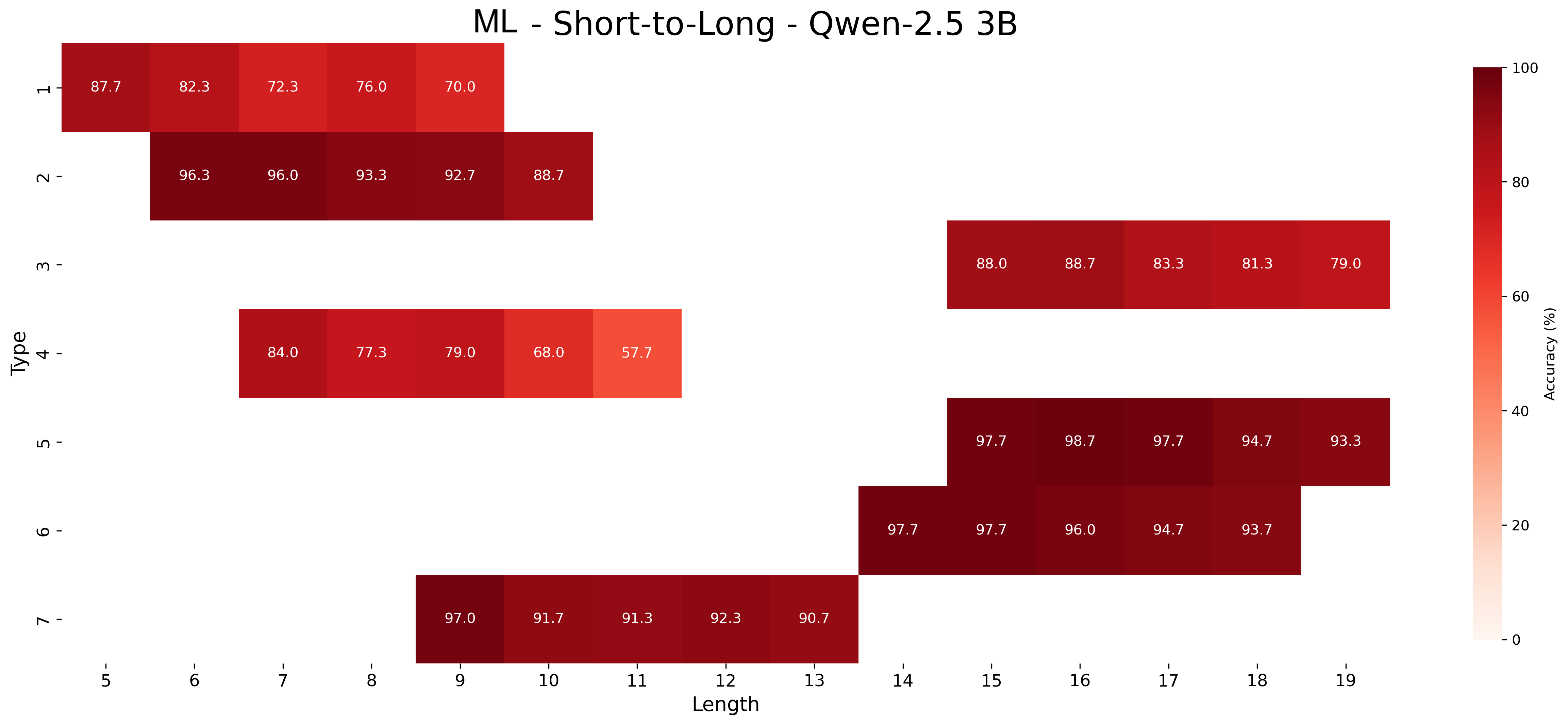}
    \hspace{3pt}
    \includegraphics[width=0.45\linewidth]{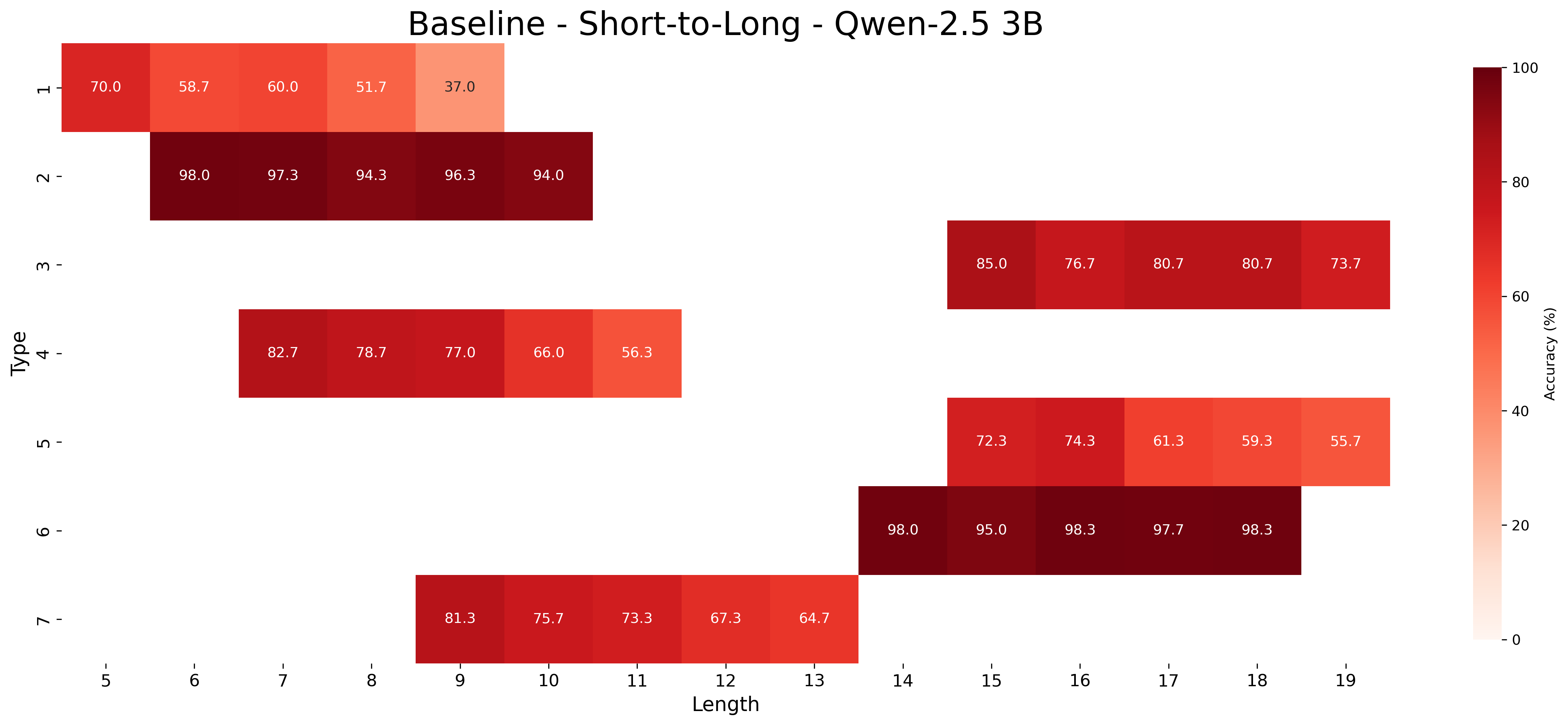}
    \caption{Accuracy of ML (Left) and Baseline (Right) Qwen-2.5 3B on short to long generalization decomposed by inference type and length.}
    \label{fig:qwen-3-short-to-long}
\end{figure*}

\begin{figure*}[htbp]
    \centering
    \includegraphics[width=0.45\linewidth]{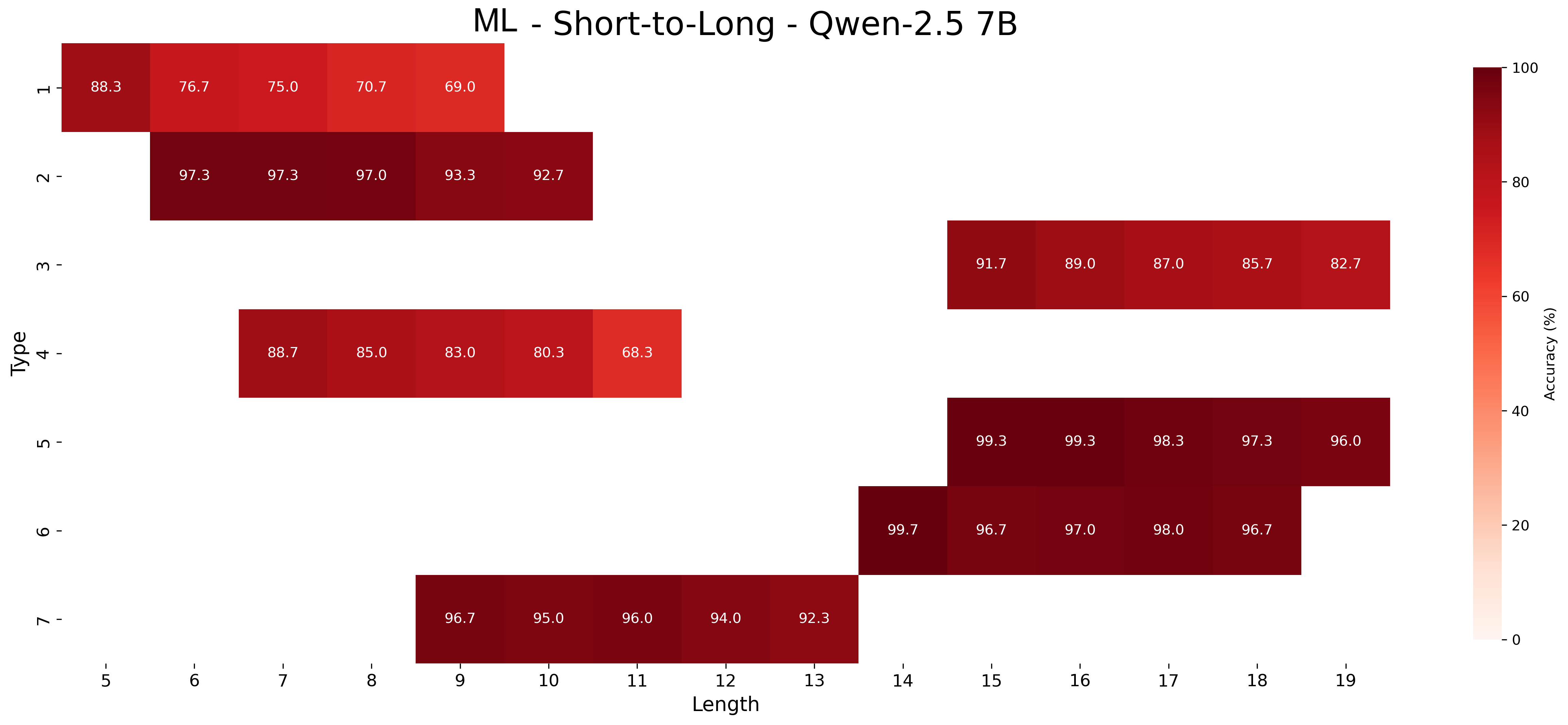} 
    \hspace{3pt}
    \includegraphics[width=0.45\linewidth]{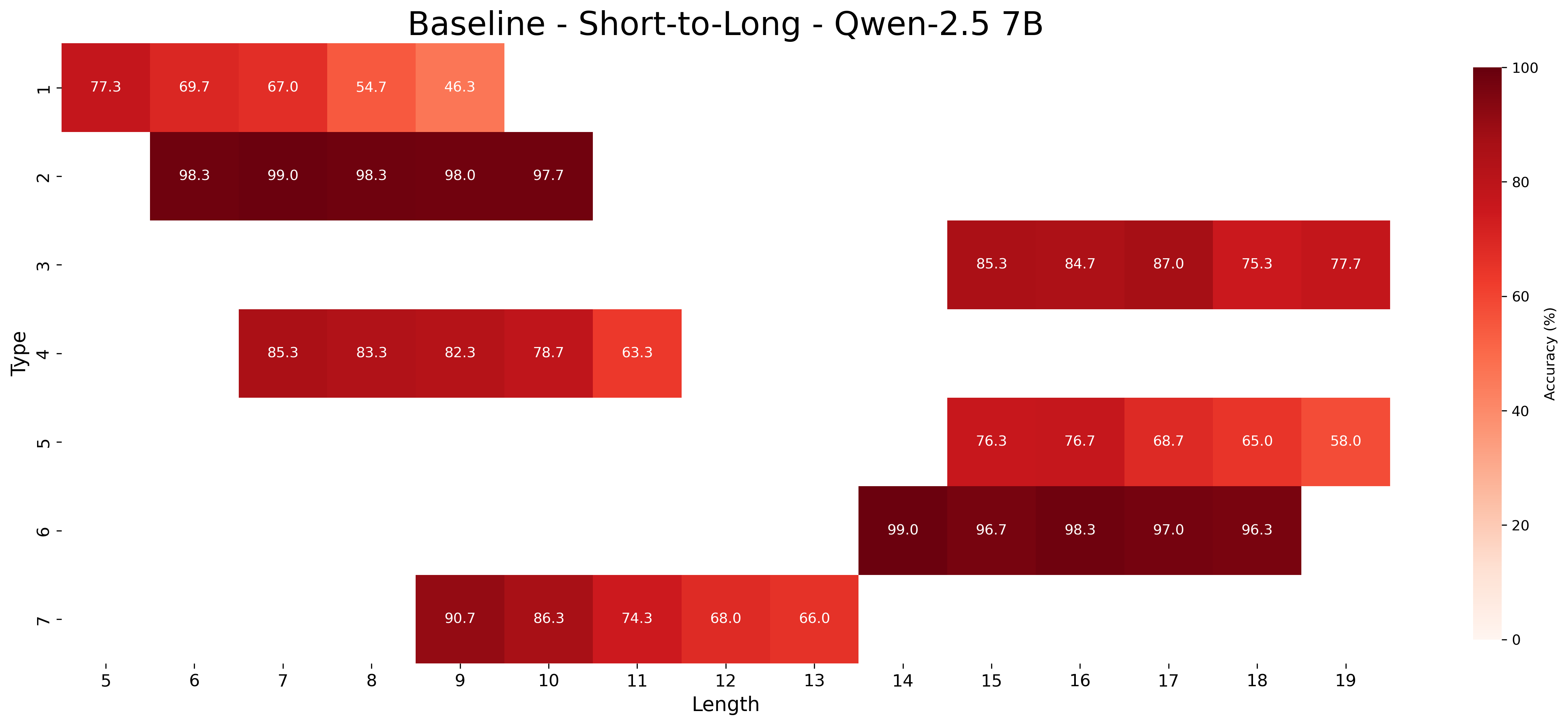}
    \caption{Accuracy of ML (Left) and Baseline (Right) Qwen-2.5 7B on short to long generalization decomposed by inference type and length.}
    \label{fig:qwen-7-short-to-long}
\end{figure*}

\paragraph{ML and baseline.}
We report the average accuracy for each inference type and length combination in both the core and length generalization settings for the Qwen-2.5 models. Figures \ref{fig:qwen-1.5-core}, \ref{fig:qwen-3-core}, and \ref{fig:qwen-7-core} show the accuracies for core generalization for the 1.5B, 3B, and 7B models, respectively, in both the ML and baseline settings. Figures \ref{fig:qwen-1.5-short-to-long}, \ref{fig:qwen-3-short-to-long}, and \ref{fig:qwen-7-short-to-long} show the accuracies for short to long generalization, while Figures \ref{fig:qwen-1.5-long-to-short}, \ref{fig:qwen-3-long-to-short}, and \ref{fig:qwen-7-long-to-short} show the accuracies for long to short generalization for the same models, again in both the ML and baseline settings.

Across model sizes and approaches, the easiest types of inferences are type 2 and type 6. In contrast, types 1, 3, and 4 are typically the most challenging. A notable difference between the ML and baseline models is that the latter consistently struggle with type 5 inferences, whereas the former show stronger performance. However, apart from type 5 inferences, ML models generally perform better but still tend to struggle or excel in similar type and length combinations as the baseline models.

These patterns also hold in the length generalization setting, with the additional observation that performance tends to degrade as the distance between the lengths used for training and those used for testing increases.

\begin{figure*}[htbp]
    \centering
    \includegraphics[width=0.8\linewidth]{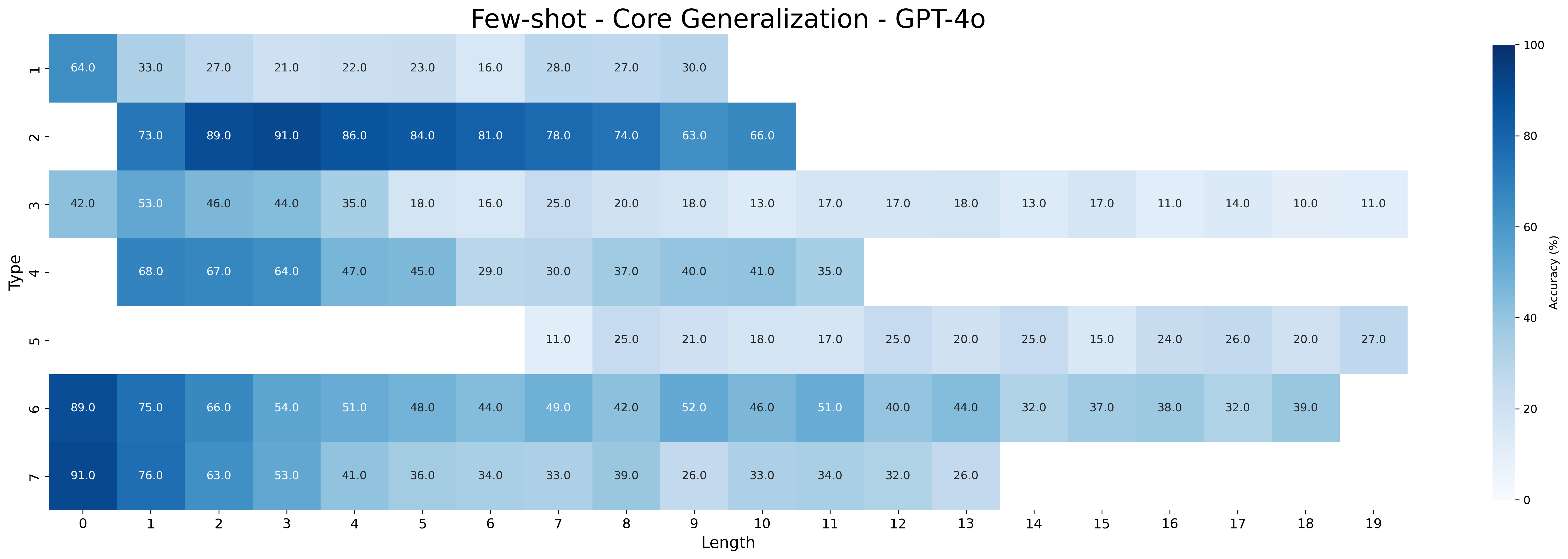} \\[10pt]
    \includegraphics[width=0.8\linewidth]{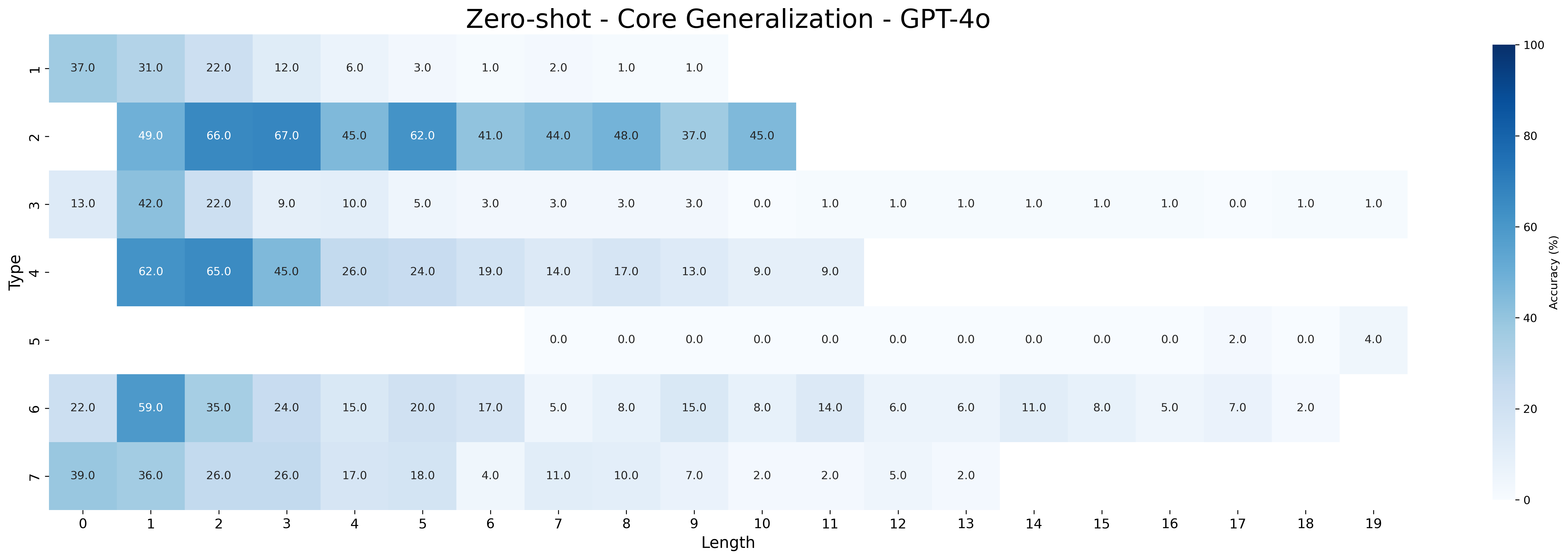}
    \caption{Accuracy of Few-shot (Top) and Zero-shot (Bottom) GPT-4o on core generalization decomposed by inference type and length.}
    \label{fig:gpt-4o-core}
\end{figure*}

\begin{figure*}[htbp]
    \centering
    \includegraphics[width=0.8\linewidth]{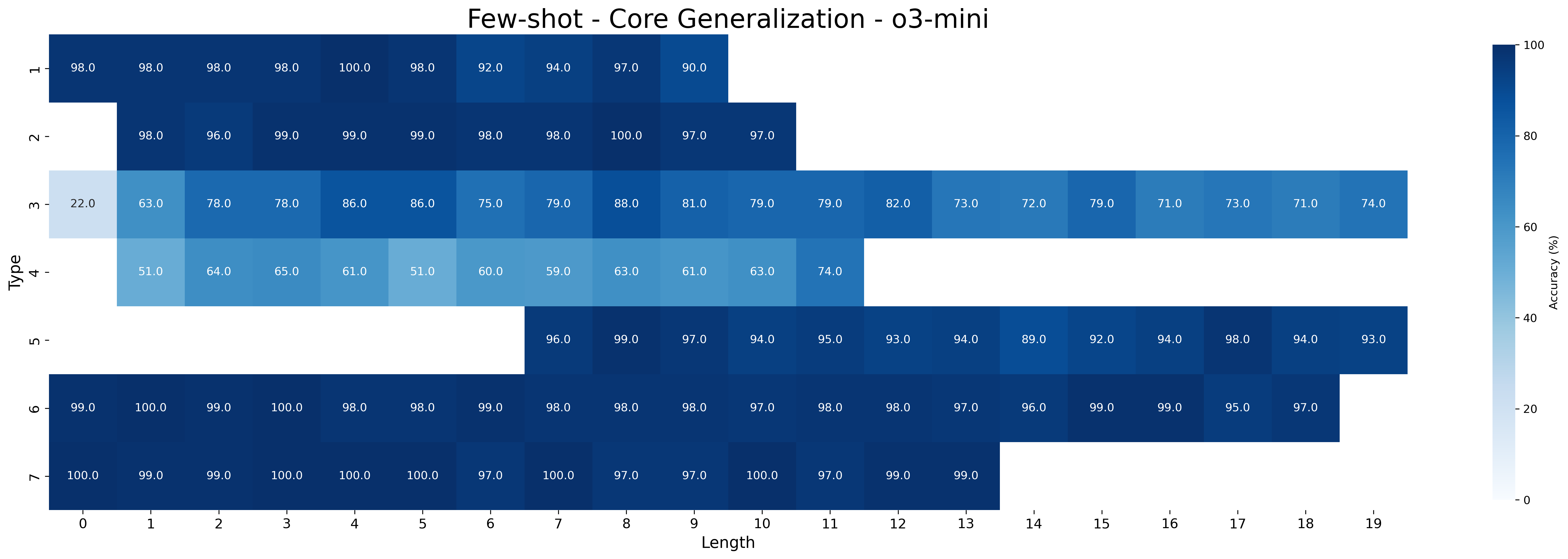} \\[10pt]
    \includegraphics[width=0.8\linewidth]{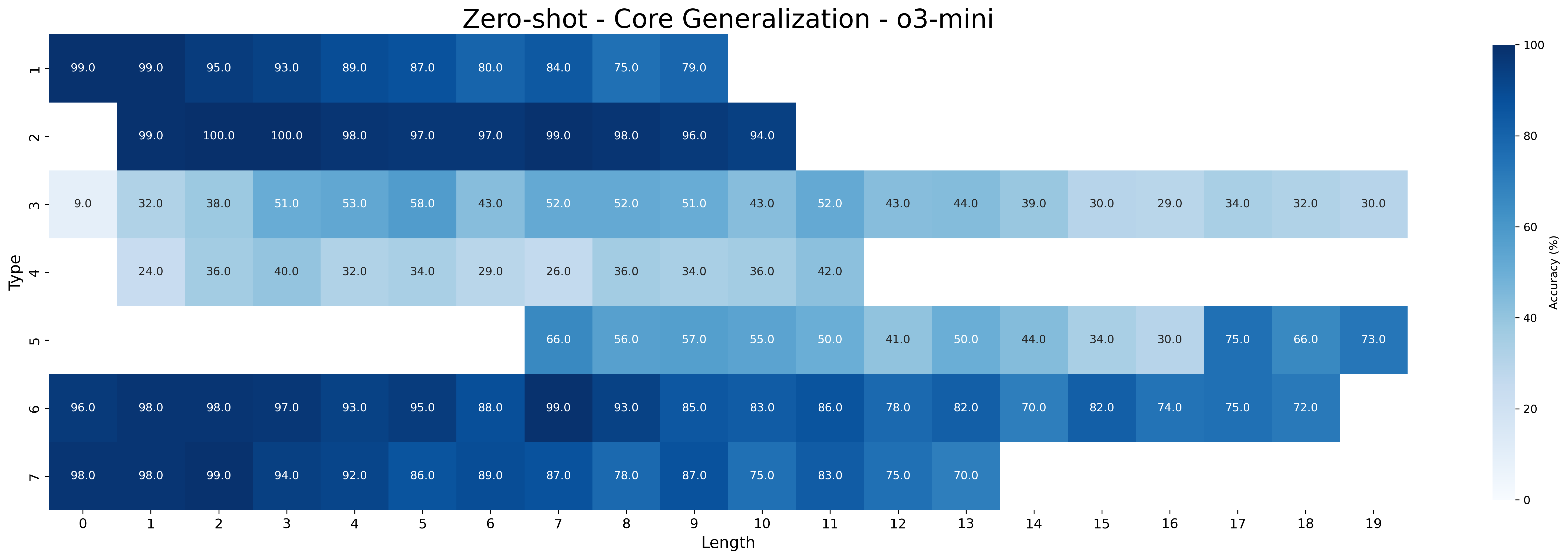}
    \caption{Accuracy of Few-shot (Top) and Zero-shot (Bottom) o3-mini on core generalization decomposed by inference type and length.}
    \label{fig:o3-core}
\end{figure*}

\paragraph{Closed models.}
Figures \ref{fig:gpt-4o-core} and \ref{fig:o3-core} show the accuracies for zero-shot and few-shot prompting of GPT-4o and o3-mini, respectively. Both models show substantial improvement in the few-shot setting. GPT-4o is the lowest-performing model according to Table \ref{tab:core}, a result further supported by the detailed breakdown in this section. It consistently achieves high accuracy only on type 2 inferences, which are the easiest and rely primarily on simple transitivity. o3-mini struggles more with types 3 and 4. Additionally, a clear difference in performance on type 5 inferences is observed between the zero-shot and few-shot settings. This resembles the difference seen in Qwen-2.5 models between ML and baseline. These results show that even pretrained models tend to struggle with the same types of syllogistic inferences as fine-tuned models, with a few exceptions, such as type 5 inferences.

\begin{table*}[t]
  \centering
  \begin{tabular}{llccccc}
    \toprule
    \textbf{Model} & \textbf{Method} & \textbf{NVM} [\%] & \textbf{Avg. NVM} & \textbf{MAP} [\%] & \textbf{Avg. MAP} & \textbf{HP} [\%]  \\
    \midrule
    \multirow{2}{*}{Qwen-2.5 7B} & ML & 17.86 & 2.80 & 80.36 & 3.32 & 75.00 \\
                                     & $\mathrm{Baseline}_{-S}$ & 6.67 & 5.19 & 91.43 & 5.39 & 80.95 \\
                                     
    \midrule
    \multirow{2}{*}{GPT-4o}    & Few-shot & 28.13 & 2.92 & 70.54 & 5.76 & 22.76 \\
                                     & Zero-shot & 14.46 & 3.50 & 83.01 & 6.45 & 17.15 \\
    \cmidrule(lr){2-7}
    \multirow{2}{*}{o3-mini}    & Few-shot & 84.57 & 2.38 & 14.23 & 2.65 & 7.21 \\
                                     & Zero-shot & 76.60 & 2.61 & 22.55 & 7.09 & 2.62 \\
                                     
    \bottomrule
  \end{tabular}
  \caption{\textbf{Error analysis.} Error analysis on core generalization in Qwen-2.5 7B, and the closed models GPT-4o and o3-mini. The table shows percentages and averages for non-minimal valid sets of premises (NVM) and missing necessary $A$ premises (MAP), and the percentage of hallucinated premises (HP).}
  \label{tab:error_analysis_second}
\end{table*}

\subsection{Lexical Generalization}
\label{app:lexical-gen}

In the main body of the paper, we evaluated core and length generalization. Here, we report an additional set of results related to \textbf{lexical generalization}. By \emph{lexical} generalization, we mean the manipulation of the vocabulary assigned to each of the terms appearing in the formulas within $\mathcal{KB}$s.

Section \ref{sec:core} presents results using the same vocabulary of pseudowords employed during training, tested on unseen $\mathcal{KB}$s. Here, we explore two more challenging settings: one using a new vocabulary of pseudowords, and another using abstract symbols (e.g., \texttt{x2435}) in place of pseudowords. This latter setting is distributionally the most distant from the training data.

Table \ref{tab:lexical_gen} presents the results of this lexical generalization experiment. Across all Qwen-2.5 model sizes (1.5B, 3B, 7B) and conditions, the ML approach consistently yields higher accuracy than the baseline, with performance improving with model scale for both approaches. Notably, for both known and unseen pseudowords, performance is similar in both the ML and baseline settings, that is, changing the pseudoword vocabulary has little impact on model performance.

In contrast, for the most challenging generalization setting\textemdash{}unseen constants\textemdash{}both approaches exhibit a significant drop in performance, but the performance gap between ML and the baseline becomes more pronounced: ML achieves 86.87\% at 7B, compared to just 57.81\% for the baseline.

\subsection{Generalization with Limited Data}
\label{app:limited-data}
Table \ref{tab:low_data} presents the performance of the models when trained in a low data regime, using only 100 examples for each combination of inference type and length. Consistent with the findings in Table \ref{tab:lexical_gen} and Table \ref{tab:rec_comp_gen}, ML significantly outperforms the baseline across all model sizes and evaluation metrics. For the core generalization performance, the ML models achieve substantially higher accuracy (e.g., 88.01\% for Qwen-2.5 7B ML vs. 68.54\% for baseline). Similarly, when evaluating generalization to shorter and longer inferences than seen during training, ML models demonstrate a clear advantage. 

Crucially, the performance gap between the meta-learning and baseline approaches is notably wider in this limited data setting compared to the standard data setting. This highlights the enhanced generalization capabilities on limited data induced by meta-learning.

\subsection{Comparison with Prompted Base Model}
\label{app:prompt_model}
To better contextualize the base Qwen-2.5 models' capabilities prior to any fine-tuning, Table \ref{tab:core_prompting} presents the performance of Qwen-2.5 models using few-shot prompting with greedy decoding compared with the performance of the ML and baseline models previously shown in the main paper. The few-shot examples are the same as the one provided to the ML and $\mathrm{Baseline}_{+S}$ models at test time. This comparison allows us to verify that the base models cannot perform premise selection effectively through in-context learning alone, thereby justifying the need for studying how different fine-tuning approaches can affect how the model learns logical inferences.

The results show dramatically poor performance across all model sizes: even the largest 7B model achieves only 11.14\% accuracy overall, while the smallest 1.5B model reaches merely 1.09\%. Although performance improves modestly with model scale, all results remain far below both the fine-tuned baseline and meta-learning approaches, which exceed 85\% accuracy even at the smallest size. Notably, even in this low-performance regime, the pattern observed in fine-tuned models persists: longer inferences remain more difficult than shorter ones across all model sizes (e.g., for the 7B model, accuracy drops from 13.89\% on short inferences to 10.49\% on long ones), further supporting our hypothesis that selecting all premises from a larger minimal set is more challenging.

\begin{table*}[t]
\centering
\begin{tabular}{l l c c c}
\toprule
 \textbf{Model} & \textbf{Method} & \textbf{All} & \textbf{Short} & \textbf{Long} \\
\midrule
 \multirow{2}{*}{Qwen-2.5 1.5B} 
    & ML           & 93.11 ± 0.61     & 94.28 ± 0.61     & 91.76 ± 0.27     \\
    & $\mathrm{Baseline}_{-S}$      & 85.56 ± 1.24     & 91.42 ± 0.82     & 80.56 ± 1.78     \\
    & Few-shot Prompting    & 1.09 ± 0.00    &      2.00 ± 0.00     &   0.51 ± 0.00   \\
\cmidrule(lr){2-5}
 \multirow{2}{*}{Qwen-2.5 3B} 
    & ML           & 96.16 ± 0.44     & 96.24 ± 0.56     & 95.55 ± 0.43     \\
    & $\mathrm{Baseline}_{-S}$       & 93.03 ± 1.15     & 95.34 ± 1.18     & 90.92 ± 1.27     \\
    & Few-shot Prompting    & 4.98 ± 0.00    &      8.20 ± 0.00     &   3.11 ± 0.00   \\
\cmidrule(lr){2-5}
 \multirow{2}{*}{Qwen-2.5 7B} 
    & ML           & 98.13 ± 0.98     & 98.26 ± 0.82     & 97.69 ± 1.40     \\
    & $\mathrm{Baseline}_{-S}$       & 95.76 ± 1.10     & 97.27 ± 1.22     & 94.13 ± 0.90     \\
    & Few-shot Prompting    & 11.14 ± 0.00    &      13.89 ± 0.00     &   10.49 ± 0.00   \\
\bottomrule
\end{tabular}
\caption{\textbf{Comparison with prompted Qwen-2.5 models on core generalization.}  
Accuracy (mean ± std) on test inferences across all type‐length combinations (All), plus breakdown into the five shortest (Short) and longest (Long) inferences for each of the seven types of inference. The table includes the results obtained by a few-shot prompted Qwen-2.5 models using greedy decoding.}
\label{tab:core_prompting}
\end{table*}

\subsection{Providing Study Examples to the Baseline at Inference Time}
\label{app:baseline-comparison}

Table~\ref{fig:baseline_comparison} presents a direct comparison between the two baseline variants: $\mathrm{Baseline}_{-S}$ (without study examples at inference time) and $\mathrm{Baseline}_{+S}$ (with study examples). In the main paper, we reported only $\mathrm{Baseline}_{-S}$ results, as it consistently performed better\textemdash{}or at least on par on average\textemdash{}across the settings highlighted there. Here, we provide a more detailed analysis.

Overall, both baseline variants underperform relative to the meta-learning approach, confirming that access to structured meta-learning episodes is crucial for systematic generalization. However, incorporating study examples into the baseline input does confer benefits in specific scenarios. In particular, $\mathrm{Baseline}_{+S}$ can outperform $\mathrm{Baseline}_{-S}$ in the low-data regime. This is likely because baseline models trained with limited data benefit from both training on task examples and leveraging their previously developed in-context learning capabilities, enabling better generalization from demonstrations of the inference task. For example, Qwen-2.5 3B achieves 75.80\% (with study examples) versus 66.51\% (without). A similar pattern emerges in the lexical generalization setting: $\mathrm{Baseline}_{+S}$ exhibits stronger robustness to unseen constants, which represent the largest distributional shift in vocabulary, again suggesting that study examples help the model generalize in-context from the learned vocabulary to novel symbols. For instance, Qwen-2.5 7B improves from 57.81\% ($-S$) to 74.28\% ($+S$) when evaluated on unseen constants.

In the length generalization case, the effect of study examples depends critically on their alignment with test inference lengths. Specifically, disaligned study examples consistently reduce performance relative to $\mathrm{Baseline}_{-S}$. For instance, in the Short~$\rightarrow$~Long task for Qwen-2.5 1.5B, disaligned examples yield 51.49\% versus 63.53\% without study examples, indicating that irrelevant in-context examples can confuse the model, particularly for longer inference lengths. Conversely, aligned study examples improve performance over disaligned ones, demonstrating that the baseline can exploit in-context cues when examples match the target length distribution. However, even with aligned examples, $\mathrm{Baseline}_{+S}$ rarely surpasses $\mathrm{Baseline}_{-S}$ for larger models (e.g., Qwen-2.5 7B, Short~$\rightarrow$~Long: 79.90\% aligned versus 80.76\% without examples).

Since $\mathrm{Baseline}_{-S}$ tends to perform better in core generalization tasks, and the benefits of study examples at test time are inconsistent in length generalization scenarios, we reported $\mathrm{Baseline}_{-S}$ results in the main paper.

\subsection{Additional Error Analysis}
\label{app:additional_errors}
In this section, we present the additional error analysis results for Qwen-2.5 7B both in ML and baseline setting on the core generalization experiment. Additionally, we also show the error analysis results for GPT-4o and o3-mini. The detailed breakdown of these errors is presented in Table \ref{tab:error_analysis_second}.

\paragraph{ML and baseline.}
For the Qwen-2.5 7B model, ML shows a higher percentage of non-minimal valid set of premises (NVM) errors (17.86\%) compared to the baseline (6.67\%) on core generalization. However, when these NVM errors occur, ML includes fewer unnecessary premises on average (Avg. NVM of 2.80) than the baseline (Avg. NVM of 5.19). Conversely, the baseline model exhibits a higher proportion of errors due to missing necessary A premises (MAP) at 91.43\%, with an average of 5.39 missing premises. This is higher than ML, which has a MAP percentage of 80.36\% and an average of 3.32 missing premises. Both methods show high rates of hallucinated premises (HP), with ML at 75.00\% and the baseline slightly higher at 80.95\%. These results suggest that not only ML has generally a higher core generalization performance than the baseline, but also that ML errors tend to be closer to the correct set of premises.

\paragraph{Closed models.}
The error analysis for closed models reveals distinct patterns for GPT-4o and o3-mini. For GPT-4o, MAP errors are predominant in both few-shot (70.54\%) and zero-shot (83.01\%) settings. The average number of missing $A$ premises is also high (5.76 for few-shot and 6.45 for zero-shot) and indicates that the model struggles to provide all the necessary premises to derive hypotheses.

In contrast, o3-mini primarily struggles with NVM errors, which constitute 84.57\% of errors in the few-shot setting and 76.60\% in the zero-shot setting. The average number of unnecessary premises is relatively low and similar in both settings (2.38 for few-shot, 2.61 for zero-shot). This shows that the model is capable of providing logically valid set of premises from which hypotheses can be derived but, on the other hand, struggles with the concept of minimality. An interesting characteristic of o3-mini is its very low HP rate, at 7.21\% for few-shot and an even lower 2.62\% for zero-shot, which is considerably better than both Qwen-2.5 7B and GPT-4o.

\begin{figure*}[t]
\centering

% --- Core Generalization ---
\begin{subfigure}{\textwidth}
\centering
\begin{tabular}{c l l c c c}
\toprule
 & \textbf{Model} & \textbf{Method} & \textbf{All} & \textbf{Short} & \textbf{Long} \\
\midrule
\multirow{6}{*}{\rotatebox[origin=c]{90}{Core}} 
 & \multirow{2}{*}{Qwen-2.5 1.5B} 
              & $\mathrm{Baseline}_{-S}$ & 85.56 ± 1.24 & 91.42 ± 0.82 & 80.56 ± 1.78 \\
 &            & $\mathrm{Baseline}_{+S}$ & 77.89 ± 3.13 & 85.93 ± 2.17 & 70.58 ± 4.45 \\
\cmidrule(lr){2-6}
 & \multirow{2}{*}{Qwen-2.5 3B} 
              & $\mathrm{Baseline}_{-S}$ & 93.03 ± 1.15 & 95.34 ± 1.18 & 90.92 ± 1.27 \\
 &            & $\mathrm{Baseline}_{+S}$ & 87.18 ± 4.32 & 91.98 ± 2.24 & 82.99 ± 6.41 \\
\cmidrule(lr){2-6}
 & \multirow{2}{*}{Qwen-2.5 7B} 
              & $\mathrm{Baseline}_{-S}$ & 95.76 ± 1.10 & 97.27 ± 1.22 & 94.13 ± 0.90 \\
 &            & $\mathrm{Baseline}_{+S}$ & 92.91 ± 1.85 & 95.59 ± 1.06 & 90.29 ± 2.68 \\
\bottomrule
\end{tabular}
\caption{\textbf{Core generalization.}}
\end{subfigure}

\vspace{0.5cm}

% --- Generalization to Lengths ---
\begin{subfigure}{\textwidth}
\centering
\begin{tabular}{l l c c c c}
\toprule
\multirow{2}{*}{\textbf{Model}} & \multirow{2}{*}{\textbf{Method}} & \multicolumn{2}{c}{\textbf{Short $\rightarrow$ Long}} & \multicolumn{2}{c}{\textbf{Long $\rightarrow$ Short}} \\
\cmidrule(lr){3-4} \cmidrule(lr){5-6}
 & & \textbf{Disaligned} & \textbf{Aligned} & \textbf{Disaligned} & \textbf{Aligned} \\
\midrule
\multirow{2}{*}{Qwen-2.5 1.5B} 
              & $\mathrm{Baseline}_{-S}$ & 63.53 ± 1.16 & 63.53 ± 1.16 & 56.67 ± 1.22 & 56.67 ± 1.22 \\
              & $\mathrm{Baseline}_{+S}$ & 51.49 ± 2.30 & 62.88 ± 4.59 & 60.38 ± 1.38 & 61.90 ± 2.08 \\
\cmidrule(lr){1-6}
\multirow{2}{*}{Qwen-2.5 3B} 
              & $\mathrm{Baseline}_{-S}$ & 76.78 ± 1.63 & 76.78 ± 1.63 & 71.88 ± 1.49 & 71.88 ± 1.49 \\
              & $\mathrm{Baseline}_{+S}$ & 65.98 ± 4.02 & 69.55 ± 0.69 & 74.29 ± 1.95 & 74.84 ± 2.70 \\
\cmidrule(lr){1-6}
\multirow{2}{*}{Qwen-2.5 7B} 
              & $\mathrm{Baseline}_{-S}$ & 80.76 ± 2.65 & 80.76 ± 2.65 & 71.08 ± 1.55 & 71.08 ± 1.55 \\
              & $\mathrm{Baseline}_{+S}$ & 75.34 ± 2.35 & 79.90 ± 1.53 & 73.88 ± 2.70 & 76.04 ± 1.95 \\
\bottomrule
\end{tabular}
\caption{\textbf{Generalization to unseen lengths.}}
\end{subfigure}

\vspace{0.5cm}

% --- Low Data ---
\begin{subfigure}{\textwidth}
\centering
\begin{tabular}{l l c c c}
\toprule
\textbf{Model} & \textbf{Method} & \textbf{Core} & \textbf{L$\rightarrow$S} & \textbf{S$\rightarrow$L} \\
\midrule
\multirow{2}{*}{Qwen-2.5 1.5B} 
              & $\mathrm{Baseline}_{-S}$ & 55.14 ± 0.53 & 29.37 ± 1.85 & 30.22 ± 1.52 \\
              & $\mathrm{Baseline}_{+S}$ & 66.18 ± 0.51 & 42.42 ± 1.48 & 28.35 ± 4.65 \\
\cmidrule(lr){1-5}
\multirow{2}{*}{Qwen-2.5 3B} 
              & $\mathrm{Baseline}_{-S}$ & 66.51 ± 0.19 & 43.66 ± 1.93 & 43.67 ± 2.05 \\
              & $\mathrm{Baseline}_{+S}$ & 75.80 ± 0.15 & 58.50 ± 1.79 & 43.62 ± 4.13 \\
\cmidrule(lr){1-5}
\multirow{2}{*}{Qwen-2.5 7B} 
              & $\mathrm{Baseline}_{-S}$ & 68.54 ± 2.25 & 45.27 ± 0.95 & 43.94 ± 2.82 \\
              & $\mathrm{Baseline}_{+S}$ & 74.69 ± 3.16 & 57.42 ± 3.07 & 45.60 ± 4.56 \\
\bottomrule
\end{tabular}
\caption{\textbf{Low-data regime.}}
\end{subfigure}

\vspace{0.5cm}

% --- Lexical ---
\begin{subfigure}{\textwidth}
\centering
\begin{tabular}{l l c c c}
\toprule
\textbf{Model} & \textbf{Method} & \textbf{Core} & \textbf{Unseen Pseudowords} & \textbf{Unseen Constants} \\
\midrule
\multirow{2}{*}{Qwen-2.5 1.5B} 
              & $\mathrm{Baseline}_{-S}$ & 85.56 ± 1.24 & 83.34 ± 1.90 & 38.49 ± 1.06 \\
              & $\mathrm{Baseline}_{+S}$ & 77.89 ± 3.13 & 76.85 ± 2.85 & 54.22 ± 5.75 \\
\cmidrule(lr){1-5}
\multirow{2}{*}{Qwen-2.5 3B} 
              & $\mathrm{Baseline}_{-S}$ & 93.03 ± 1.15 & 91.49 ± 0.68 & 53.12 ± 2.03 \\
              & $\mathrm{Baseline}_{+S}$ & 87.18 ± 4.32 & 85.94 ± 4.11 & 68.71 ± 6.74 \\
\cmidrule(lr){1-5}
\multirow{2}{*}{Qwen-2.5 7B} 
              & $\mathrm{Baseline}_{-S}$ & 95.76 ± 1.10 & 94.89 ± 1.55 & 57.81 ± 2.17 \\
              & $\mathrm{Baseline}_{+S}$ & 92.91 ± 1.85 & 92.13 ± 1.93 & 74.28 ± 2.20 \\
\bottomrule
\end{tabular}
\caption{\textbf{Lexical generalization.}}
\end{subfigure}

\caption{\textbf{Baseline models with and without study examples.} Each subfigure compares $\mathrm{Baseline}{-S}$ (without study examples at inference) and $\mathrm{Baseline}{+S}$ (with study examples). Results are shown for (a) core generalization, (b) generalization to unseen lengths, (c) low-data regime, and (d) lexical generalization. While $\mathrm{Baseline}{-S}$ generally performs better in fully trained settings, $\mathrm{Baseline}{+S}$ shows advantages in low-data and lexical generalization (unseen constants) scenarios, though both remain below meta-learning performance.}
\label{fig:baseline_comparison}
\end{figure*}

\end{document}

%% file: tikz/schema.tex
\begin{tikzpicture}[
    node distance=0.8cm,
    box/.style={
        draw=none,
        rounded corners,
        minimum width=3cm,
        minimum height=1.2cm,
        align=center,
        font=\sffamily,
        fill opacity=0.95
    },
    dashedbox/.style={
        draw=black,
        dashed,
        rounded corners,
        align=center
    },
    title/.style={font=\small\bfseries\sffamily},
    subtitle/.style={font=\scriptsize\itshape\sffamily},
    content/.style={font=\scriptsize\sffamily}
]

% Big Episode Box (Outer Container)
\node[dashedbox, minimum width=7.5cm, minimum height=7.5cm] (bigbox) at (0,0) {};
\node[title] at ($(bigbox.north west)+(1cm,0.3cm)$) {Episode $\mathcal{T}$};

% Knowledge Base
\node[box, fill=gray!20, minimum width=7.0cm, minimum height=2.0cm] (kb) at ($(bigbox.north)+(0,-1.3cm)$) {};
\node[title] at ($(kb.north)+(0,-0.3cm)$) {Knowledge Base ($\mathcal{KB}$)};
\node[content, align=left] at ($(kb.center)+(0,-0.2)$) {
    \textbf{knowledge base:} All x1 are x2, All x2 are x4, All x3 are x5, \\[2pt]
     All x10 are x11, All x4 are x6, All x2 are x3, All x5 are x7, \\[2pt]
     Some x5 are not x1, All x9 are x10, All x6 are x8, All x8 are x9, \\[2pt]
     Some x11 are not x4
};

% Study Examples
\node[box, fill=blue!20, minimum width=6.2cm, minimum height=2.0cm] (study) at ($(kb.south)+(0,-1.2cm)$) {};
\node[title] at ($(study.north)+(0,-0.3cm)$) {Study Examples ($S^{\text{supp}}$)};
\node[content, align=left] at ($(study.south)+(0,0.75)$) {
    \textbf{<STUDY> hypothesis:} All x8 are x11 \\[2pt]
    \textbf{premises:} All x8 are x9, All x9 are x10, All x10 are x11; \\[2pt]
    \textbf{hypothesis:} All x1 are x3 \\[2pt]
    \textbf{premises:} All x1 are x2, All x2 are x3; ...
};

% Query Hypothesis
\node[box, fill=orange!25, minimum width=4.6cm, minimum height=1cm] (query) at ($(study.south)+(0,-0.7cm)$) {};
\node[title] at ($(query.north)+(0,-0.3cm)$) {Query Hypothesis ($x^{\text{query}}$)};
\node[content] at ($(query.south)+(0,0.3cm)$) {\textbf{<QUERY> hypothesis:} All x3 are x7};

% Predicted Premises
\node[box, fill=green!20, minimum width=5.0cm, minimum height=1.0cm] (premises) at ($(query.south)+(0,-1.0cm)$) {};
\node[title] at ($(premises.north)+(0,-0.3cm)$) {Query Premises ($y^{\text{query}}$)};
\node[content] at ($(premises.south)+(0,0.3cm)$) {\textbf{premises:} All x3 are x5, All x5 are x7};

% Input Box (includes KB, Study, Query)
\node[dashedbox, minimum width=7.2cm, minimum height=5.7cm] (inputbox) at ($(bigbox.north)+(0,-3.0cm)$) {}; % ($(kb)!0.5!(query)$)
\node[subtitle, rotate=90] at ($(inputbox.west)+(0.25cm,0.0cm)$) {\scriptsize\bfseries\sffamily Input};

% Output Box
\node[dashedbox, minimum width=7.2cm, minimum height=1.3cm] (outputbox) at ($(premises)$) {};
\node[subtitle, rotate=90] at ($(outputbox.west)+(0.25cm,0.0cm)$) {\scriptsize\bfseries\sffamily Output};

% Arrows
%\draw[-, thick] (kb.south) -- (study.north);
%\draw[-, thick] (study.south) -- (query.north);
%\draw[-, thick, dashed] (inputbox.south) -- (outputbox.north);

\end{tikzpicture}

%% file: tikz/generalization.tex
\begin{tikzpicture}[
  node distance=0.3cm and 1cm,
  train/.style={
  fill=gray!10, text=black,
  text width=0.42\linewidth, align=left, font=\scriptsize\ttfamily,
  rounded corners, inner sep=6pt
  },
  test/.style={
  fill=gray!10, text=black,
  text width=0.42\linewidth, align=left, font=\scriptsize\ttfamily,
  rounded corners, inner sep=6pt
  },
  labelbox/.style={
    draw=none, font=\bfseries\sffamily\small, align=left
  },
  arrow/.style={
    -{Latex[length=2mm]}, thick, color=black
  }
]
% Column headers
\node[labelbox] at (0, 5) {Training};
\node[labelbox, right=7.4cm] at (0, 5) {Testing};

% Row 1
\node[train=white] (train1) at (0,4) {
\textbf{Longer inferences:}\\[3pt]
``\textcolor{blue!90!black}{\textbf{all x1 are x2, all x2 are x3}}, all x3 are x4, all x4 are x5, all x5 are x6 $\vdash$ all x1 are x6''
};
\node[test=white, right=of train1] (test1) {
\textbf{Shorter inferences:}\\[3pt]
``\textcolor{blue!90!black}{\textbf{all x1 are x2, all x2 are x3}} $\vdash$ all x1 are x3''
};

% Row 2
\node[train=white] (train2) at (0,2.3) {
\textbf{Shorter inferences:}\\[3pt]
``\textcolor{red!90!black}{\textbf{all x1 are x2, all x2 are x3, all x3 are x4}} $\vdash$ all x1 are x4''
};
\node[test=white, right=of train2] (test2) {
\textbf{Longer inferences:}\\[3pt]
``\textcolor{red!90!black}{\textbf{all x1 are x2, all x2 are x3, all x3 are x4}}, all x4 are x5, all x5 are x6 $\vdash$ all x1 are x6''
};

% Arrows
\draw[->, thick] (train1) -- (test1);
\draw[->, thick] (train2) -- (test2);

\end{tikzpicture}